\documentclass[10pt,twocolumn,letterpaper]{article}

\usepackage{cvpr}
\usepackage{times}
\usepackage{epsfig}
\usepackage{graphicx}
\usepackage{amsmath}
\usepackage{amssymb}
\usepackage{subcaption}
\usepackage{xcolor,colortbl}

\usepackage[pagebackref=true,breaklinks=true,letterpaper=true,colorlinks,bookmarks=false]{hyperref}

\cvprfinalcopy 

\def\cvprPaperID{1386} 

\ifcvprfinal\fi
\begin{document}

\title{segDeepM: Exploiting Segmentation and Context in Deep Neural Networks for Object Detection}

\author{Yukun Zhu \quad
Raquel Urtasun \quad
Ruslan Salakhutdinov \quad
Sanja Fidler\\
University of Toronto\\
{\tt\small \{yukun,urtasun,rsalakhu,fidler\}@cs.toronto.edu}
}

\maketitle

\begin{abstract}
In this paper, we propose an approach that exploits object segmentation in order to improve the accuracy of object detection. We frame the problem as inference in a Markov Random Field, in which each detection hypothesis scores object appearance as well as contextual information using Convolutional Neural Networks, and allows the hypothesis to choose and score a segment out of a large pool of accurate object segmentation proposals. This enables the detector to incorporate additional evidence when it is available and thus results in more accurate detections. Our experiments show an improvement of $4.1\%$ in mAP over the R-CNN baseline on PASCAL VOC 2010, and $3.4\%$ over the current state-of-the-art, demonstrating the power of our approach.
\end{abstract}

\section{Introduction}

In the past two years, Convolutional Neural Networks (CNNs) have revolutionized computer vision. They have been applied to a variety of general vision problems,
 such as recognition~\cite{krizhevsky2012imagenet,girshick2013rich}, segmentation~\cite{HariharanSimultaneousECCV2014}, stereo~\cite{Memisevic}, flow~\cite{Weinzaepfel}, and even text-from-image generation~\cite{Kiros}, consistently outperforming past work. This is mainly due to their high generalization power  achieved by learning complex, non-linear dependencies across millions of labelled examples. 

It has recently been shown that 
increasing the depth of the network 
increases the performance by an additional impressive margin on the ImageNet challenge~\cite{verydeep,GoogLeNet}. It remains to be seen whether recognition can be solved by simply pushing the limits of computation (the size of the networks) and increasing the amount of the training data. 
We believe that the main challenge in the next few years will be to design 
computationally simpler and more efficient models that can achieve a similar or better performance 
compared to the very deep networks.

\begin{figure}
\includegraphics[width=1\linewidth,trim=0 120 50 0,clip=true]{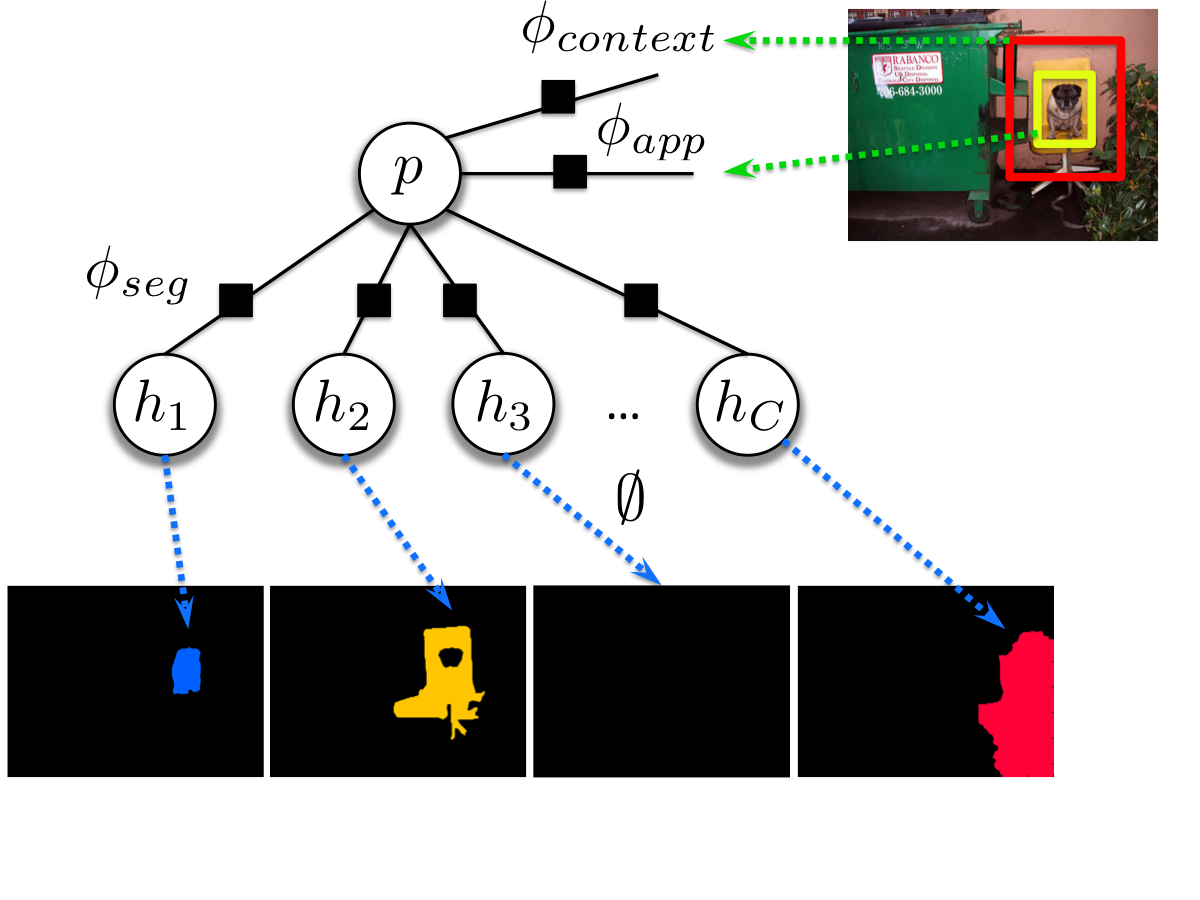}
\vspace{-5mm}
\caption{Proposed segDeepM model.}
\label{figure:intro}
\vspace{-2mm}
\end{figure}

For object detection, a successful approach has been to generate a large pool of candidate boxes~\cite{van2011segmentation} and classify them using CNNs~\cite{girshick2013rich}.  The quality of such a detector thus largely depends on the quality of the object hypotheses. Interestingly, however, using much better proposals obtained via a high-end bottom-up segmentation approach~\cite{HariharanSimultaneousECCV2014} has resulted only in small improvements in accuracy. 

In this paper, we show how to exploit a small number of accurate object segment proposals in order to significantly improve object detection performance. We frame the detection problem as inference in a Markov Random Field as in Figure~\ref{figure:intro},
in which each detection hypothesis scores object appearance as well as 
contextual information using Convolutional Neural Networks. 
Each hypothesis can choose and score a segment out of a small pool of accurate object segmentation proposals. 
This enables our approach to place more accurate object bounding boxes in parts of the image where an object segmentation hypothesis~\cite{carreira2012semantic} exists or where strong contextual cues are available.  We additionally show that a significant performance boost can be obtained by a sequential approach, where the network iterates between adjusting its spatial scope (the bounding box) and classifying its content. This strategy reduces the dependency on the initial candidate boxes obtained by~\cite{van2011segmentation} and enables our approach to recover from 
the potentially bad initial localization.

We show that our model, called segDeepM, outperforms the baseline R-CNN~\cite{girshick2013rich} approach by $3.2\%$ with almost no extra computational cost. 
We get a total of  $5\%$ improvement by incorporating contextual information at the cost of doubling the running time of the method. 
On PASCAL VOC 2010 test, our method achieves $4.1\%$ improvement over R-CNN and $1.4\%$ over the current state-of-the-art. 

\section{Related Work}

In the past years, a variety of  segmentation algorithms that exploit object detections as a top-down cue have been explored. The standard approach has been to use detection features as unary potentials in an MRF~\cite{koltun11}, or as candidate bounding boxes for holistic MRFs~\cite{Yao12,ladicky10}. In~\cite{parkhi11},  segmentation within the detection boxes has been performed using a GrabCut method. In~\cite{bourdev10}, object segmentations are found by aligning the masks obtained from Poselets~\cite{bourdev10,maire11}.

There have been a few approaches to  use segmentation to improve object detection.~\cite{gu09} cast votes for the object's location by using a Hough transform with a set of regions.~\cite{dai12} uses DPM to find a rough object location and refines it  according to color information and occlusion boundaries. In~\cite{HariharanSimultaneousECCV2014}, segmentation is used to mask-out the background inside the detection, resulting in improved performance. Segmentation and detection has also been addressed in a joint formulation in~\cite{yang11} by combining shape information obtained via DPM parts as well as color and boundary cues.

Our work is inspired by the success of segDPM~\cite{FidlerBottomCVPR2013}. By augmenting the DPM detector~\cite{felzenszwalb2010object} with very simple segmentation features that can be computed in constant time, segDPM improved the detection performance by $8\%$ on the challenging PASCAL VOC dataset. 
The approach used segments computed from the final segmentation output of CPMC~\cite{carreira2012semantic} in order to place accurate boxes in parts of the image where segmentation for the object class of interest was available. This idea was subsequently exploited in~\cite{mottaghirole} by augmenting the DPM with an additional set of deformable context ``parts'' which scored contextual segmentation features around the object. In~\cite{PartsCVPR14}, the segDPM detector~\cite{FidlerBottomCVPR2013} was augmented with part visibility reasoning, achieving state-of-the-art results for detection of articulated classes. In~\cite{dong2014towards}, the authors extended segDPM to incorporate segmentation compatibility also at the part level. 

In this paper, we 
build on R-CNN framework~\cite{girshick2013rich} and transfer the core ideas of segDPM. 
We use appearance features from~\cite{krizhevsky2012imagenet,girshick2013rich}, a rich contextual appearance description around the object, and a MRF model that is able to exploit segmentation in a more efficient way than segDPM.

\vspace{-1mm}
\section{Our Approach}

The goal of our approach is to efficiently exploit segmentation and contextual cues in order to facilitate object detection. Following the R-CNN setup, we compute the Selective Search boxes~\cite{van2011segmentation} yielding approximately 2000 object candidates per image. For each box we extract the last feature layer of the CNN network~\cite{krizhevsky2012imagenet}, that is fine-tuned on the PASCAL dataset as proposed in~\cite{girshick2013rich}. We obtain object segment proposals via the CPMC approach~\cite{carreira2010constrained}, although our approach is independent of this choice. Following~\cite{carreira2012semantic}, we take the top $150$ proposals given by an object-independent ranker, and train class-specific classifiers for all classes of interest by the second-order pooling method O2P~\cite{carreira2012semantic}. We remove all segments that have less than $1500$ pixels. Our method will make use of these segments along with their class-specific scores. This is slightly different than segDPM which takes only 1 or 2 segments carved out from the final O2P's pixel-level labeling of the image.

In the remainder of this section we first define our model and describe its segmentation and contextual features. We next discuss inference and learning. Finally, 
we detail a sequential inference scheme that iterates between correcting the input bounding boxes and scoring them with our model.

\subsection{The segDeepM Model}
We define our model as a Markov Random Field with random variables that reason about detection boxes, object segments, and context. Similar to~\cite{felzenszwalb2010object,FidlerBottomCVPR2013}, we define $p$ as a random variable denoting the location and scale of a candidate bounding box in the image.  
We also define ${\bf h}$ to be a set of random variables, one for each class, i.e. ${\bf h}=(h_1,h_2,\ldots,h_C)^T$. 
Each random variable $h_c\in\{0,1,\ldots,H({\bf x})\}$ represents an index into 
the set of all candidate segments. Here $C$ is the total number of object 
classes of interest and $H({\bf x})$ is the total number of segments in image $x$. 
The random variable $h_c$ allows each candidate detection box to \emph{choose} a segment for each class and score its confidence according to the agreement with the segment. 
The idea is to (1) boost the confidence of boxes that are well aligned with a 
high scoring object region proposal for the class of interest, and
(2) adjust its score based on the proximity and confidence of region proposals for other classes, serving as context for the model.  
This is different from segDPM that only had a single random variable $h$ which selected a segment belonging to the detector's class. It is also different from~\cite{mottaghirole} in that the model chooses contextual segments, and does not score context in a fixed segmentation window. Note that $h_c=0$ indicates that no segment is selected for class $c$. This means that either no segment for a class of interest is in the vicinity of the detection hypothesis, or that none of the regions corresponding to the contextual class $c$ help classification of the current box. We define the energy of a configuration as follows:
\begin{align}
\label{eq:energy}
E(p,{\bf h};{\bf x})&= \omega_{app}^T \cdot \phi_{app}(p;{\bf x}) + \omega_{seg}^T \cdot \phi_{seg}(p,{\bf h};{\bf x})  \\&+ \omega_{ctx}^T \cdot \phi_{ctx}(p;{\bf x}), \nonumber 
\end{align}
where $\phi_{app}(x,p)$, $\phi_{seg}(p,{\bf h};{\bf x})$, and $\phi_{ctx}(p;{\bf x})$  are the candidate's appearance, segmentation, and contextual 
potential functions (features), respectively. We describe the potentials in detail below.

\subsection{Details of Potential Functions}

{\bf Appearance:} To extract the appearance features we follow~\cite{girshick2013rich}. The image in each candidate detection's box is warped to a fixed size $227\times227\times 3$. We run the image through the CNN~\cite{krizhevsky2012imagenet} trained on the ImageNet dataset and fine-tuned on PASCAL's data~\cite{girshick2013rich}. As our appearance feature $\phi_{app}(p;{\bf x})$ we use the $4096$-dimensional feature extracted from the $fc7$ layer.

{\bf Segmentation:} Similar to~\cite{FidlerBottomCVPR2013}, our segmentation features 
attempt to capture the agreement between the candidate's bounding box and a particular segment. The features are complementary in nature, and, when combined within the model, aim at placing the box tightly around each segment. We emphasize that the weights for each feature will be learned, thus allowing the model to adjust the importance of each feature's contribution to the joint energy.  

We use slightly more complex features tailored to exploit a much larger set of segments than~\cite{FidlerBottomCVPR2013}. In particular, we use a grid feature that aims to capture a loose geometric arrangement of the segment inside the candidate's box. We also incorporate class information, where the model is allowed to choose a different segment for each class, depending on the contextual information contained in a segment with respect to the class of the detector. 

We use multiple segmentation features, one for each class, thus our segmentation term decomposes:
\begin{equation}
\omega_{seg}^T \cdot \phi_{seg}(p,{\bf h};{\bf x})=\hspace{-2mm}\sum_{c\in\{1,\dots,C\}}\sum_{type} \omega_{type}^T \cdot \phi_{type}(p,h_c;{\bf x}).
\nonumber 
\end{equation}

Specifically, we consider the following features:

\textbf{SegmentGrid-In:} Let $S(h_c)$ denote the binary mask of the segment chosen by $h_c$. For a particular candidate box~$p$, we crop the segment's mask via the bounding box of $p$ and compute the SegmentGrid-in feature on a $K \times K$ grid $\mathcal{G}$ placed over the cropped mask. The $k^{th}$ dimension represents the percentage of segment's pixels inside the $k^{th}$ block, relative to the number of all pixels in $S(h_c)$.
\begin{equation}\label{eqn:seg-in}
\phi_{seggrid-in}(x,p,h_c,k) = \frac{1}{|S(h_c)|}\sum_{i\in \mathcal{G}(p,k)}S(h_c,i), 
\end{equation}
where $\mathcal{G}(p,k)$ is the $k^{th}$ block of pixels in grid $\mathcal{G}$, and $S(h_c,i)$ indexes the segment's mask in pixel $i$. That is, $S(h_c,i)=1$ when pixel $i$ is part of the segment and $S(h_c,i)=0$ otherwise. For $c$ matching the detector's class, this feature will attempt to place a box slightly bigger than the segment while at the same time trying to localize it such that the spatial distribution of pixels within each grid matches the class' expected shape. For $c$ other than the detector's class, this feature will try to place the box such that it intersects as little as possible with the segments of other classes. The dimensionality of this feature is $K\times K\times C$.

\textbf{Segment-Out:} This feature follows~\cite{FidlerBottomCVPR2013}, and computes the percentage of segment pixels outside the candidate  box. Unlike the SegmentGrid-In, this feature computes a single value for each segment/bounding box pair. 
\begin{equation}\label{eqn:seg-out}
\phi_{seg-out}(p,h) = \frac{1}{|S(h)|}\sum_{i\not\in \mathcal{B}(p)}S(h,i),
\end{equation}
where $\mathcal{B}(p)$ is the bounding box corresponding to $p$. The aim of this feature is to place boxes that are smaller compared to the segments, which, in combination with SegmentGrid-In, achieves a tight fit around the segments.

\textbf{BackgroundGrid-In:} This feature is also computed with a $K \times K$ grid $\mathcal{G}$ for each bounding box $p$. We compute the percentage of pixels in each grid cell that are {\bf not} part of the segment:
\begin{equation}\label{eqn:back-in}
\phi_{back-in}(p,h,k) = \frac{1}{M-|S(h)|}\sum_{i\in \mathcal{G}(p,k)}\left(1-S(h,i)\right),
\end{equation}
with $M$ the area of the largest segment for the image.

\textbf{Background-Out:} This scalar feature measures the $\%$ of segment's background outside of the candidate's box:
\begin{equation}\label{eqn:seg-out}
\phi_{back-out}(p,h) = \frac{1}{M-|S(h)|}\sum_{i\not\in \mathcal{B}(p)}\left(1-S(h,i)\right).
\end{equation}

\textbf{Overlap:} Similarly to~\cite{FidlerBottomCVPR2013}, we use another feature to measure the alignment of the candidate's box and the segment $S(h)$. It is computed as the intersection-over-union (IOU) between the  box or $p$ and a tightly fit bounding box around the segment $S(h)$.
\begin{equation}\label{eqn:overlap}
\phi_{overlap}(x,p,h) = \frac{\mathcal{B}(p)\cap \mathcal{B}(S(x,h))}{\mathcal{B}(p)\cup \mathcal{B}(S(x,h))} -\lambda,
\end{equation}
where $\mathcal{B}(S(x,h))$ is tight box around $S(x,h)$, and $\lambda$ a bias term which we set to $\lambda = -0.7$ in our experiments.

\textbf{SegmentClass:} Since we are dealing with many segments per image, we add an additional feature to our model.  We train the O2P~\cite{carreira2012semantic} rankers for each class which uses several region-aware features as input into our segmentation features. Each ranker is trained to predict the IOU overlap of the given segment with the ground-truth object's segment. The output of all the class-specific rankers defines the following feature:
\begin{equation}\label{eqn:overlap}
\phi_{potential}(h,c) =  \frac{1}{1+e^{-s(h,c)}},
\end{equation}
where $s(h,c)$ is the score of class $c$ for segment $S(h)$.

SegmentGrid-in, segment-out, backgroundGrid-in, and background-out can be efficiently computed via integral images~\cite{FidlerBottomCVPR2013}. Note that ~\cite{FidlerBottomCVPR2013}'s features are a special case of these features with a grid size $K=1$. Overlap and 
segment features can also be quickly computed using matrix operations. 

{\bf Context:} 
CNNs are typically trained for the task of  image classification where in most cases an 
input image is much larger than the object. This means that part of their success may be due to learning complex dependencies between the objects and their  contextual information (\eg sky for aeroplane, road for car and bus).
However, the appearance features that we use are only computed based on the candidate's box, thus hardly capturing useful information from the scene. We thus add an additional feature that looks at a bigger scope than the candidate's box.

In particular, we enlarge each input candidate box by a fixed percentage $\rho$ along its horizontal and vertical direction. For big boxes, or those close to the image boundary, we clip the enlarged region to be fully inside the image. We keep the object labels for each expanded box the same as that for the original boxes, even if the expanded box now encloses objects of other classes.  We then warp the image in each enlarged box to $227\times227$ and fine-tune the original ImageNet-trained CNN using these images and labels. We call the fine-tuned network the \textit{expanded CNN}. For our contextual features  $\phi_{ctx}(x,p)$ we extract the $fc7$ layer features of the expanded CNN by running the warped image in the enlarged window through the network.

\subsection{Inference}
In the inference stage of our model, we score each candidate box $p$ as follows:
\begin{align} 
f_{\boldsymbol w}(x,p)&=\max_{{\bf h}}\big( \omega_{app}^T \cdot \phi_{app}(x,p) + 
\omega_{ctx}^T \cdot \phi_{ctx}(x,p) \nonumber \\ 
&+ \omega_{seg}^T \cdot \phi_{seg}(x,p,{\bf h})\big).
\label{eqn:inference}
\end{align}
Observe that the first two terms in Eq.~\ref{eqn:inference} can be computed efficiently by matrix multiplication, and the only part that depends on ${\bf h}$ is its last term. Although there could be exponential number of candidates for ${\bf h}$, we can greedily search each dimension of  ${\bf h}$ and find the best segment $S(h_c)$ w.r.t. model parameters $\omega_{seg}$ for each class $c$. Since our segmentation features do not depend on the pairwise relationships in ${\bf h}$, this greedy approach is guaranteed to find the global maximum of $\omega_{seg}^T \cdot \phi_{seg}(x,p,{\bf h})$. Finally, we sum the three terms to obtain the score of each bounding box location $p$.

\begin{table*}\scriptsize
\begin{center}
\addtolength{\tabcolsep}{-0.1pt}
\begin{tabular}{|p{1.4cm}|p{0.06cm}p{0.06cm}p{0.06cm}p{0.06cm}|p{0.24cm}p{0.24cm}p{0.24cm}p{0.24cm}p{0.24cm}p{0.24cm}p{0.24cm}p{0.24cm}p{0.24cm}p{0.24cm}p{0.24cm}p{0.24cm}p{0.24cm}p{0.24cm}p{0.28cm}p{0.24cm}p{0.24cm}p{0.24cm}p{0.24cm}p{0.26cm}|p{0.32cm}|}
\hline
 & seg & exp & ibr & br & plane & bike  & bird  & boat  & bottle & bus   & car   & cat   & chair & cow   & table & dog   & horse & motor & person & plant & sheep & sofa  & train & tv    & mAP     \\
\hline\hline
RCNN      &  &  &    &   & 69.9  & 64.2 & 48.0 & 30.2 & 26.9   & 63.3 & 56.0 & 67.6 & 26.8  & 44.7 & 29.6  & 61.7 & 55.7  & 69.8  & 56.4   & 26.6  & 56.7  & 35.6 & 54.4  & 57.7 & 50.1 \\
RCNN+CPMC &   &   &   &   & 71.5 & 65.3 & 48.6 & 31.5 & 27.9 & 64.3 & 57.2 & 67.6 & 26.7 & 46.2 & 33.6 & 62.8 & 57.8 & 70.7 & 57.9 & 26.6 & 54.0 & 37.8 & 57.0 & 57.6 & 51.1\\
segDPM+CNN  & $\surd$   &    &      &      & 72.8 & 64.1 & 50.7 & 32.1 & 28.2 & 64.9 & 55.9 & 72.4 & 27.7 & 50.6 & 31.7 & 65.9 & 59.3 & 71.1 & 57.1 & 26.5 & 59.4 & 38.8 & 57.1 & 57.6 & 52.2\\
segDeepM  & $\surd$   &    &      &      & 73.8 & 64.0 & 52.4 & 32.7 & 28.2 & 66.4 & 56.7 & 73.1 & 28.1 & 51.4 & 34.0 & 66.1 & 59.9 & 71.0 & 56.6 & 29.5 & 59.5 & 43.9 & 61.6 & 58.0 & 53.3 \\
segDeepM  &    & $\surd$   &      &   & 72.2  & 65.2 & 52.4 & 36.3 & 29.4   & 67.3 & 59.0 & 71.0 & 28.9  & 49.1 & 30.6  & 67.6 & 59.3  & 72.6  & 59.1   & 28.7  & 60.6  & 38.6 & 58.2  & 60.3 & 53.3 \\
segDeepM  &   &   & $\surd$     &  & 71.4  & 64.3 & 50.2 & 31.8 & 30.6   & 66.0 & 57.5 & 68.7 & 25.6  & 49.7 & 30.5  & 64.7 & 58.3  & 69.9  & 60.7   & 26.9  & 54.4  & 35.0 & 57.1  & 55.5 & 51.4 \\
segDeepM  & $\surd$   & $\surd$   &      &   & 74.5  & 64.8 & 55.3 & 36.3 & 31.2   & 69.0 & 59.0 & 73.8 & 29.7  & 53.3 & 33.7  & 68.8 & 62.3  & 73.1  & 59.3   & 29.8  & 63.1  & 41.3 & 63.4  & 60.0 & 55.1 \\
segDeepM  & $\surd$   & $\surd$   & $\surd$     &  & 77.1  & 67.4 & 58.2 & 36.9 & 37.4   & 71.3 & 61.1 & 74.4 & 29.3  & 56.3 & 34.8  & 69.8 & 64.1  & 72.7  & 64.1   & 31.5  & 60.0  & 39.7 & 64.8  & 58.2 & 56.4 \\
\hline
RCNN      &  &  &    & $\surd$     & 72.9  & 65.8 & 54.0 & 34.5 & 31.2   & 68.0 & 59.8 & 72.3 & 26.6  & 51.3 & 35.0  & 65.7 & 59.7  & 71.7  & 60.7   & 28.0  & 60.6  & 37.1 & 60.3  & 59.9 & 53.7 \\
segDeepM & $\surd$ &  &  & $\surd$ & 76.9 & 66.8 & 57.8 & 36.2 & 32.2 & 71.4 & 60.0 & 75.4 & 27.7 & 53.8 & 38.6 & 68.6 & 64.4 & 72.6 & 61.1 & 30.4 & 61.5 & 43.2 & 64.1 & 60.9 & 56.2 \\
segDeepM &  & $\surd$ &  & $\surd$ & 76.7 & 68.7 & 58.0 & 39.9 & 34.6 & 71.1 & 62.1 & \textbf{75.9} & 30.3 & 54.6 & 36.2 & 69.6 & 63.3 & 74.0 & 63.5 & 31.3 & 62.5 & 37.9 & 66.3 & 61.0 & 56.9 \\
segDeepM &  &  & $\surd$ & $\surd$ & 77.2 & 66.6 & 55.2 & 34.5 & 34.6 & 67.5 & 60.0 & 70.4 & 27.1 & 53.4 & 35.9 & 66.4 & 63.4 & 71.9 & 63.0 & 32.0 & 55.7 & 38.5 & 62.0 & 58.0 & 54.7 \\
segDeepM & $\surd$ & $\surd$ &  & $\surd$ & 77.2 & 67.6 & 59.8 & 40.2 & 35.7 & \textbf{72.0} & \textbf{62.1} & 75.7 & 30.4 & \textbf{58.1} & 37.2 & 69.9 & 64.8 & \textbf{73.9} & 63.4 & 32.4 & \textbf{63.9} & 43.1 & \textbf{68.4} & \textbf{61.6} & 57.9\\
segDeepM  & $\surd$   & $\surd$   & $\surd$     & $\surd$     & \textbf{79.0}  & \textbf{70.6} & \textbf{61.9} & \textbf{40.4} & \textbf{39.0}  & 71.6 & 61.9 & 74.7 & \textbf{31.3}  & 56.6 & \textbf{39.2}  & \textbf{70.4} & \textbf{66.5}  & 73.5  & \textbf{65.6}  & \textbf{35.3}  & 60.7  & \textbf{44.3} & 68.0  & 58.7 & \textbf{58.5}  \\
\hline
\end{tabular}
\end{center}
\vspace{-5mm}
\caption{Detection results (in \% AP) on PASCAL VOC 2010 $val$ for R-CNN and segDeepM detectors.}
\label{table:val}.
\vspace{-3mm}
\end{table*}

\subsection{Learning}

Given a set of images with $N$ candidate boxes $\{p_n\}$ and their annotations $\{y(x_n,p_n)\}$, together with a collection of segments for each image $\{S(x_n,h_n)\}$ and associated potentials $\{\phi(x_n,h_n)\}$ with $n=1,..,N$, training our model can be written as follows:
\begin{equation}
\label{eqn:latentSVM}
	\min_{\boldsymbol{\omega}} \lVert {\boldsymbol \omega}\rVert^2 + C\sum_{n=1}^N \max\left(0,1-y(x_n,p_n)f_{\boldsymbol w}(x_n,p_n)\right),
\end{equation}
where $\boldsymbol\omega$ is a vector of all the weights in our model.

The learning problem in~\eqref{eqn:latentSVM} is a latent SVM~\cite{felzenszwalb2010object} where we treat the assignment variable ${\bf h}$ as a latent variable for each training instance. To optimize Equation~\eqref{eqn:latentSVM},  we iterate two steps following~\cite{felzenszwalb2010object}:
\begin{enumerate}
\item \textit{label each positive example:} for each $(x,p)$ with $y(x,p)=1$, we compute ${\bf h^*}=\arg\max_{{\bf h}}f_{\boldsymbol w}(x,p)$ with current the model parameters $\boldsymbol\omega$;
\item \textit{update the weights:}  we do hard-negative mining over a set of negative instances until reaching a certain memory limit. We then use stochastic gradient descent to optimize  the weights $\boldsymbol \omega$.
\end{enumerate}

Latent SVM is guaranteed to converge to a local minimum only, thus we need to carefully pick a good initialization for positive examples at the first iteration. We use the overlap feature $\phi_{overlap}$ as the indicator and set each dimension of ${\bf h^*}$ as ${\bf h}^*=\arg\max_{\bf h} \phi_{overlap}(p,{\bf h})$. This encourages the method to pick segments that best overlaps with the candidate's box.

Although our segmentation features are efficient to compute, we need to recompute them for all positive examples during the first step and for all hard negative examples during the second step of training. In our implementation, we cache a pool of segmentation features $\phi_{seg}(x,p,{\bf h})$ for all training instances to avoid computing them in every iteration. With the compact segmentation feature, our method achieves similar running speed with that of R-CNN~\cite{girshick2013rich}.

\subsection{Iterative Bounding Box Prediction}
\label{sec:bboxpred}

As a typical {\bf postprocessing} step, object detection approaches usually perform bounding box prediction on their final candidate set~\cite{felzenszwalb2010object,girshick2013rich}. This typically results in a few percent improvement in accuracy, since the approach is able to make an informative re-localization based on complex features. Following~\cite{girshick2013rich}, we also use $pool5$ layer features in order to do bounding box prediction.

In this paper we take this idea one step further by doing bounding box prediction and scoring the model in an iterative fashion. 
Our motivation is that better localization can lead to improved
predictions.   
In particular, we first extract the CNN features and regress to a corrected set of boxes. We then re-extract the features on the new boxes and score our model. We only re-extract the features on the boxes which have changed by more than 20\% percent from the original set. We can then repeat this process, by doing bounding box prediction again, re-extracting the features, and re-scoring. Our experiments show that this procedure converges to a set of stable boxes after two iterations.


\vspace{-1mm}
\begin{table*}\scriptsize
\begin{center}
\begin{tabular}{|p{0.9cm}|p{0.2cm}p{0.2cm}p{0.2cm}p{0.2cm}|p{0.24cm}p{0.24cm}p{0.24cm}p{0.24cm}p{0.24cm}p{0.24cm}p{0.24cm}p{0.24cm}p{0.24cm}p{0.24cm}p{0.24cm}p{0.24cm}p{0.24cm}p{0.24cm}p{0.24cm}p{0.24cm}p{0.24cm}p{0.24cm}p{0.24cm}p{0.27cm}|p{0.32cm}|}
\hline
 & seg & exp & ibr & br & plane & bike  & bird  & boat  & bottle & bus   & car   & cat   & chair & cow   & table & dog   & horse & motor & person & plant & sheep & sofa  & train & tv    & mAP     \\
\hline\hline
RCNN     &   &   &   &   & 74.4 & 69.0 & 55.6 & 34.5 & 35.2 & 70.8 & 63.0 & 81.4 & 35.0 & 57.9 & 39.3 & 77.7 & 70.2 & 75.8 & 61.7 & 29.1 & 66.9 & 56.7 & 63.8 & 58.7 & 58.8 \\
segDeepM & $\surd$ &   &   &   & 77.5 & 67.7 & 59.0 & 33.4 & 35.9 & 71.3 & 62.4 & 82.8 & 35.6 & 61.4 & 42.7 & 79.1 & 70.7 & 76.7 & 62.3 & 31.3 & 67.1 & 54.0 & 65.5 & 60.0 & 59.8 \\
segDeepM &   & $\surd$ &   &   & 77.4 & 72.9 & 62.6 & 36.8 & 39.4 & 71.5 & 64.9 & 83.7 & 37.1 & 62.0 & 40.4 & 81.0 & 73.1 & 77.9 & 65.7 & 34.7 & 68.0 & 59.1 & 70.0 & 58.7 & 61.8 \\
segDeepM & $\surd$ & $\surd$ &   &   & 79.2 & 69.8 & 63.6 & 36.4 & 39.5 & 72.9 & 65.2 & 83.5 & 38.4 & 63.6 & 43.8 & 80.8 & 75.1 & 78.3 & 66.2 & 33.3 & 68.6 & 56.0 & 70.7 & 60.2 & 62.2 \\
RCNN     &   &   &  & $\surd$ & 78.6 & 72.1 & 62.1 & 40.4 & 40.0 & 71.2 & 65.0 & 84.2 & 36.7 & 59.5 & 41.8 & 80.3 & 74.5 & 78.0 & 65.8 & 33.0 & 67.3 & 59.9 & 68.7 & 61.3 & 62.0 \\
segDeepM & $\surd$ &  &  & $\surd$ & 79.0 & 71.3 & 63.0 & 38.9 & 40.0 & 72.8 & 63.4 & 84.9 & 36.7 & 62.1 & 44.3 & 80.2 & 76.0 & 78.7 & 66.0 & 35.8 & 68.3 & 57.8 & 67.5 & 62.0 & 62.4 \\
segDeepM &  & $\surd$ &  & $\surd$ & \textbf{81.6 }& \textbf{74.6} & \textbf{66.6} & \textbf{41.6} & 44.6 & 70.7 & \textbf{68.0} & 84.9 & 39.7 & 62.5 & 44.2 & \textbf{84.1} & \textbf{77.1} & 79.2 & 69.9 & 35.7 & 67.6 & \textbf{60.9} & \textbf{72.7} & \textbf{61.3} & 64.4 \\
segDeepM & $\surd$ & $\surd$ &  & $\surd$ & 81.5 & 73.4 & \textbf{66.6} & 40.4 & \textbf{44.7} & \textbf{71.3} & 67.5 & \textbf{85.1} & \textbf{40.9} & \textbf{62.9} & \textbf{46.0} & 83.5 & 76.9 & \textbf{80.0} & \textbf{70.0} & \textbf{37.1} & \textbf{68.9} & 60.4 & 72.2 & 61.1 & \textbf{64.5} \\
\hline
\end{tabular}
\end{center}
\vspace{-5.2mm}
\caption{Detection results (in \% AP) on PASCAL VOC 2010 $val$ for RCNN and segDeepM using 16 layer OxfordNet CNN.}
\label{table:val_16}.

\begin{center}
\vspace{-5mm}
\begin{tabular}{|p{2.2cm}|p{0.305cm}p{0.305cm}p{0.305cm}p{0.305cm}p{0.305cm}p{0.305cm}p{0.305cm}p{0.305cm}p{0.305cm}p{0.305cm}p{0.305cm}p{0.305cm}p{0.305cm}p{0.305cm}p{0.305cm}p{0.305cm}p{0.305cm}p{0.305cm}p{0.305cm}p{0.305cm}|p{0.35cm}|}
\hline
                & plane & bike & bird & boat & bottle & bus  & car  & cat  & chair & cow  & table & dog  & horse & motor & person & plant & sheep & sofa & train & tv   & mAP      \\
\hline\hline
segDeepM-16 layers    &   \textbf{82.3}   & \textbf{75.2} & \textbf{67.1} & \textbf{50.7} & \textbf{49.8} & \textbf{71.1} & \textbf{69.5} & \textbf{88.2} & \textbf{42.5} & \textbf{71.2} & \textbf{50.0} & \textbf{85.7} & \textbf{76.6} & \textbf{81.8} & \textbf{69.3} & \textbf{41.5} & \textbf{71.9} & \textbf{62.2} & \textbf{73.2} & \textbf{64.6} & \textbf{67.2} \\
segDeepM-8 layers              & 75.3 & 69.7 & 57.6   & 44.2 & 42.1 & 62.2 & 64.7  & 74.8 & 30.1  & 55.6 & 43.1  & 70.7  & 66.4   & 72.6  & 63.5  & 31.9 & 61.9  & 46.1 & 64.4 & 58.1 & 57.8 \\
\hline
BabyLearning        & 77.7 & {73.8} & 62.3 & {48.8} & 45.4 & 67.3 & 67.0   & 80.3 & {41.3} & {70.8} & {49.7} & 79.5 & 74.7 & {78.6} & 64.5 & 36.0   & {69.9} & 55.7 & 70.4 & {61.7} & {63.8} \\
R-CNN (breg)-16 ly.   & {79.3} & 72.4 & 63.1 & 44.0   & 44.4 & 64.6 & 66.3 & {84.9} & 38.8 & 67.3 & 48.4 & {82.3}& {75.0} & 76.7 & 65.7 & 35.8 & 66.2 & 54.8 & 69.1 & 58.8 & 62.9 \\
R-CNN-16 layers               & 76.5 & 70.4 & 58.0   & 40.2 & 39.6 & 61.8 & 63.7 & 81.0   & 36.2 & 64.5 & 45.7 & 80.5 & 71.9 & 74.3 & 60.6 & 31.5 & 64.7 & 52.5 & 64.6 & 57.2 & 59.8 \\
Feature Edit        & 74.8 & 69.2 & 55.7 & 41.9 & 36.1 & 64.7 & 62.3 & 69.5 & 31.3 & 53.3 & 43.7 & 69.9 & 64.0   & 71.8 & 60.5 & 32.7 & 63.0   & 44.1 & 63.6 & 56.6 & 56.4 \\
R-CNN (breg)    & 71.8 & 65.8 & 53.0   & 36.8 & 35.9 & 59.7 & 60.0   & 69.9 & 27.9 & 50.6 & 41.4 & 70.0   & 62.0   & 69.0   & 58.1 & 29.5 & 59.4 & 39.3 & 61.2 & 52.4 & 53.7 \\
R-CNN               & 67.1 & 64.1 & 46.7 & 32.0   & 30.5 & 56.4 & 57.2 & 65.9 & 27.0   & 47.3 & 40.9 & 66.6 & 57.8 & 65.9 & 53.6 & 26.7 & 56.5 & 38.1 & 52.8 & 50.2 & 50.2 \\
\hline   
\end{tabular}
\end{center}
\vspace{-5.2mm}
\caption{State-of-the-art detection results (in \%AP) on PASCAL VOC 2010 $test$. The 16 layer models adopt OxforNet, the rest use 8-layer AlexNet.}
\label{table:test}.

\begin{center}
\vspace{-5mm}
\begin{tabular}{|p{1cm}|p{0.18cm}p{0.18cm}p{0.18cm}p{0.18cm}|p{0.24cm}p{0.24cm}p{0.24cm}p{0.24cm}p{0.24cm}p{0.24cm}p{0.24cm}p{0.24cm}p{0.24cm}p{0.24cm}p{0.24cm}p{0.24cm}p{0.24cm}p{0.24cm}p{0.24cm}p{0.24cm}p{0.24cm}p{0.24cm}p{0.24cm}p{0.27cm}|p{0.32cm}|}
\hline
 & seg & exp & ibr & br & plane & bike  & bird  & boat  & bottle & bus   & car   & cat   & chair & cow   & table & dog   & horse & motor & person & plant & sheep & sofa  & train & tv    & mAP     \\
 \hline\hline
RCNN     &  &  &  &  & 68.9 & 63.5 & 45.6 & 29.1 & 26.7 & 64.4 & 55.6 & 69.5 & 26.3 & 50.3 & 36.1 & 62.2 & 55.6 & 68.7 & 56.0 & 27.6 & 54.8 & 40.2 & 54.0 & 60.5 & 50.8 \\
segDeepM & $\surd$ & $\surd$& $\surd$ &  & 75.1 & 67.6 & 56.6 & 37.9 & 34.6 & 73.8 & 60.8 & 76.1 & 27.6 & 55.4 & 39.1 & 68.9 & 63.7 & \textbf{72.4} & 63.8 & 33.5 & 60.8 & 45.9 & 65.7 & 60.9 & 57.0 \\
\hline
RCNN     &  &  &  & $\surd$ & 74.0 & 66.7 & 50.4 & 32.9 & 31.5 & 68.0 & 58.4 & 74.7 & 26.9 & 52.9 & 39.3 & 67.0 & 59.2 & 71.5 & 59.1 & 29.9 & 56.8 & 44.2 & 63.1 & 63.7 & 54.5 \\
segDeepM & $\surd$ & $\surd$ & $\surd$ & $\surd$ & \textbf{77.3} & \textbf{70.2} &\textbf{60.2} & \textbf{39.8 }& \textbf{38.3} & \textbf{75.2} & \textbf{62.3} & \textbf{76.1} & \textbf{29.4} & \textbf{55.7} & \textbf{40.8} & \textbf{70.5} & \textbf{66.4} & \textbf{72.4} & \textbf{65.8} & \textbf{35.6} & \textbf{60.9} & \textbf{46.1} & \textbf{66.7} & \textbf{63.5} & \textbf{58.7}\\
\hline
\end{tabular}
\end{center}
\vspace{-5mm}
\caption{Detection results (in \% AP) on PASCAL VOC 2012 $val$ for RCNN and segDeepM detectors. }
\label{table:val2012}.
\vspace{-2mm}
\end{table*}

\begin{figure*}[htb]
\vspace{-5mm}
	\centering
	\begin{subfigure}[b]{0.18\textwidth}
		\includegraphics[width=1\textwidth]{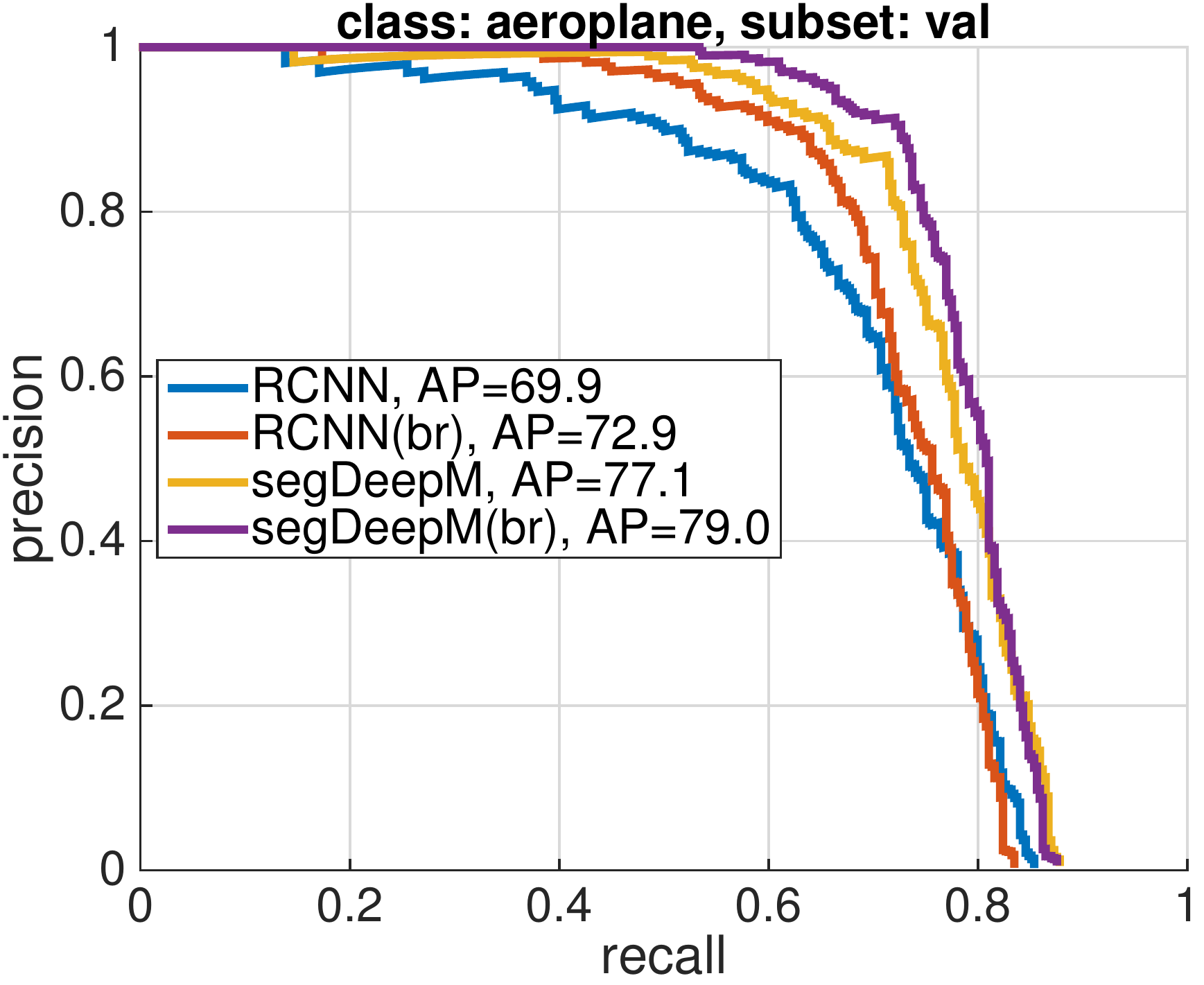}
		\caption{PR curve for plane}
	\end{subfigure}		
	\begin{subfigure}[b]{0.18\textwidth}
		\includegraphics[width=1\textwidth]{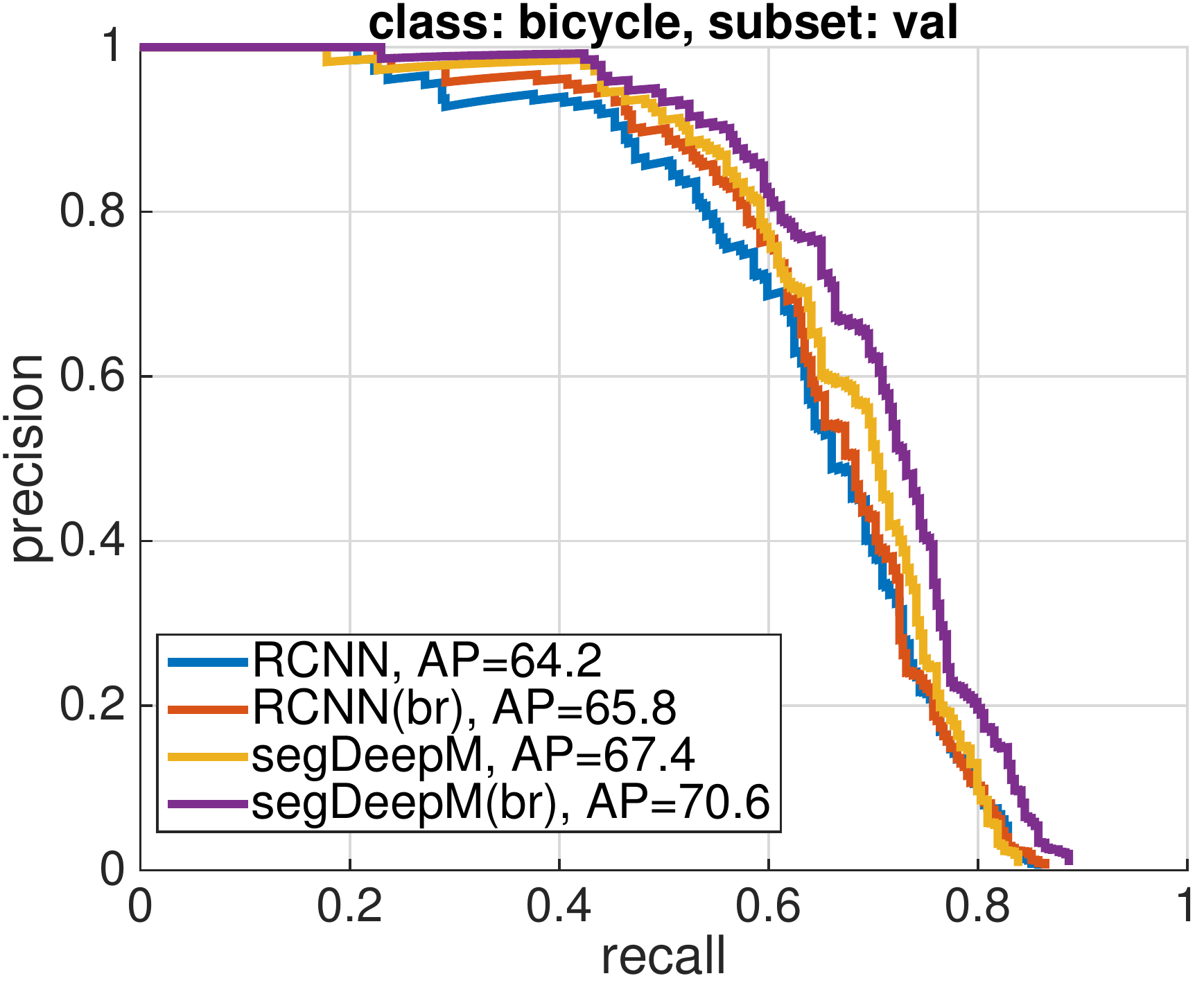}
		\caption{PR curve for bicycle}
	\end{subfigure}		
	\begin{subfigure}[b]{0.18\textwidth}
		\includegraphics[width=1\textwidth]{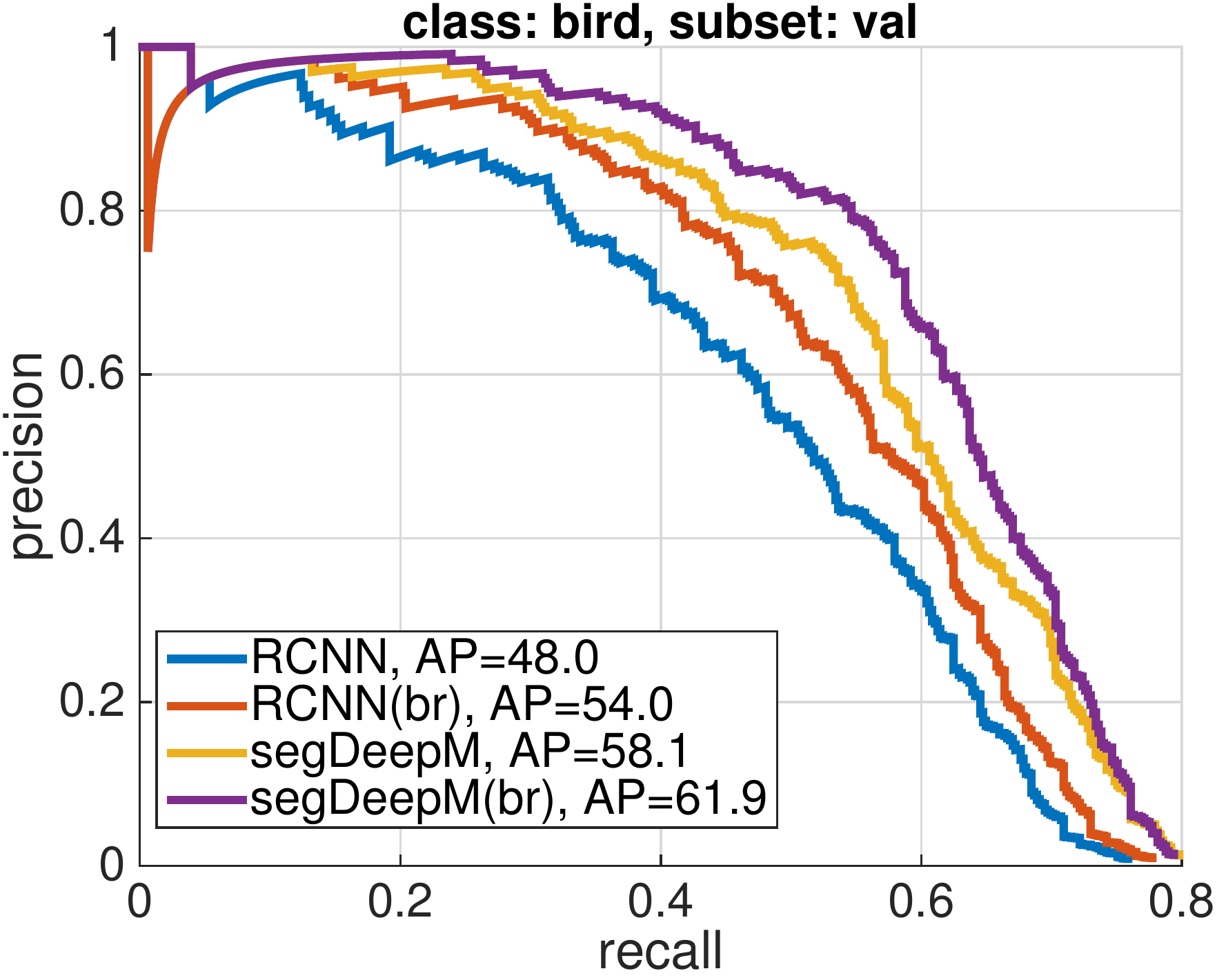}
		\caption{PR curve for bird}
	\end{subfigure}		
	\begin{subfigure}[b]{0.18\textwidth}
		\includegraphics[width=1\textwidth]{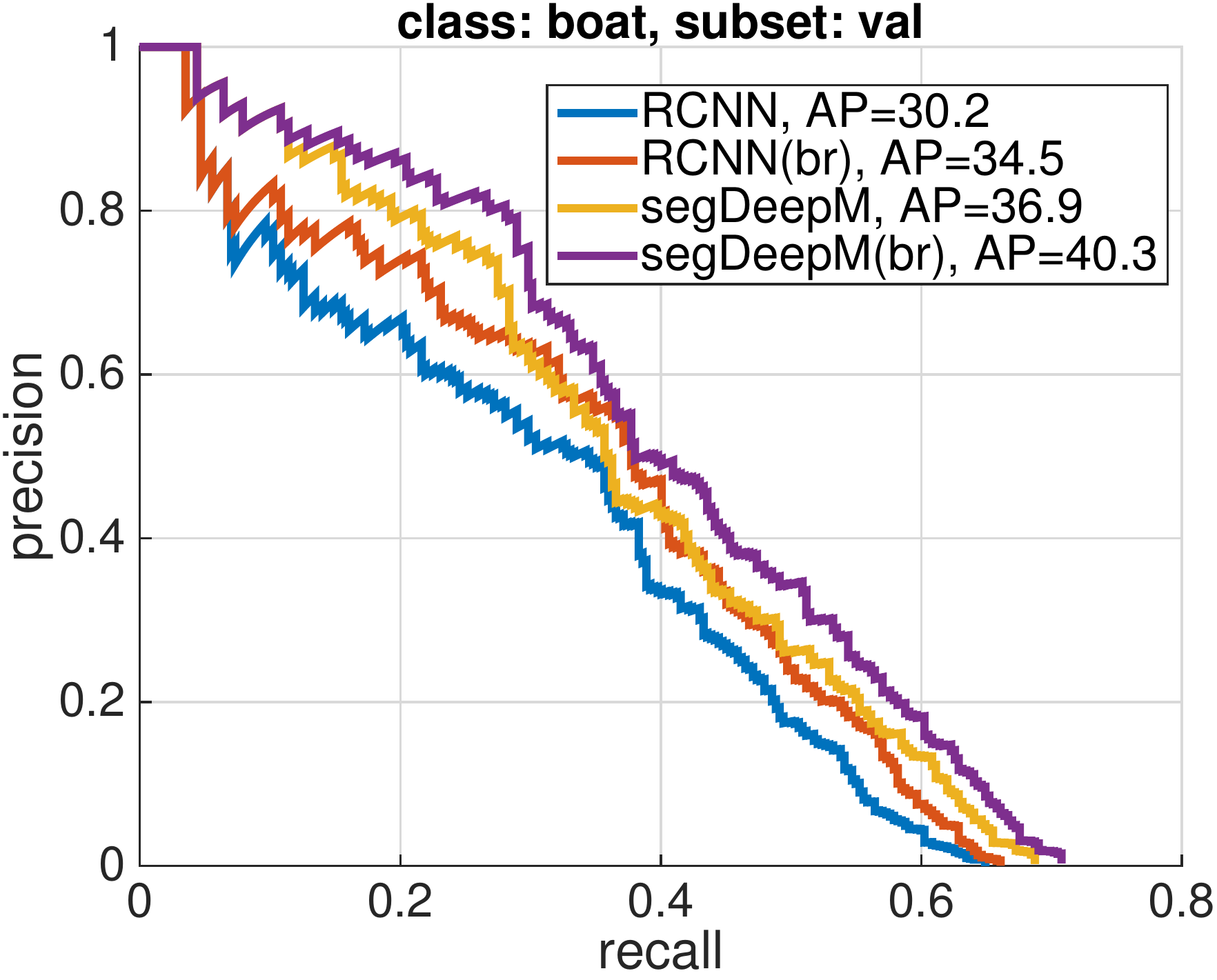}
		\caption{PR curve for boat}
	\end{subfigure}		
	\begin{subfigure}[b]{0.18\textwidth}
		\includegraphics[width=1\textwidth]{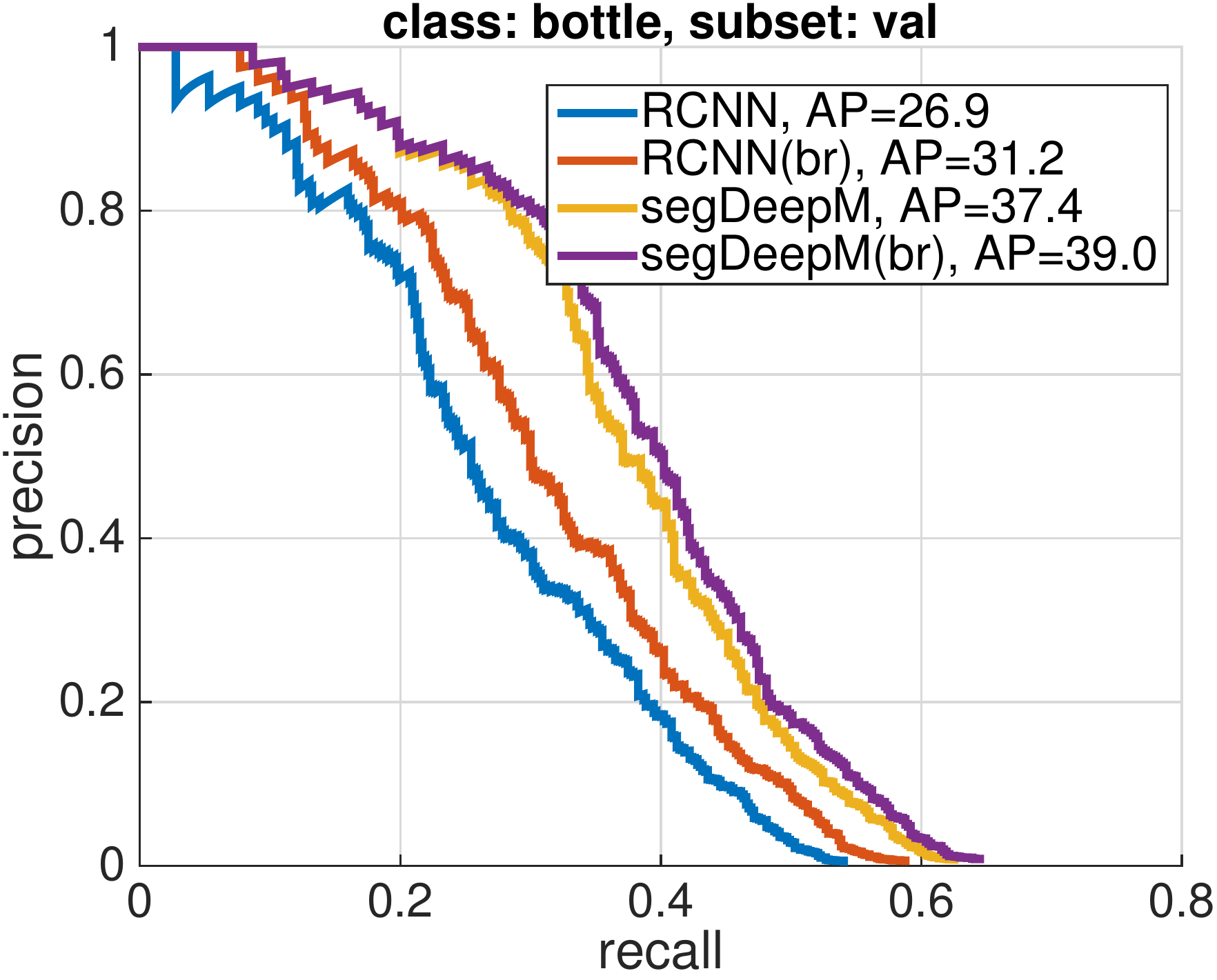}
		\caption{PR curve for bottle}
	\end{subfigure}			

	\begin{subfigure}[b]{0.18\textwidth}
		\includegraphics[width=1\textwidth]{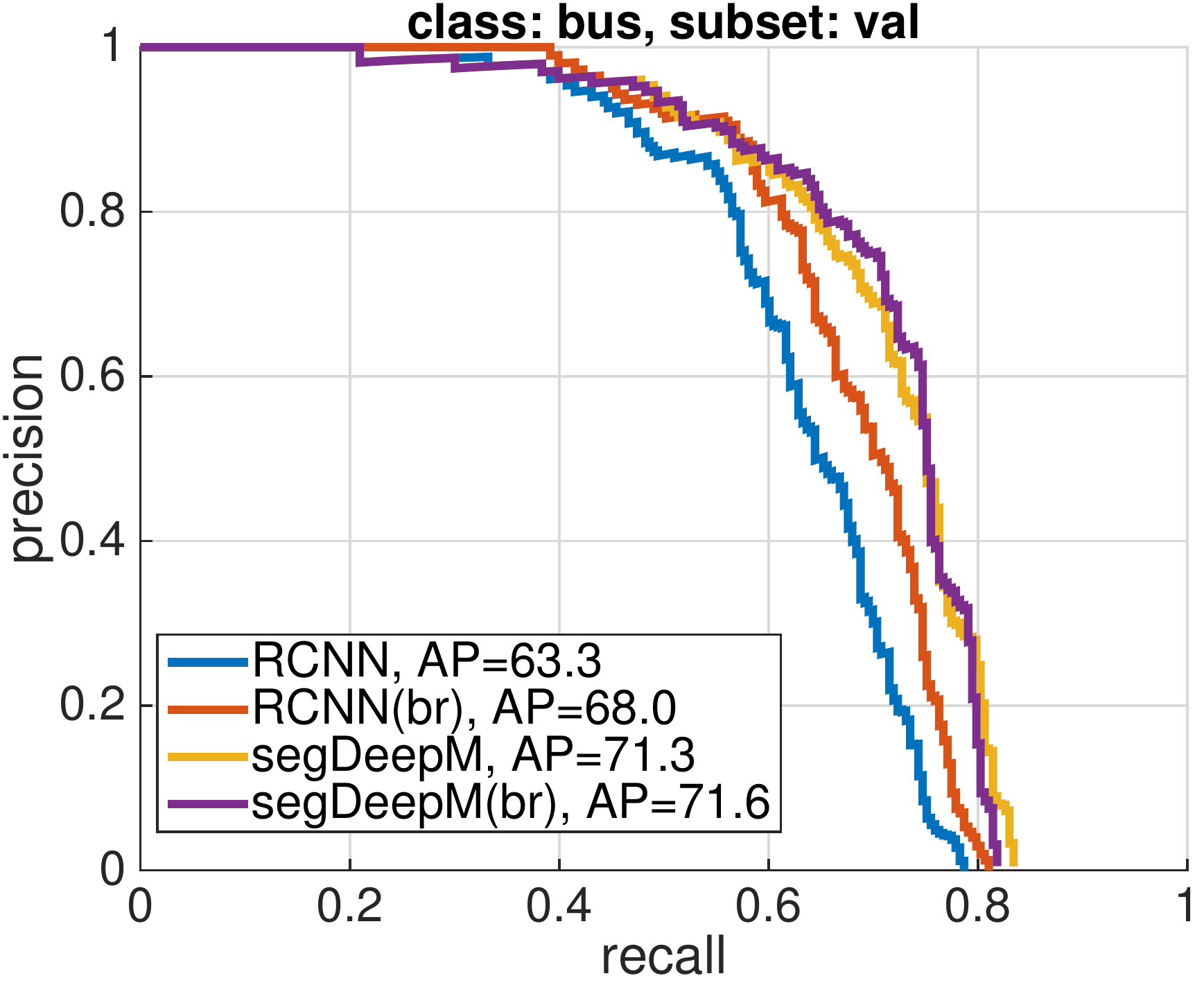}
		\caption{PR curve for bus}
	\end{subfigure}		
	\begin{subfigure}[b]{0.18\textwidth}
		\includegraphics[width=1\textwidth]{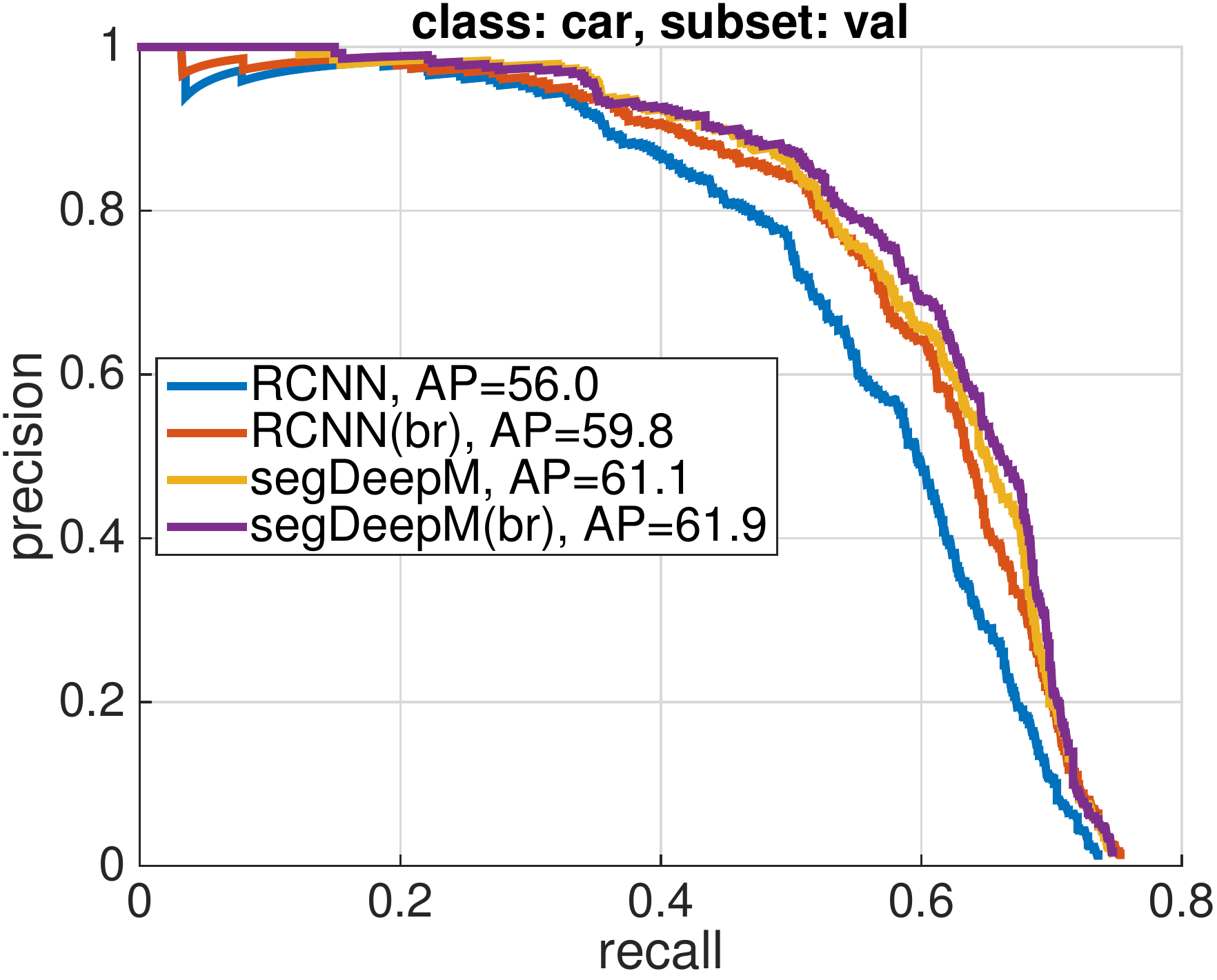}
		\caption{PR curve for car}
	\end{subfigure}		
	\begin{subfigure}[b]{0.18\textwidth}
		\includegraphics[width=1\textwidth]{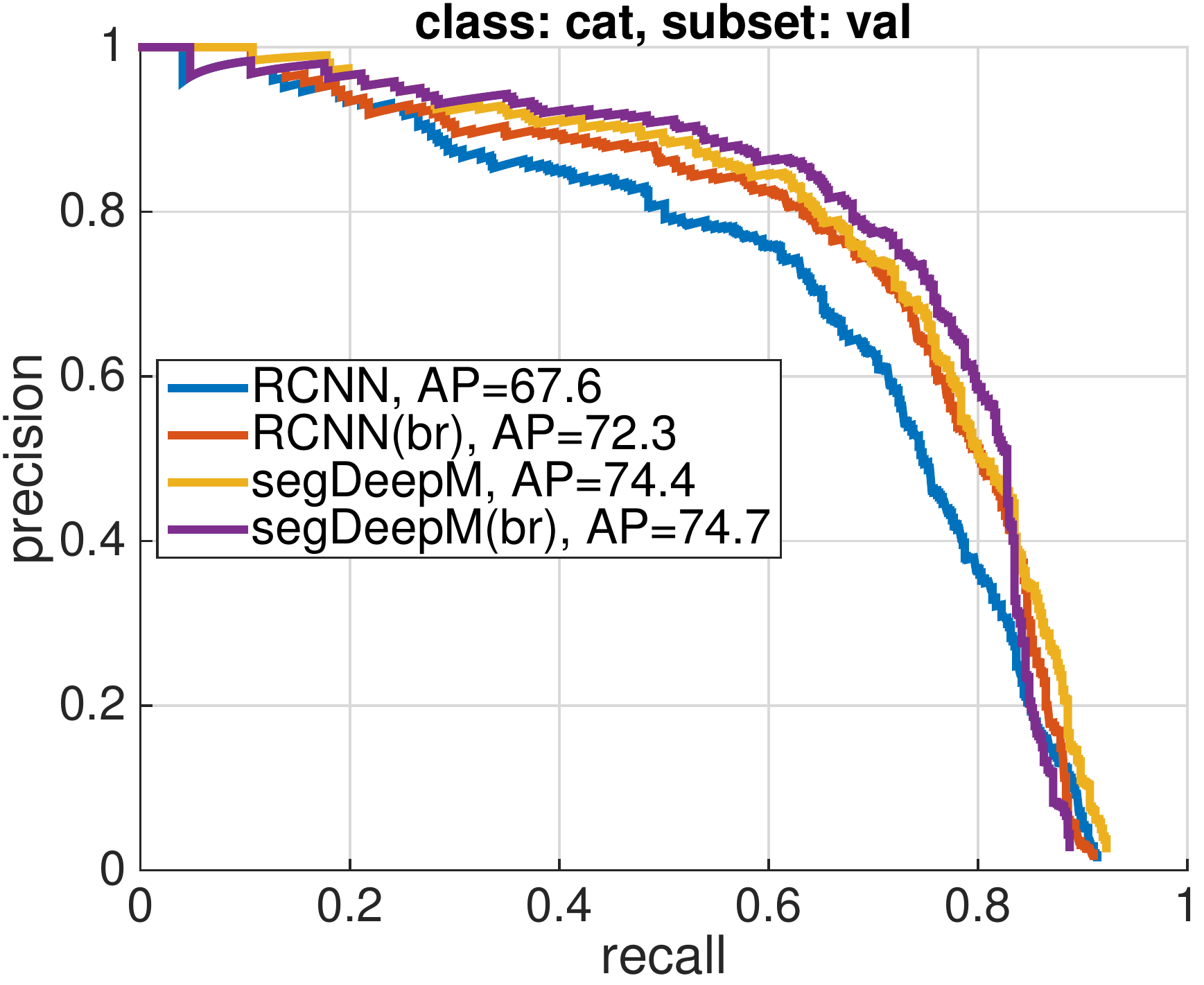}
		\caption{PR curve for cat}
	\end{subfigure}		
	\begin{subfigure}[b]{0.18\textwidth}
		\includegraphics[width=1\textwidth]{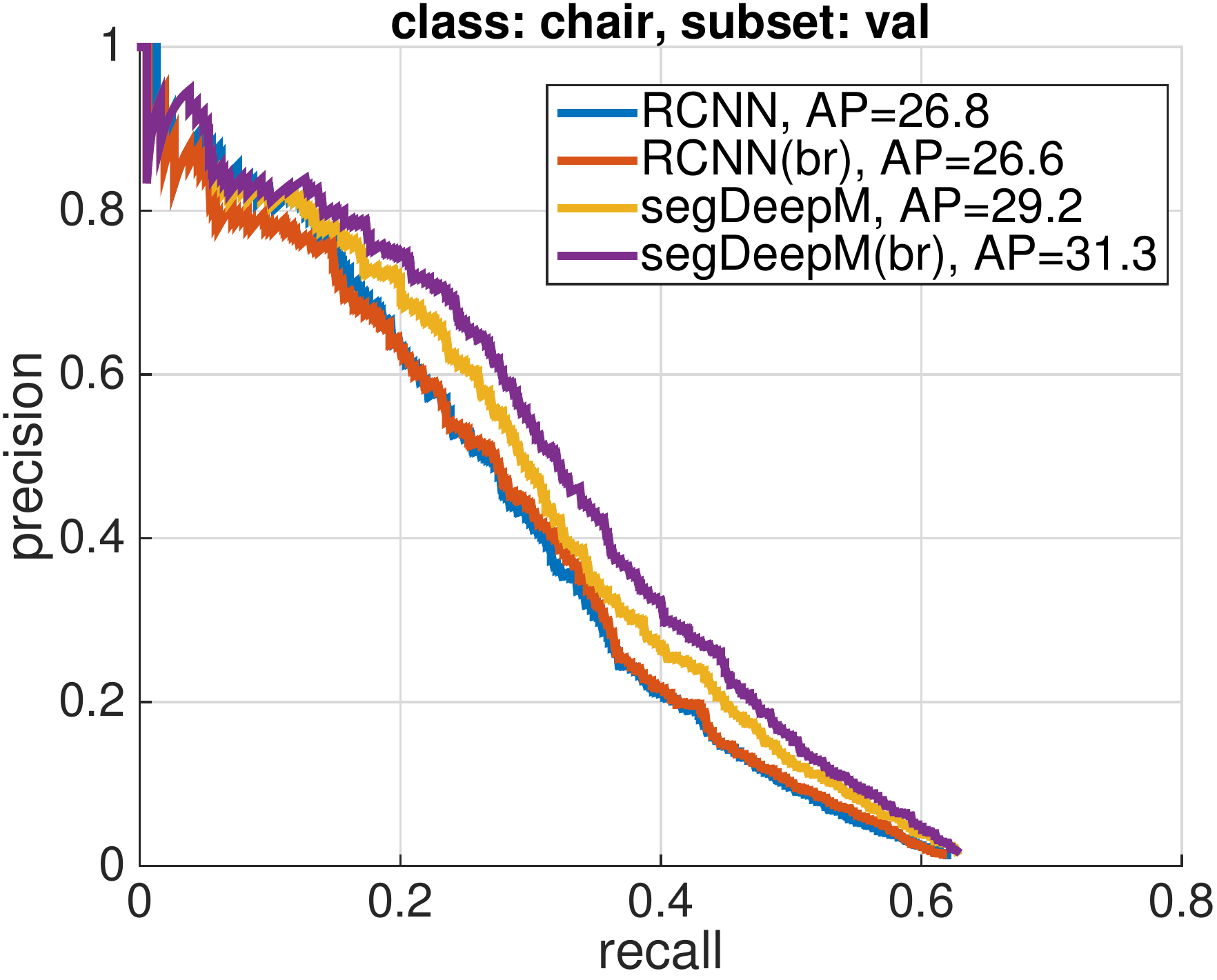}
		\caption{PR curve for chair}
	\end{subfigure}		
	\begin{subfigure}[b]{0.18\textwidth}
		\includegraphics[width=1\textwidth]{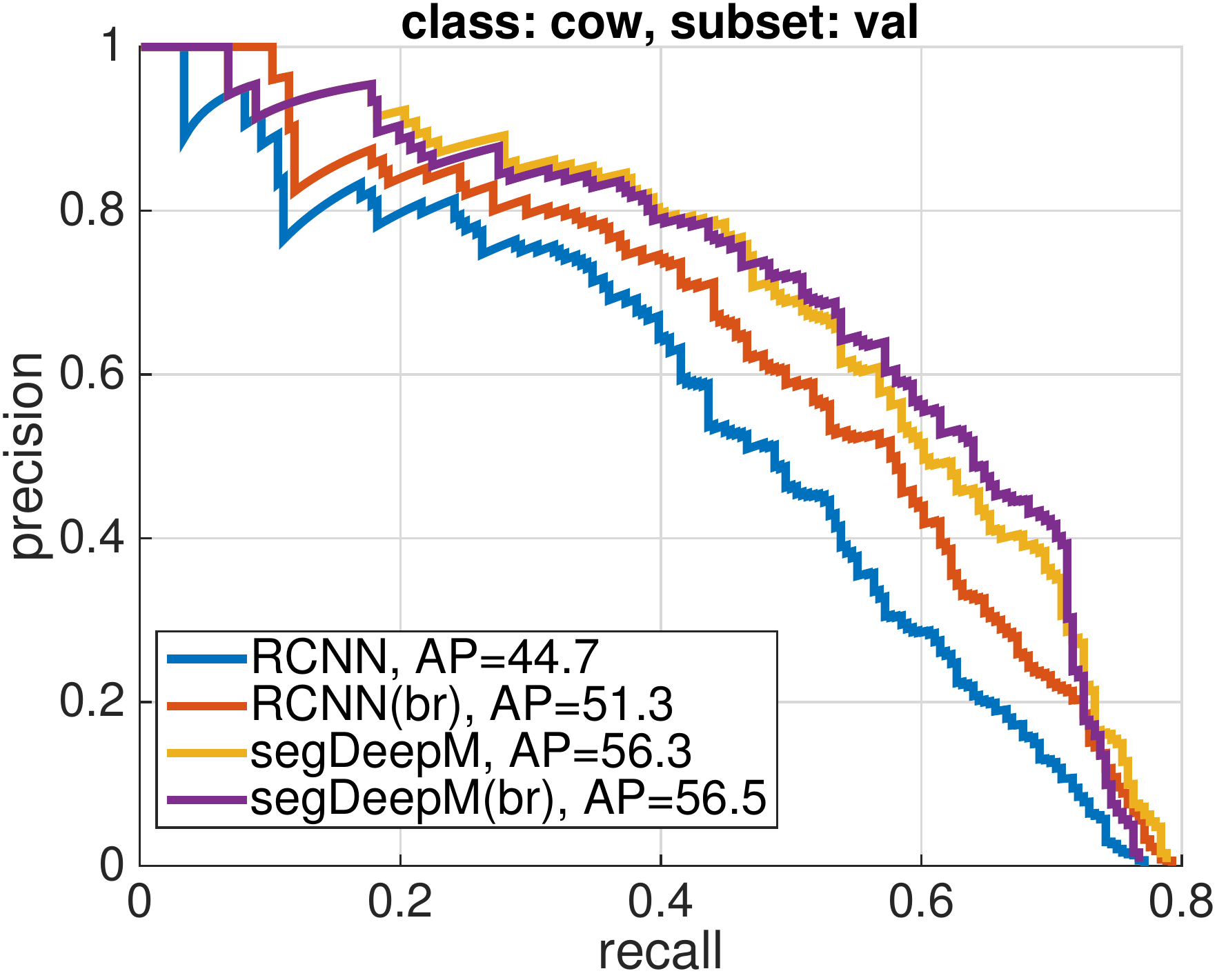}
		\caption{PR curve for cow}
	\end{subfigure}	

	\vspace{-2.5mm}	
	\caption{PR curves on PASCAL VOC 2010 $train$/$val$ split for 10 classes. All plots are in Suppl. material.}\label{figure:pr}
\end{figure*}

\begin{figure*}[t!]
\vspace{-3.2mm}
\addtolength{\tabcolsep}{-0.2pt}
	\begin{tabular}{p{1.8mm}cccccc}
		\rotatebox{90}{\hspace{7.5mm}RCNN} & \includegraphics[width=0.14\linewidth,trim=8 0 38 0,clip=true]{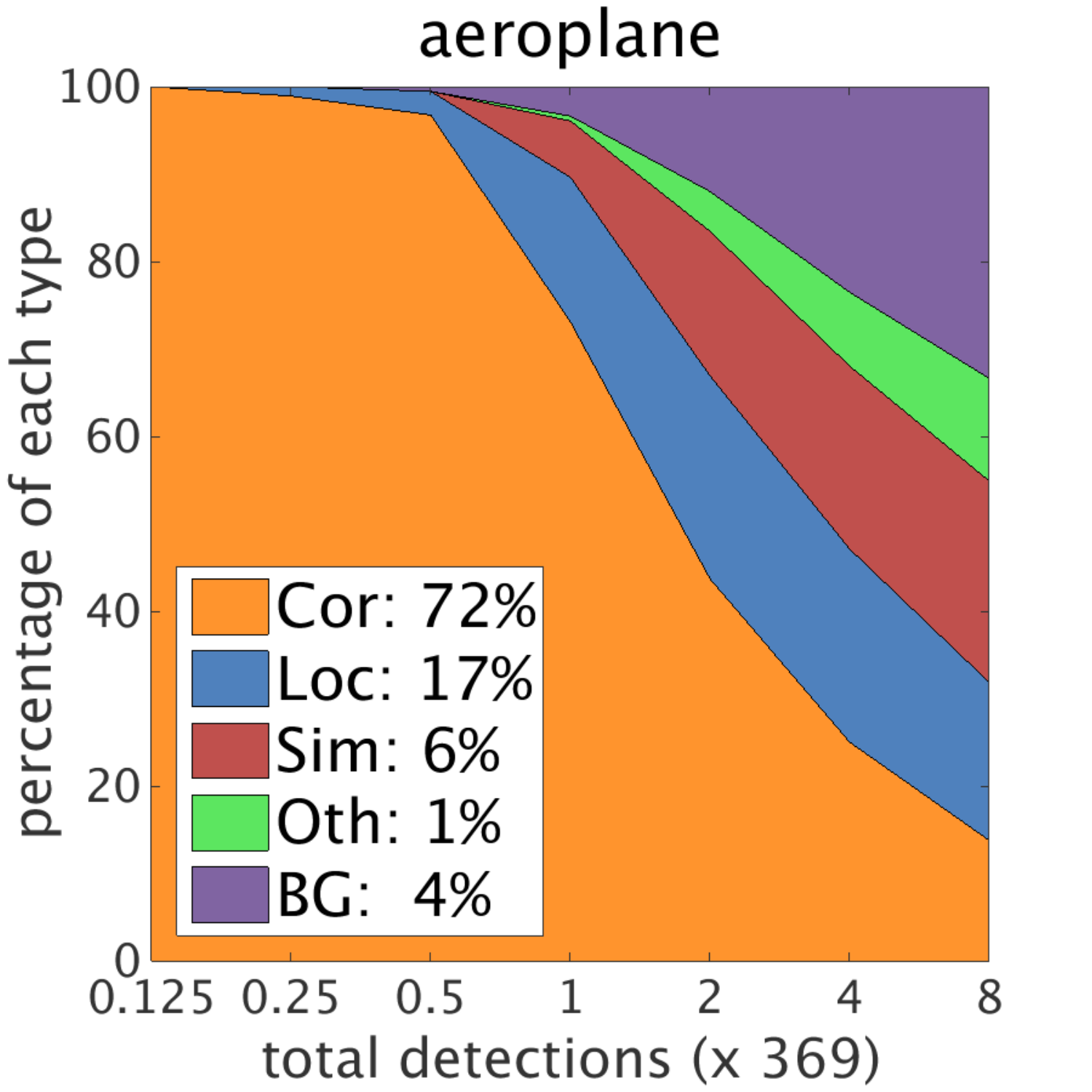} &
		\includegraphics[width=0.14\linewidth,trim=8 0 38 0,clip=true]{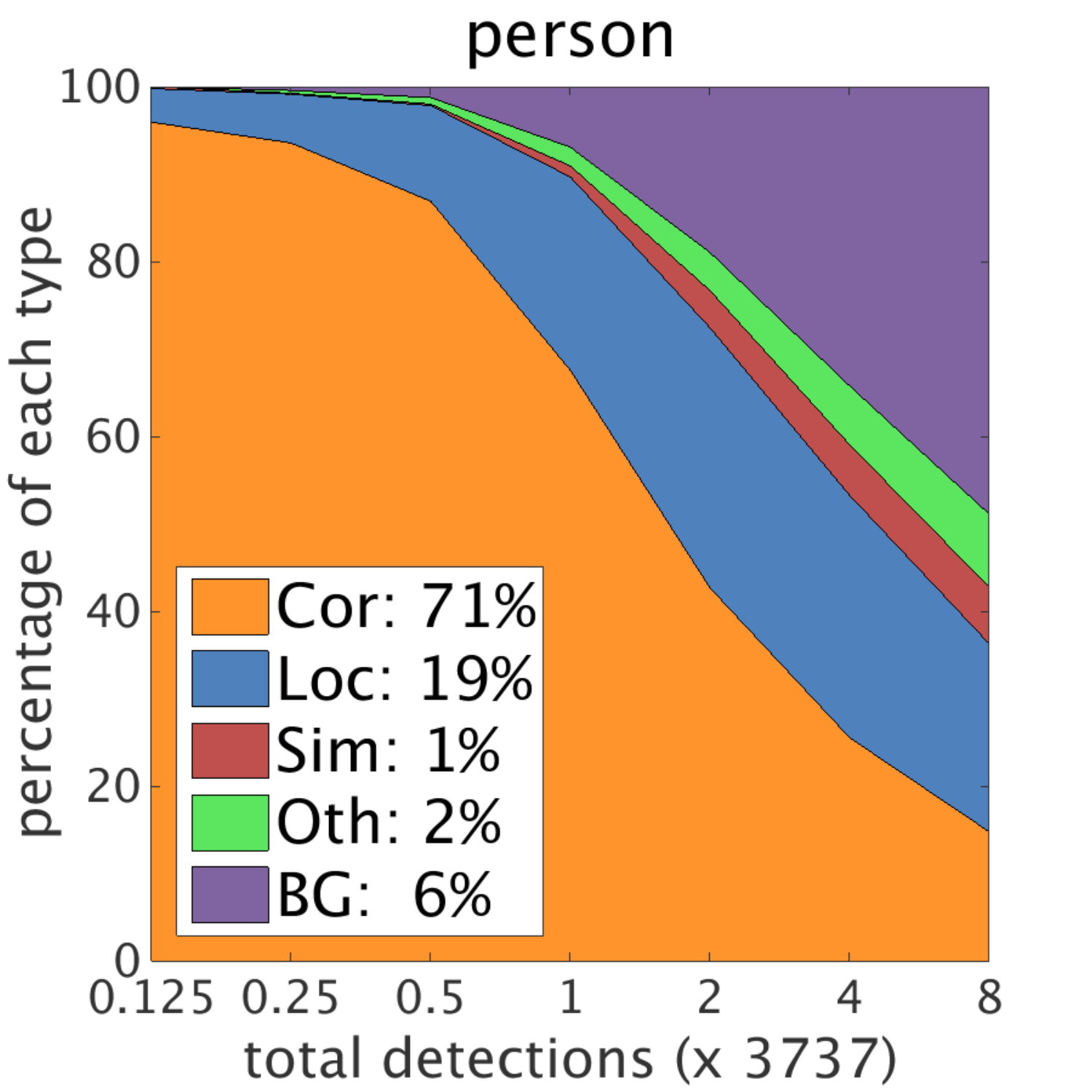} &
		\includegraphics[width=0.14\linewidth,trim=8 0 38 0,clip=true]{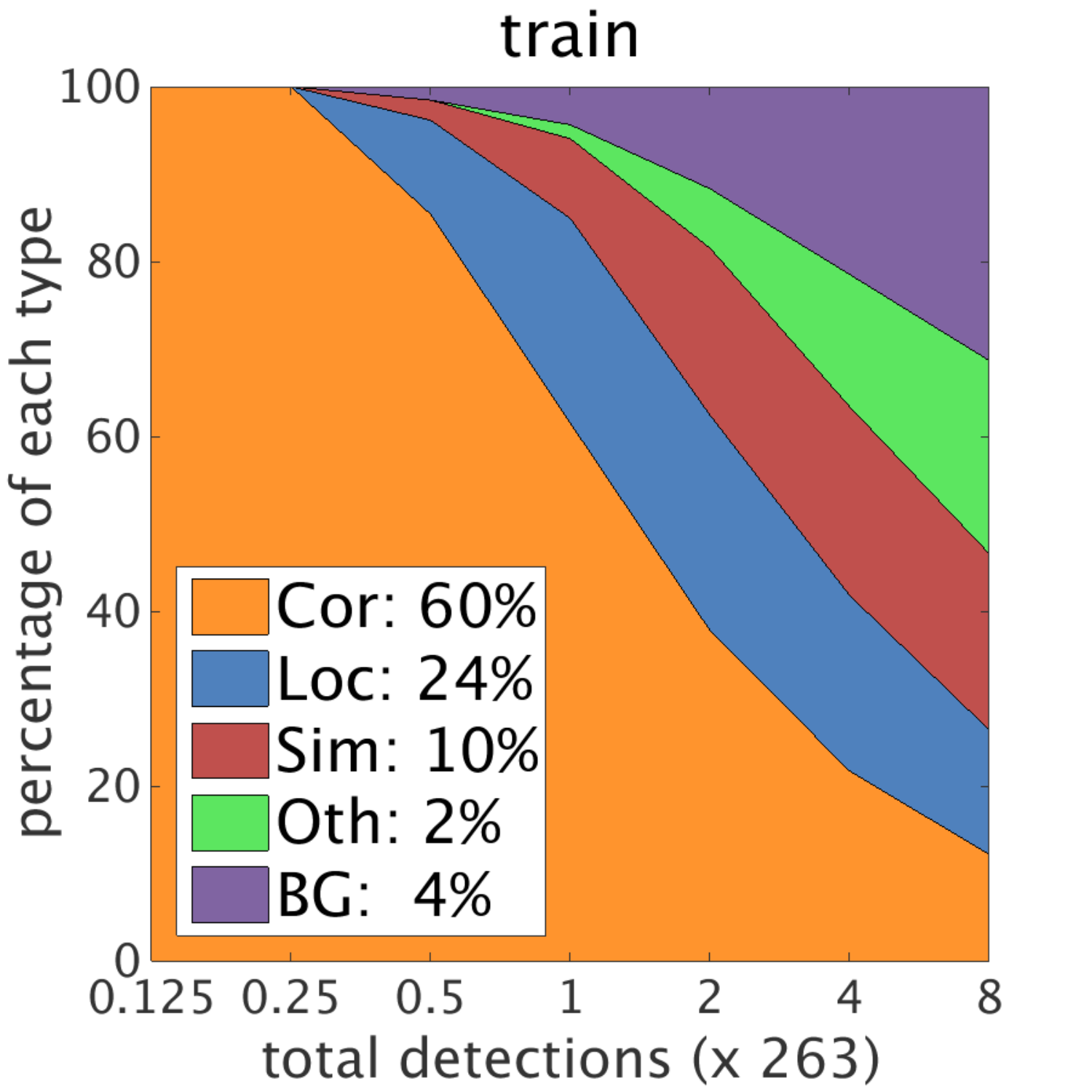} &
		\includegraphics[width=0.14\linewidth,trim=8 0 38 0,clip=true]{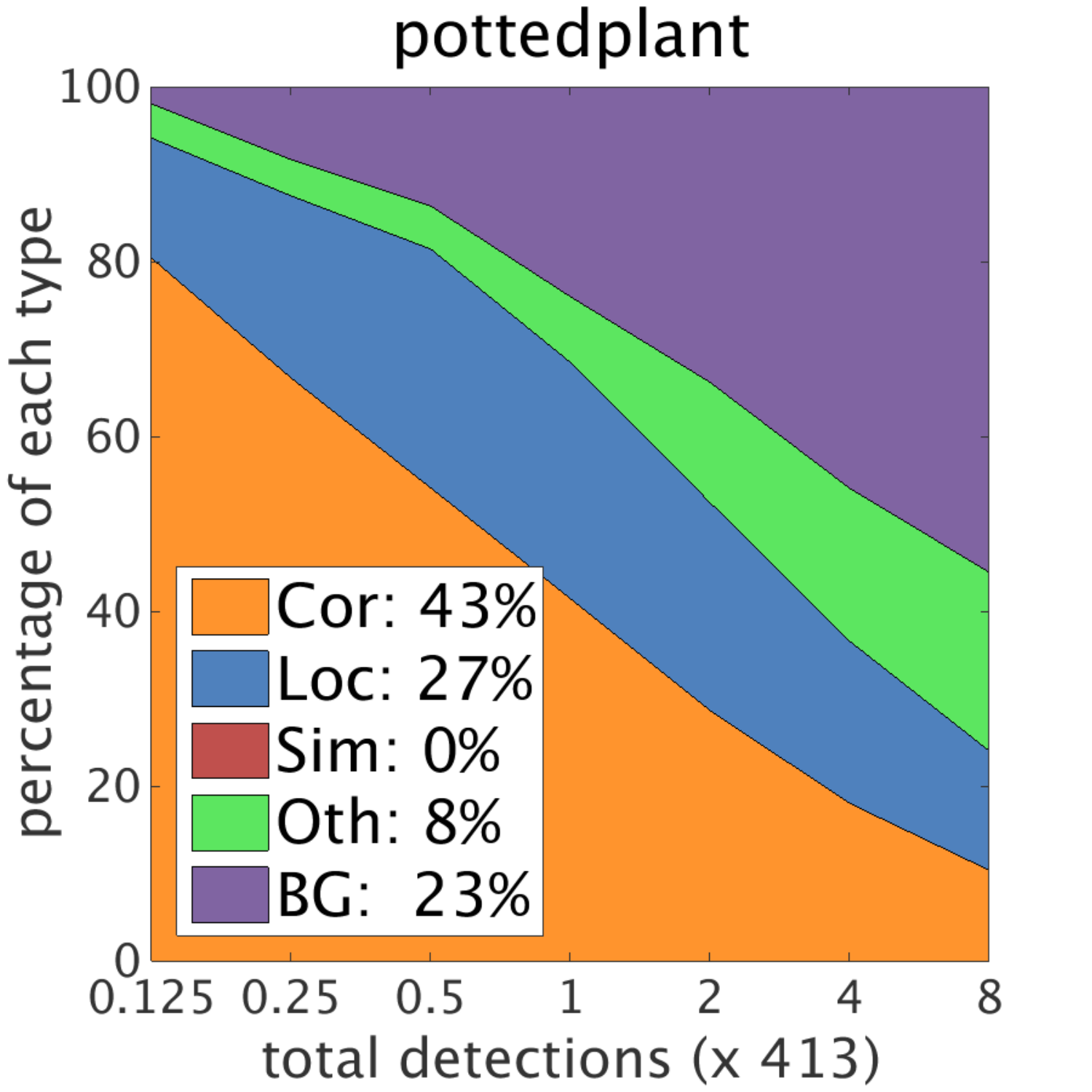} &
		\includegraphics[width=0.14\linewidth,trim=8 0 38 0,clip=true]{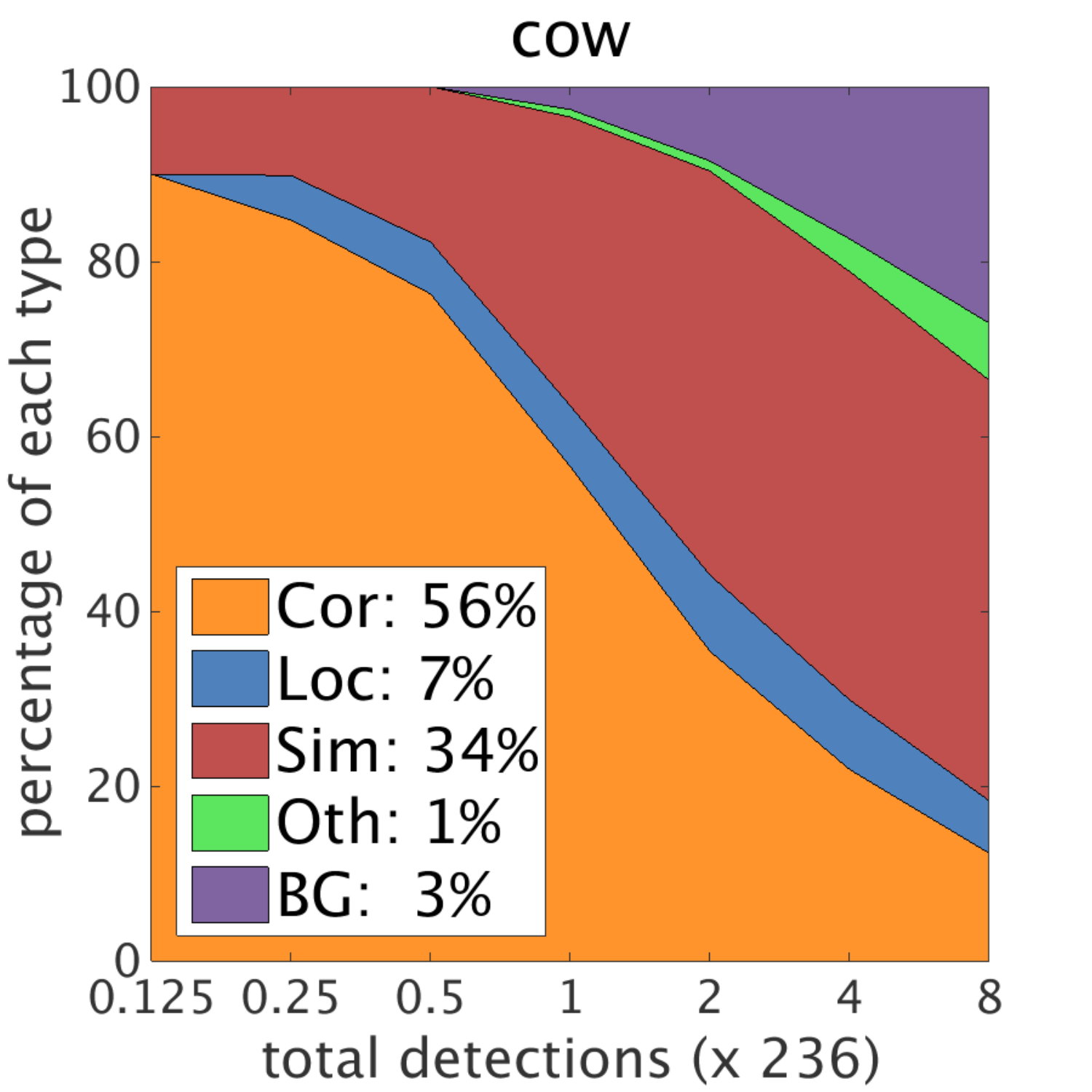} &
		\includegraphics[width=0.14\linewidth,trim=8 0 38 0,clip=true]{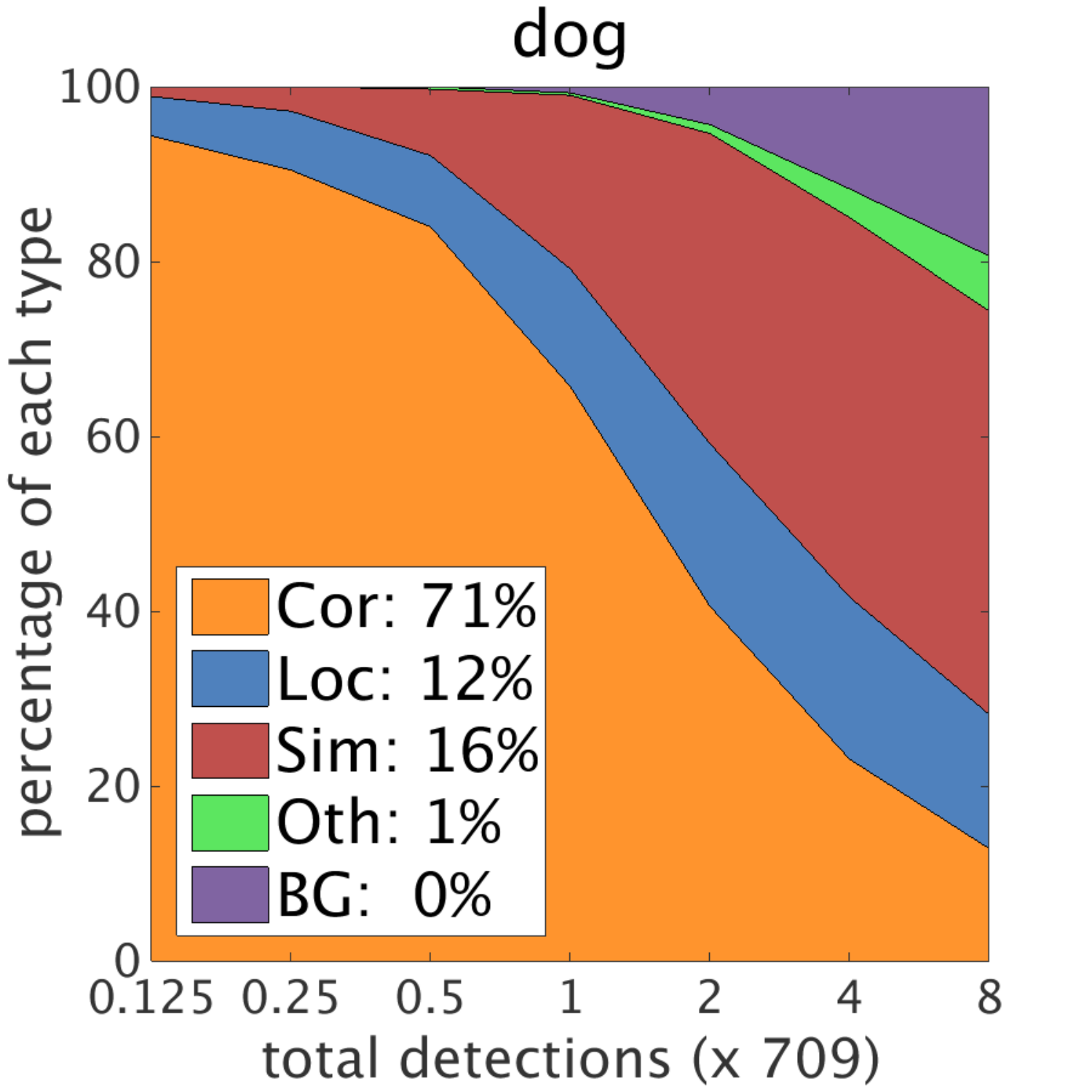}\\
		
		\rotatebox{90}{ \hspace{6mm}segDeepM} & \includegraphics[width=0.14\linewidth,trim=8 0 38 0,clip=true]{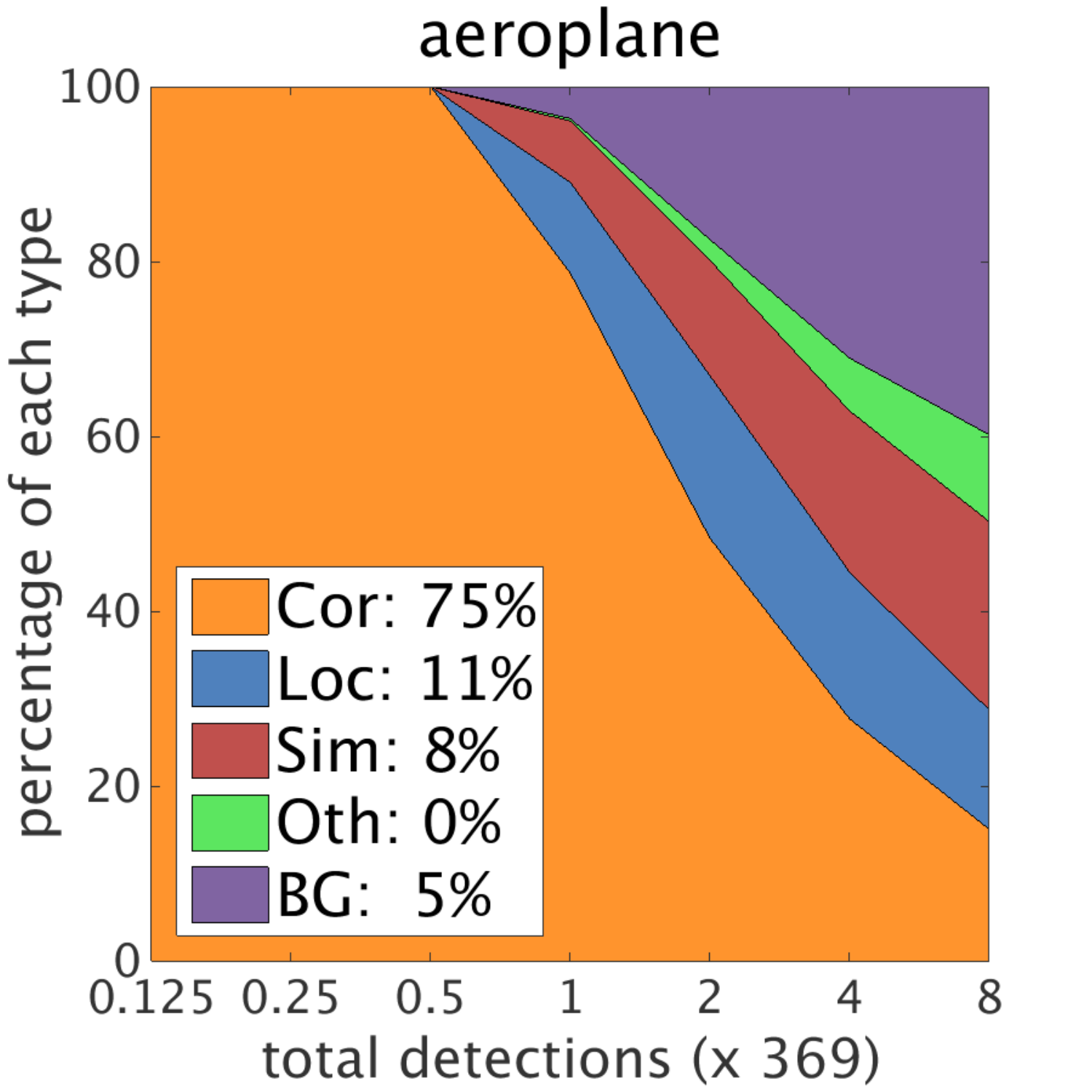}&
		\includegraphics[width=0.14\linewidth,trim=8 0 38 0,clip=true]{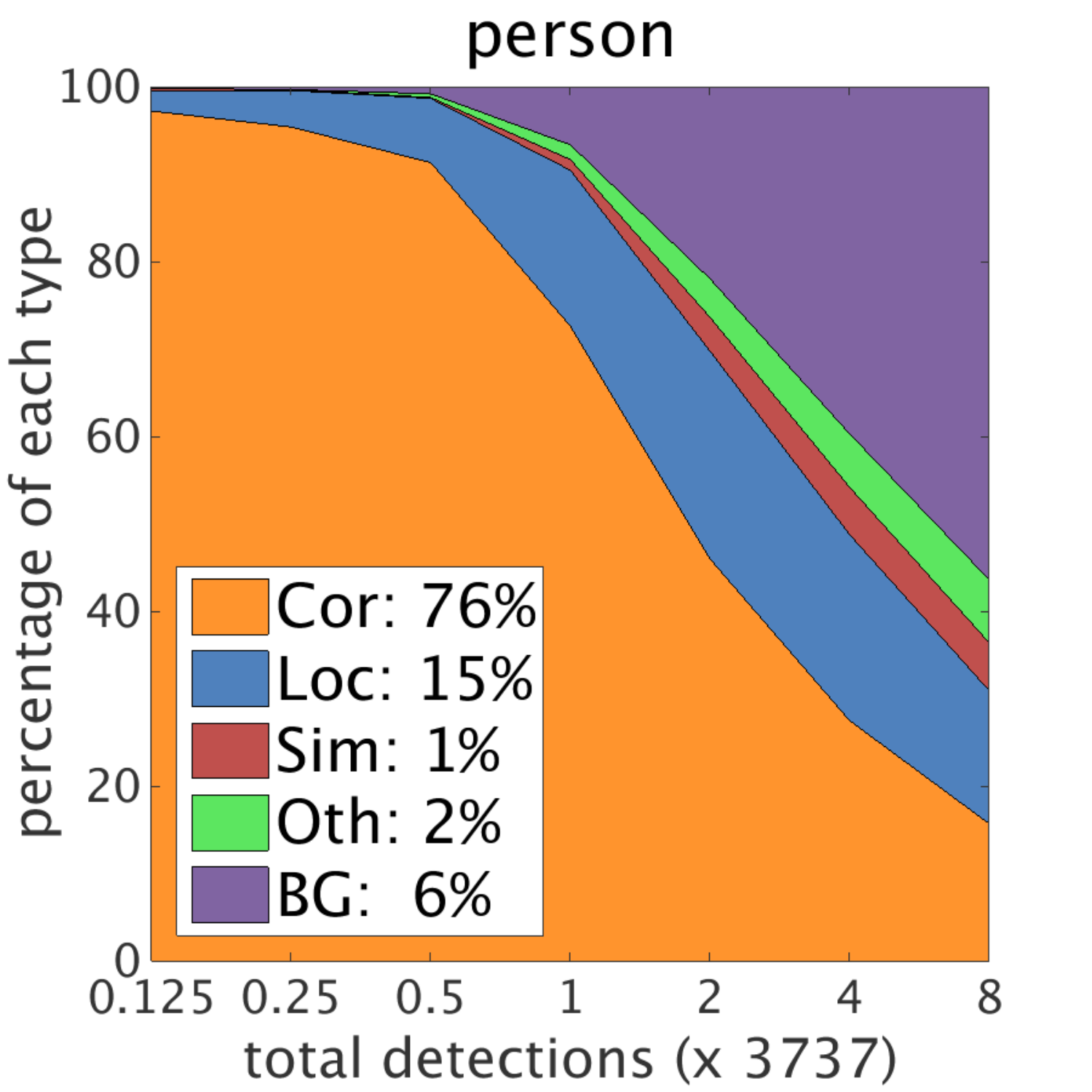} &
		\includegraphics[width=0.14\linewidth,trim=8 0 38 0,clip=true]{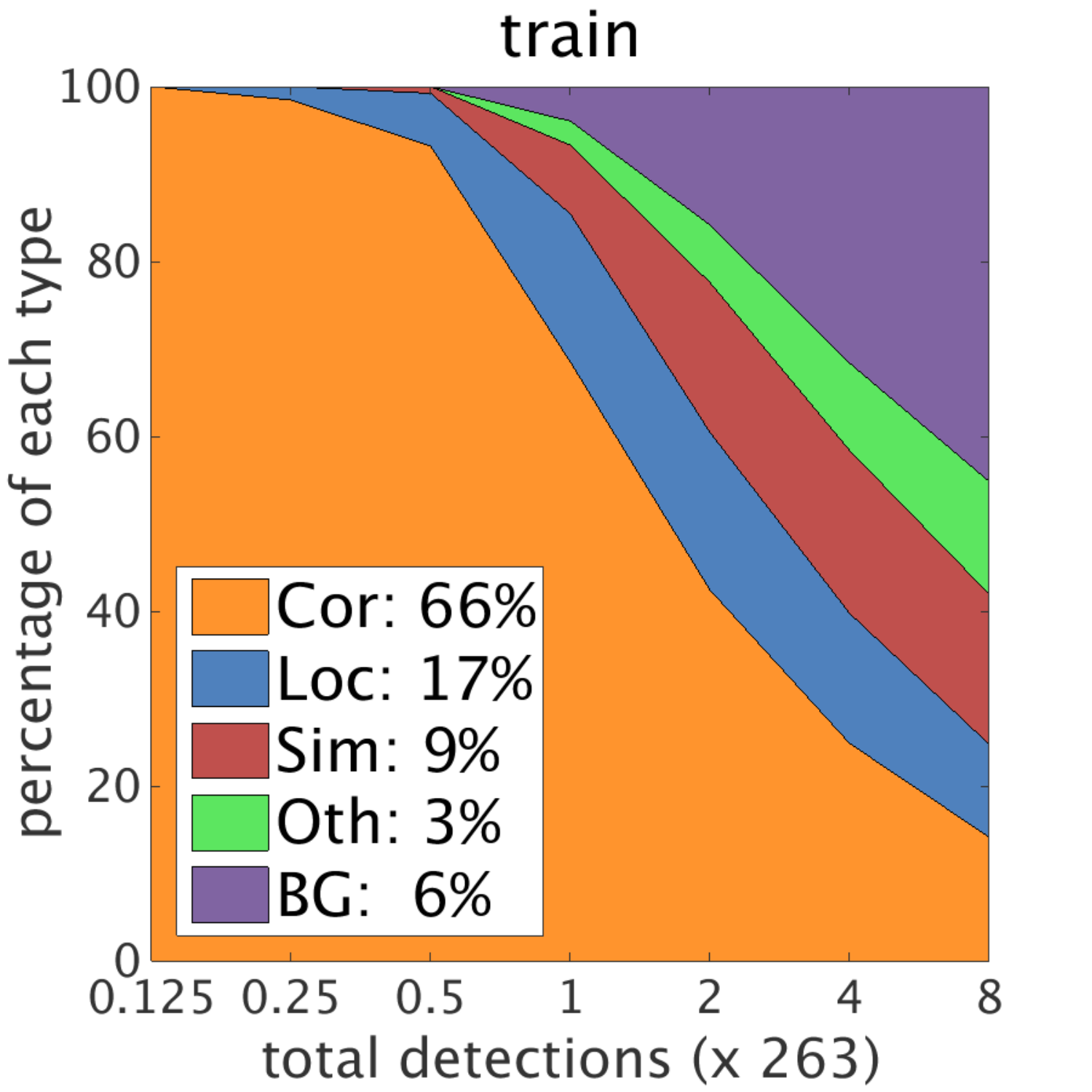} &
		\includegraphics[width=0.14\linewidth,trim=8 0 38 0,clip=true]{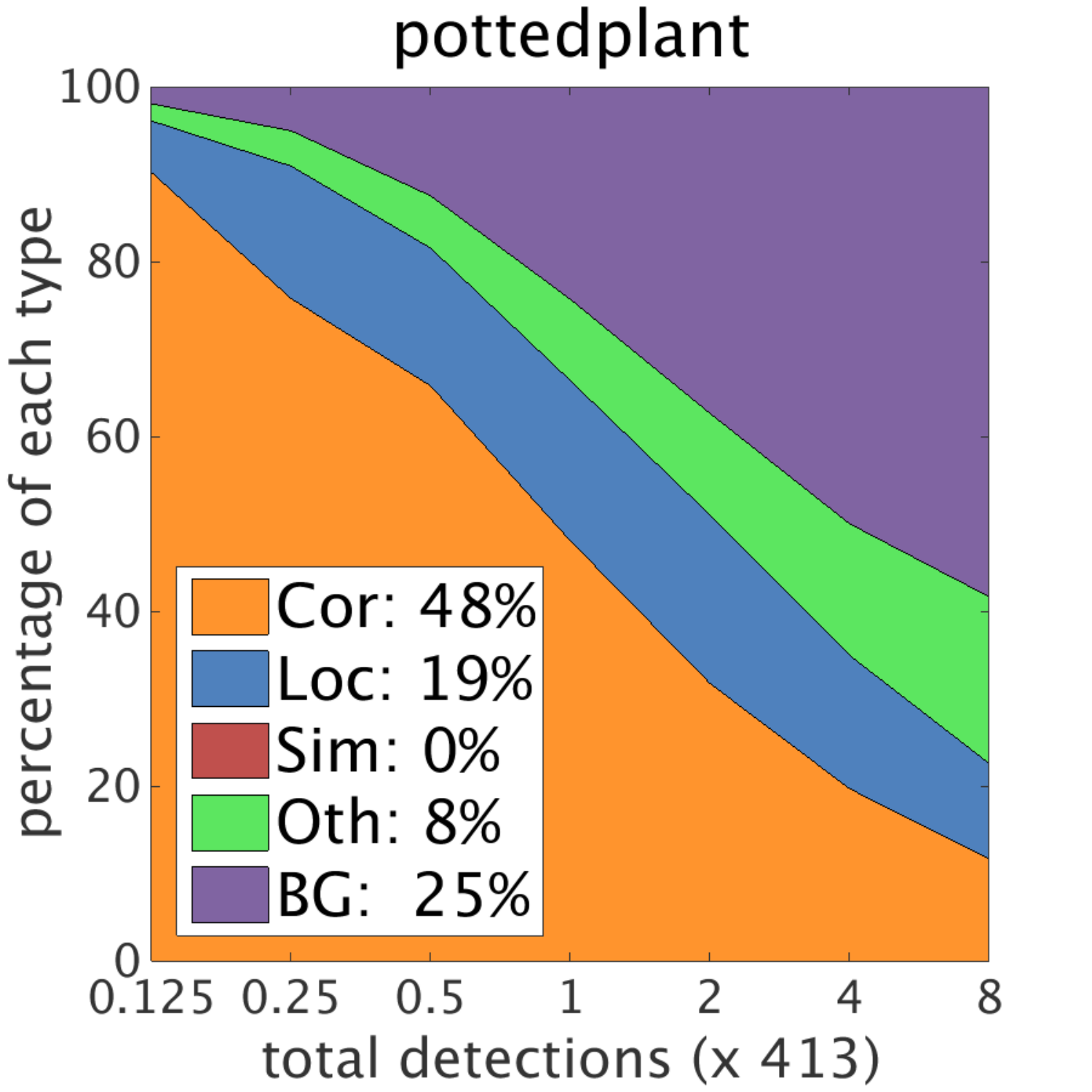}&
		\includegraphics[width=0.14\linewidth,trim=8 0 38 0,clip=true]{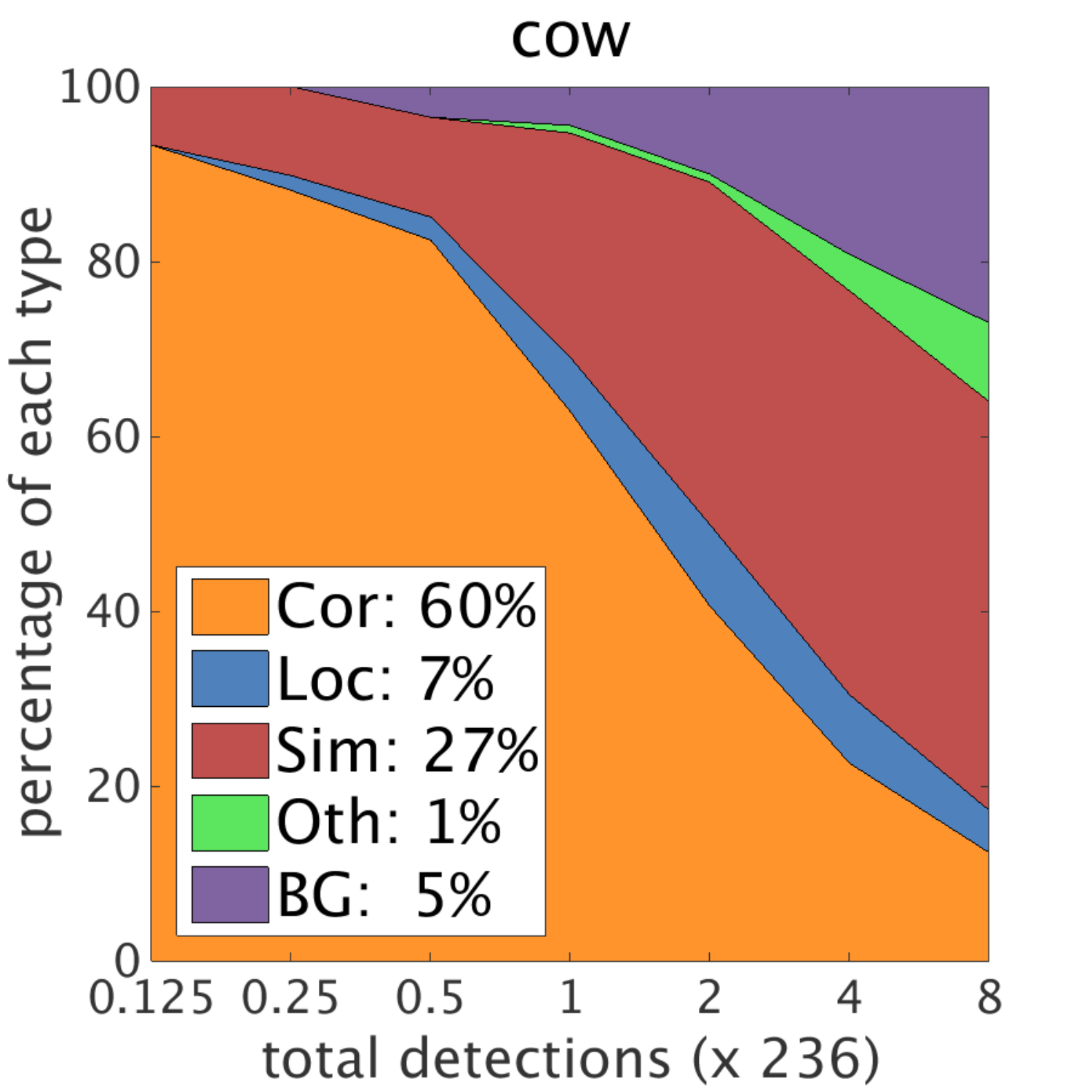} &
		\includegraphics[width=0.14\linewidth,trim=8 0 38 0,clip=true]{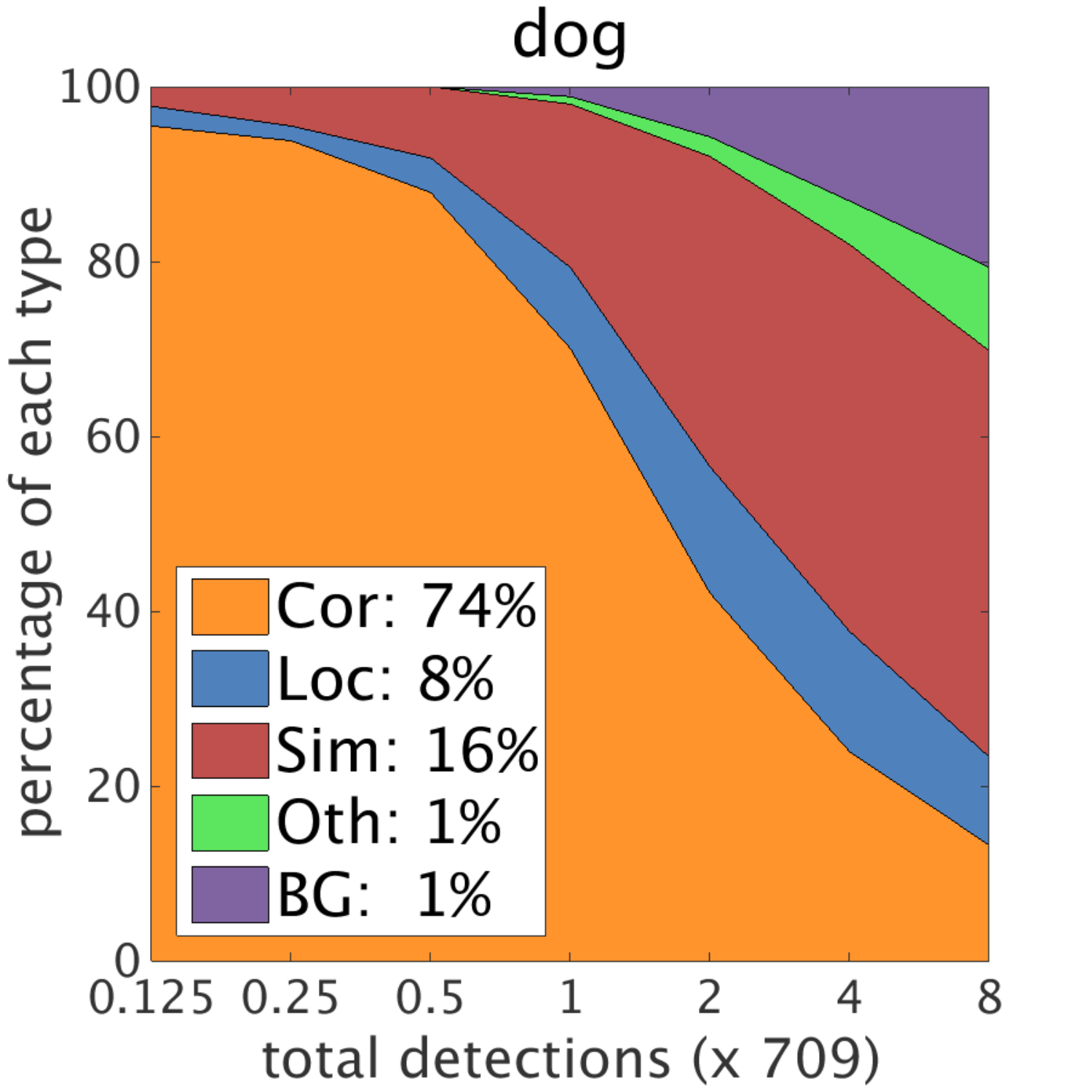}\\
	\end{tabular}
	\vspace{-4mm}
	\caption{Error analysis in style of~\cite{Hoiem14} of our detector compared to RCNN.}
	\label{fig:error}
	\vspace{-4mm}
\end{figure*}

\section{Experimental Evaluation}

We evaluate our method on the main object detection benchmark PASCAL VOC. We provide a details ablative study of different potentials and choices in our model in Subsec.~\ref{sec:ablative}. In Subsec.~\ref{sec:test} we test our method on PASCAL's held-out test set and compare it to the current state-of-the-art methods.

\subsection{A Detailed Analysis of Our Model on Val}
\label{sec:ablative}

We first evaluate our detection performance on $Val$ set of the PASCAL VOC 2010 detection dataset. We train all methods on the $train$ subset and evaluate the detection performance using the standard PASCAL criterion. We provide a detailed performance analysis of each proposed potential function, which we denote with $seg$ (segmentation) and expanded network $exp$ (the contextual network) in Table~\ref{table:val}. We also compare our iterative bounding box regression approach, referred to as $ibr$, to the standard bounding box regression,  referred to as $br$,~\cite{girshick2013rich}. 

R-CNN~\cite{girshick2013rich} serves as our main baseline.  To better justify our model, we provide an additional baseline, where we simply augment the set of Selective Search boxes used originally by the R-CNN with the CPMC proposal set. We call this approach RCNN+CPMC in the Table (second row). To contrast our model with segDPM, which originally uses segmentation features in a DPM-style formulation, we simplify our model to use their exact features. Instead of HOG, however, we use CNNs for a fair comparison. We also use their approach to generate segments, by finding connected components in the final output of CPMC-O2P segmentation~\cite{FidlerBottomCVPR2013}. This approach is referred to as segDPM+CNN (third row in Table~\ref{table:val}).

\vspace{-0.mm}
Observe that using a small set of additional segments brings a $1\%$ improvement for RCNN+CPMC over the R-CNN baseline. Using a segDPM+CNN approach yields a more significant improvement of $2.1\%$. With our segmentation features we get an additional $1.1\%$ increase over segDPM+CNN, thus justifying our feature set. Interestingly, this $3\%$ boost over R-CNN is achieved by our simple segmentation features which require only $223$ additional parameters. The Table also shows a steady improvement of each additional added potential/step, with the highest contribution achieved by the expanded contextual network.

\begin{figure}[t!]
        \centering
        \includegraphics[width=0.242\textwidth]{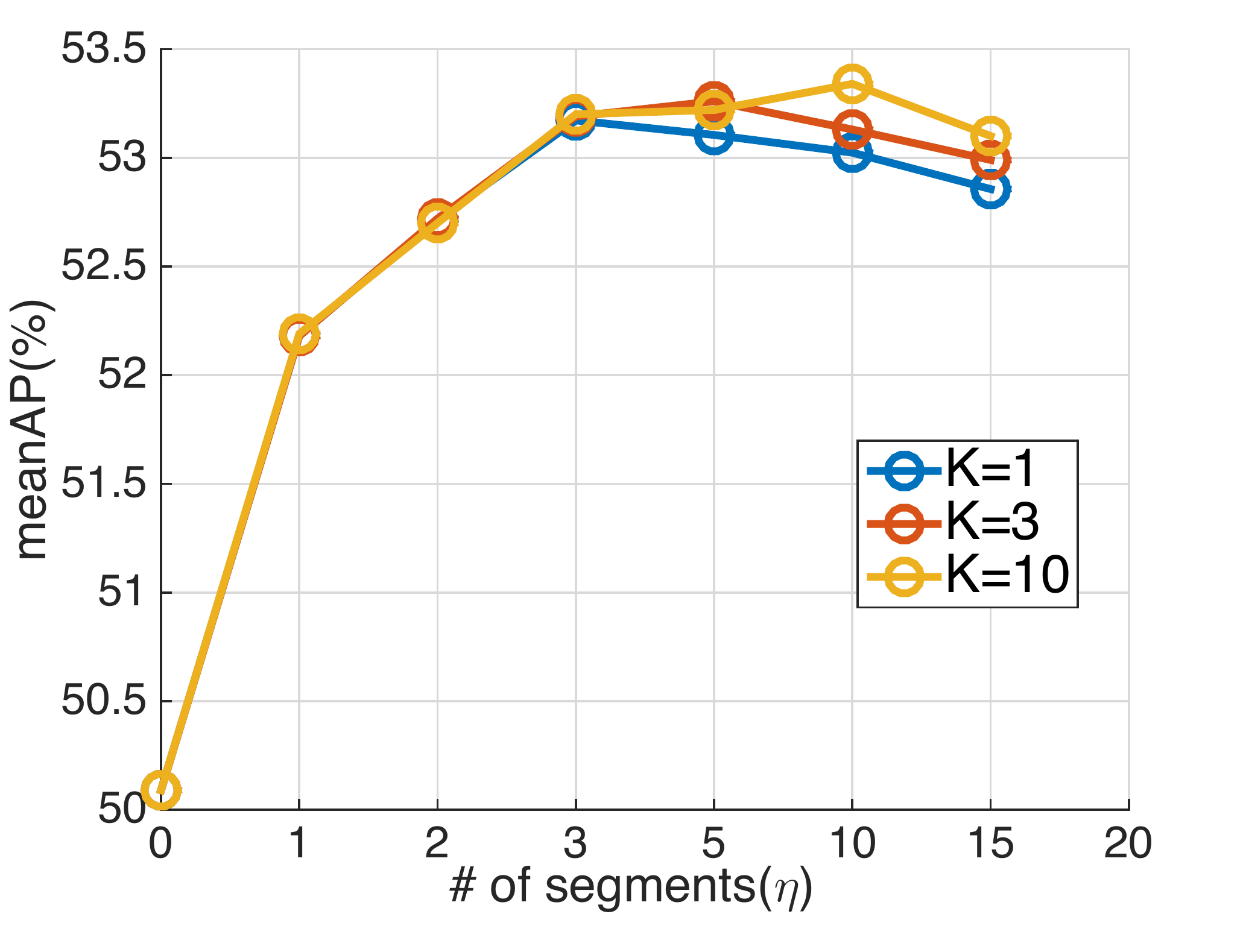}\includegraphics[width=0.242\textwidth]{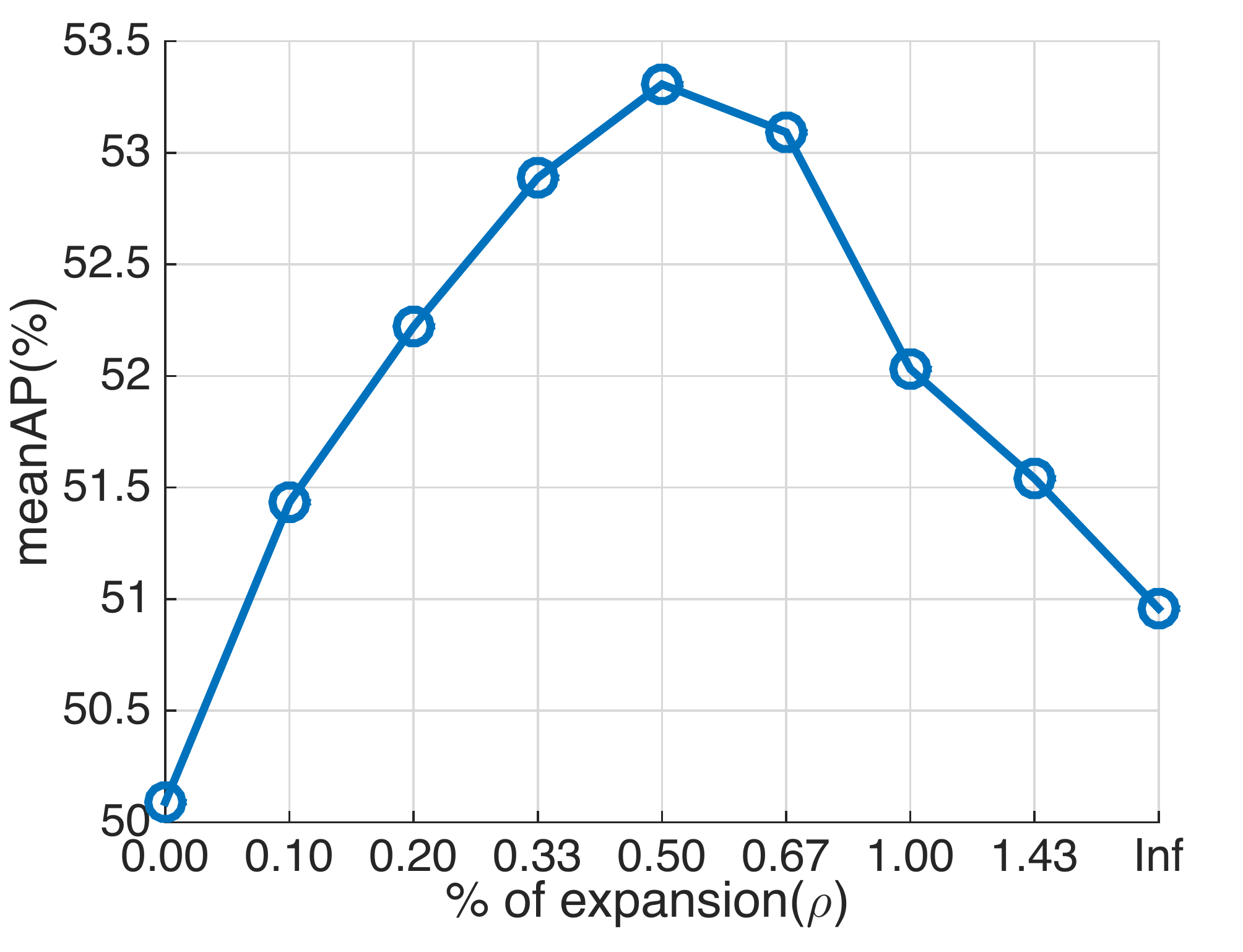}
        \vspace{-4mm}
        \caption{mAP for segDeepM \wrt  {\bf (left)} \# of segments per image ($\eta$). $\eta=0$ indicates no segments are used. {\bf (right)}  box expansion ratio $\rho$. $\rho=0$ disables context and $\rho=\infty$ indicates full image context. Only contextual features used in this experiment. Both plots for PASCAL VOC 2010 $val$.}
        \label{ap_vs_numseg}
        \vspace{-4mm}
\end{figure}

\begin{figure*}[htb]
\vspace{-2mm}
	\centering
	\begin{subfigure}[b]{0.210\textwidth}
		\includegraphics[width=1\textwidth]{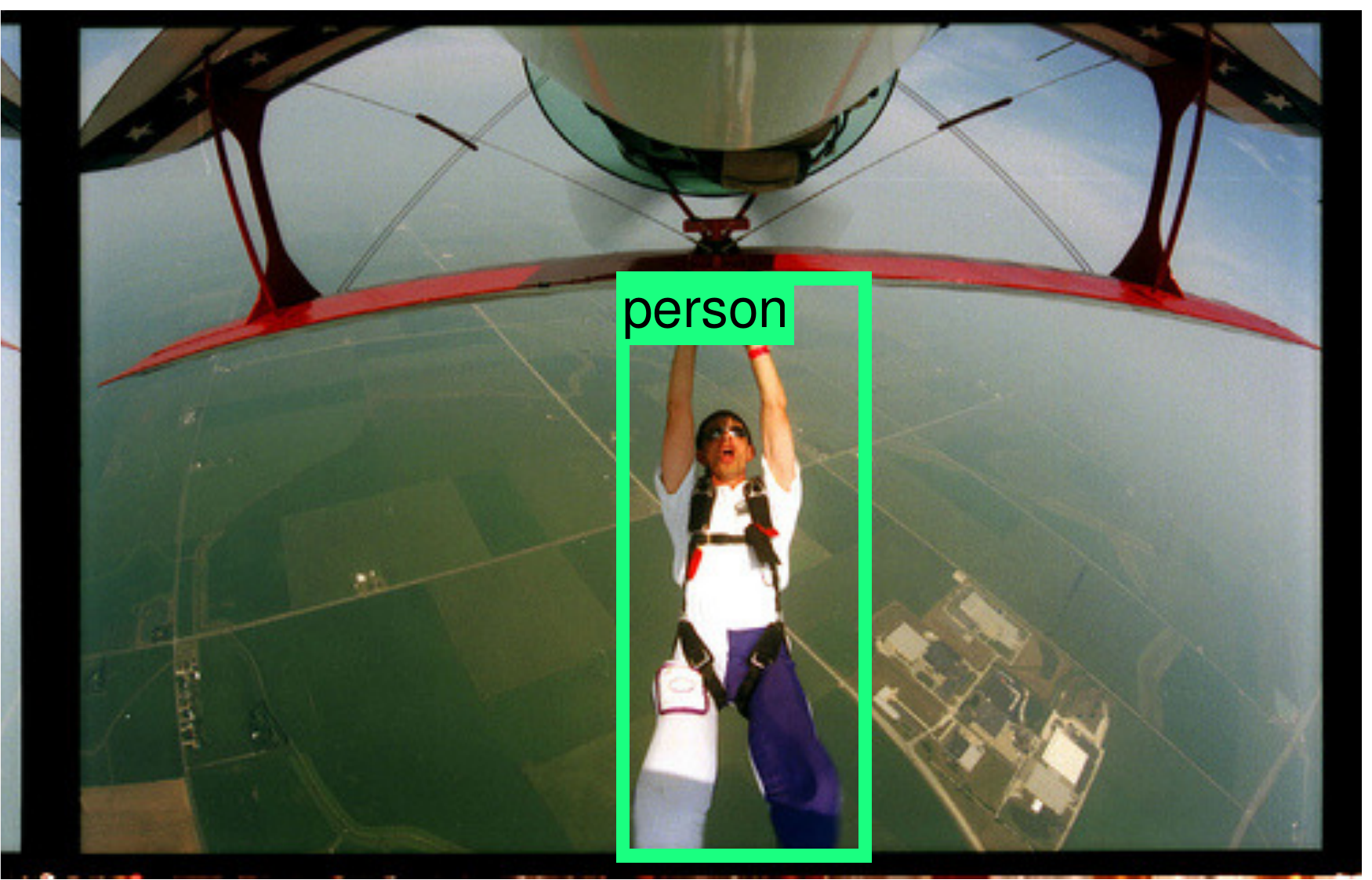}
	\end{subfigure}
	\begin{subfigure}[b]{0.210\textwidth}
		\includegraphics[width=1\textwidth]{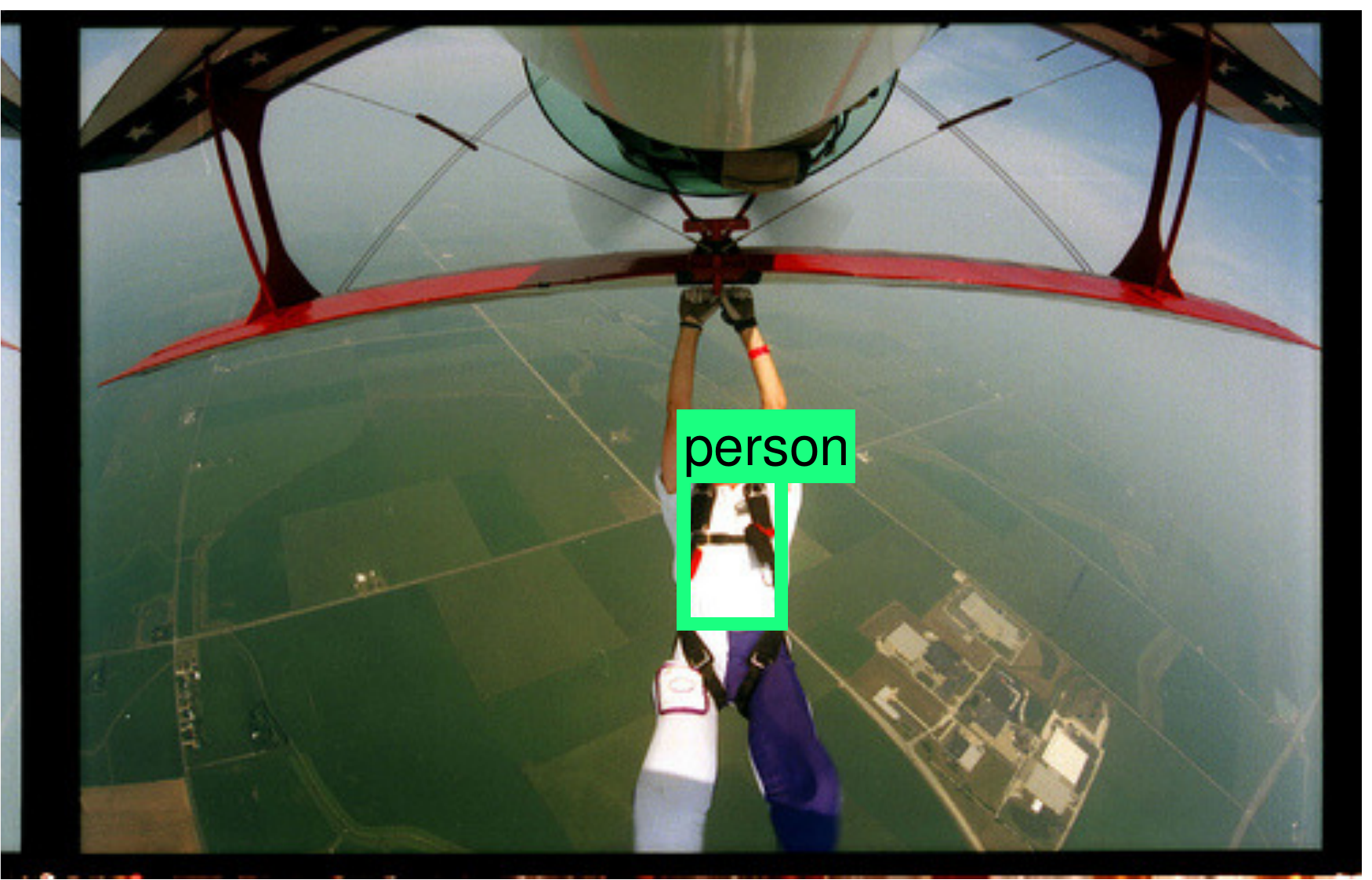}
	\end{subfigure}	
	\begin{subfigure}[b]{0.210\textwidth}
		\includegraphics[width=1\textwidth]{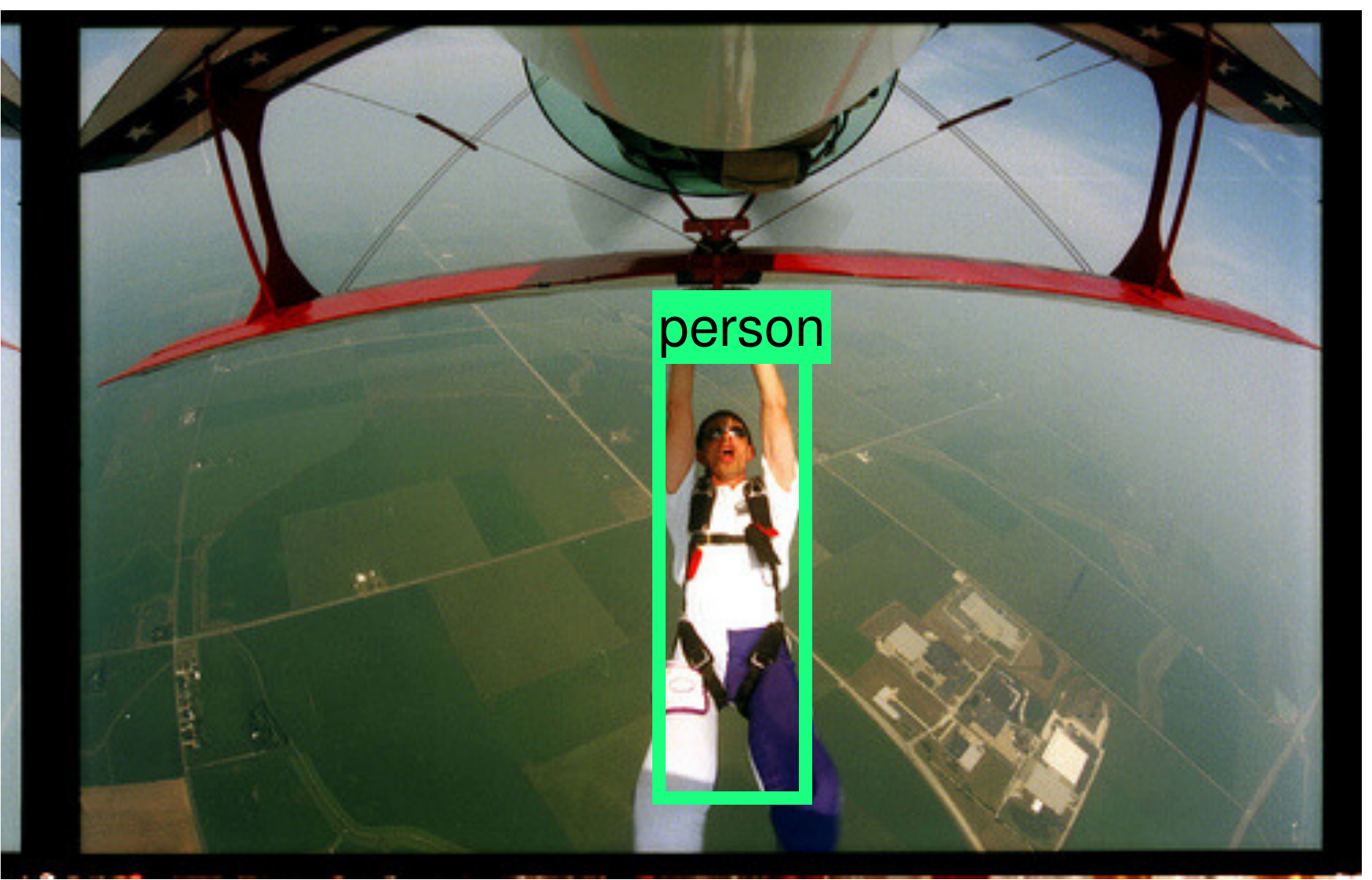}
	\end{subfigure}	
	\begin{subfigure}[b]{0.210\textwidth}
		\includegraphics[width=1\textwidth]{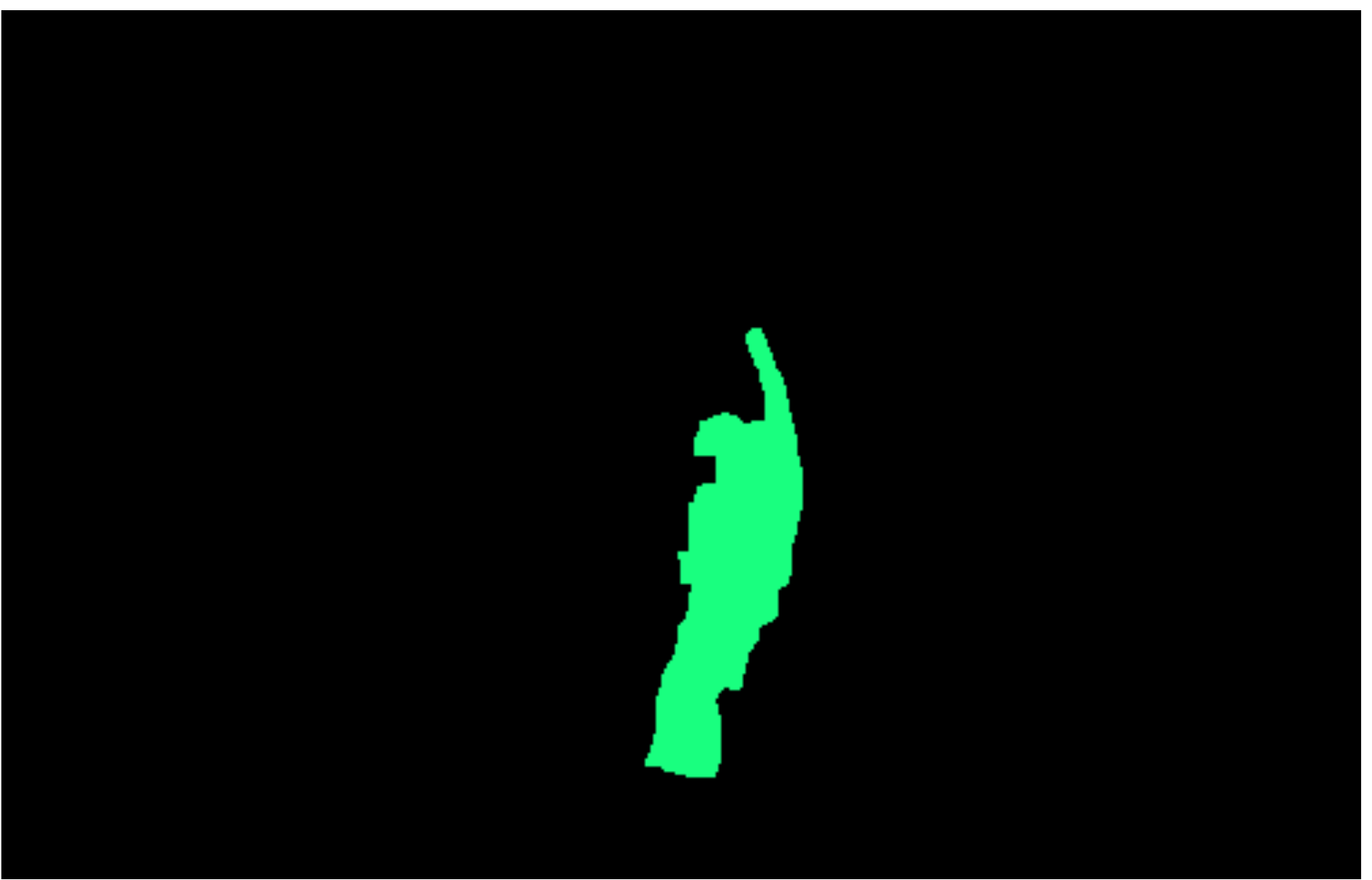}
	\end{subfigure}	
	\vspace{0.1cm}	
	\begin{subfigure}[b]{0.210\textwidth}
		\includegraphics[width=1\textwidth]{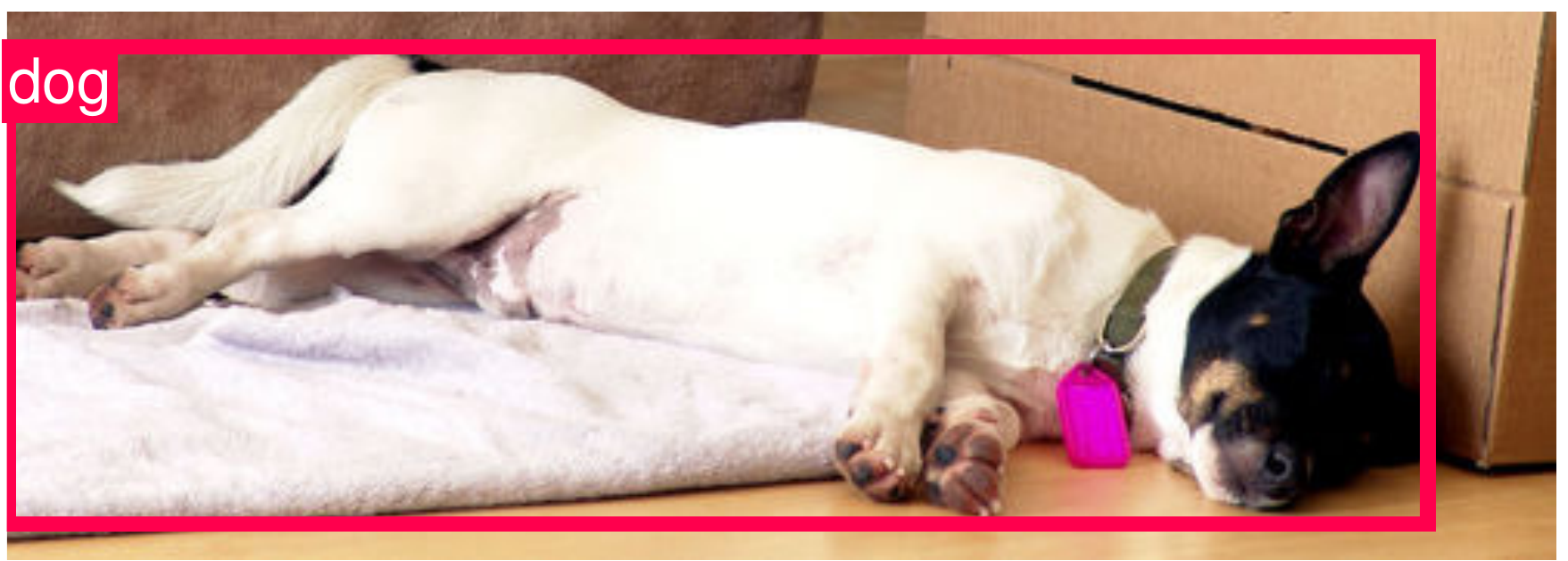}
	\end{subfigure}
	\begin{subfigure}[b]{0.210\textwidth}
		\includegraphics[width=1\textwidth]{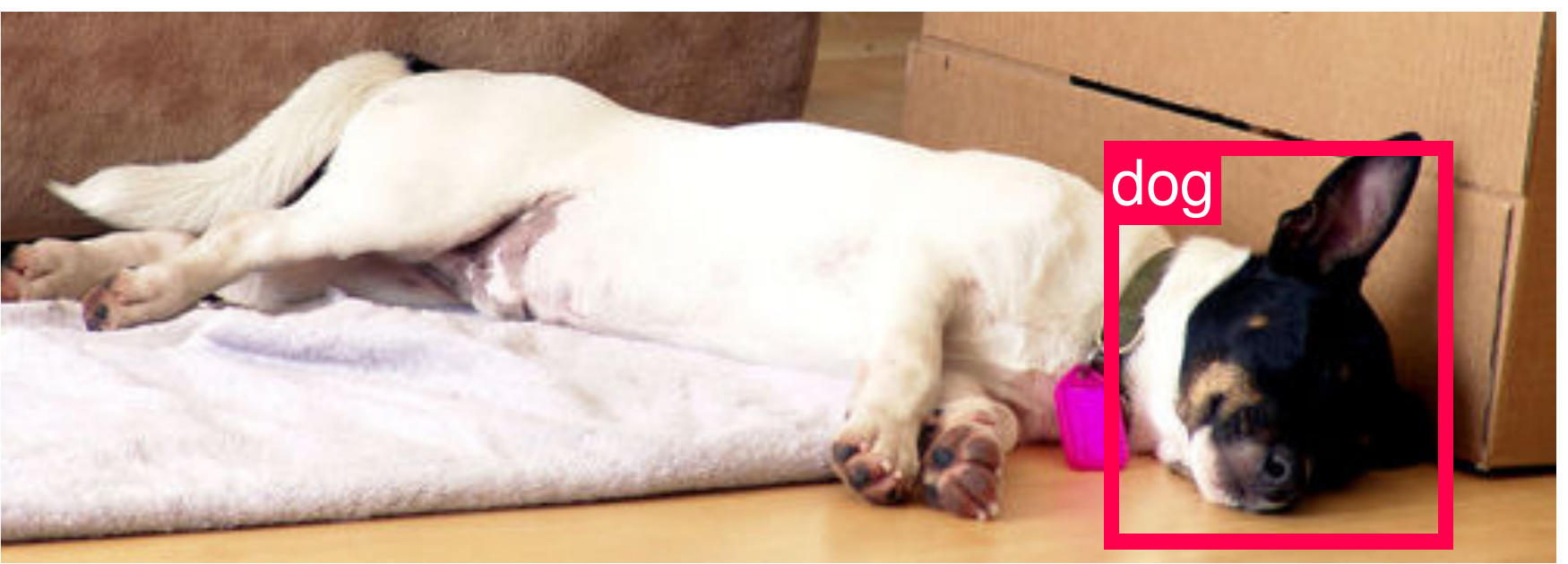}
	\end{subfigure}	
	\begin{subfigure}[b]{0.210\textwidth}
		\includegraphics[width=1\textwidth]{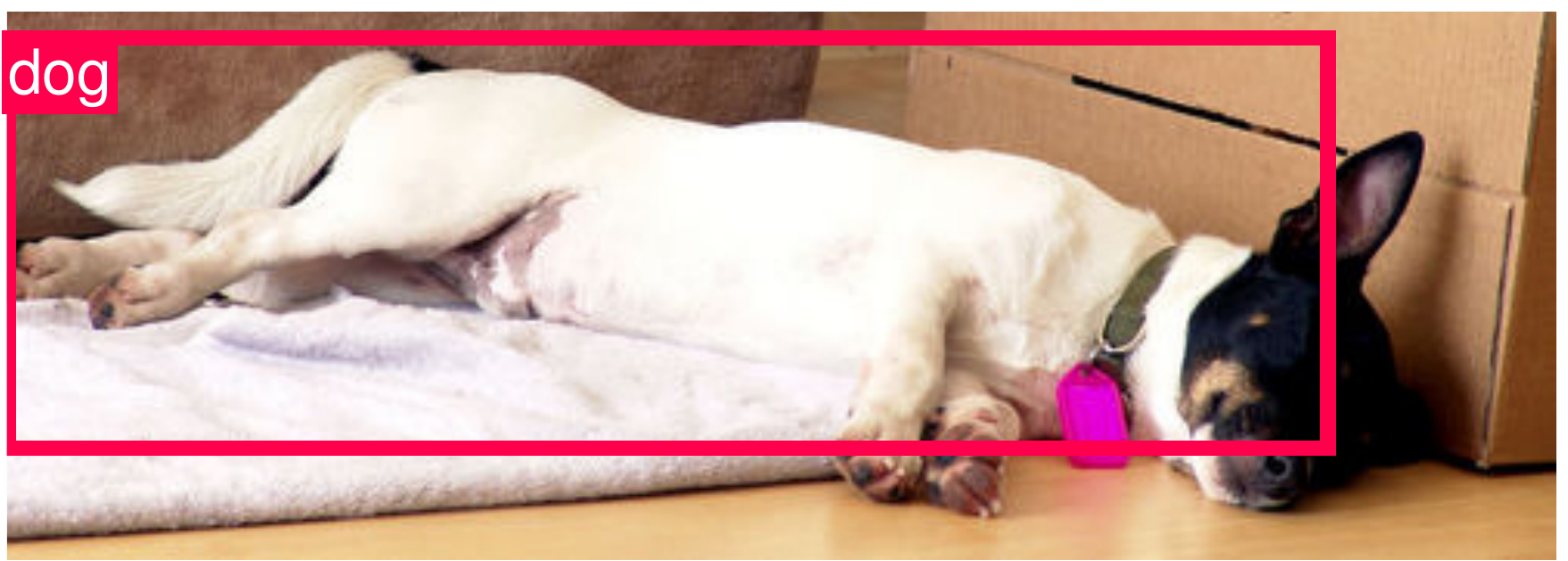}
	\end{subfigure}	
	\begin{subfigure}[b]{0.210\textwidth}
		\includegraphics[width=1\textwidth]{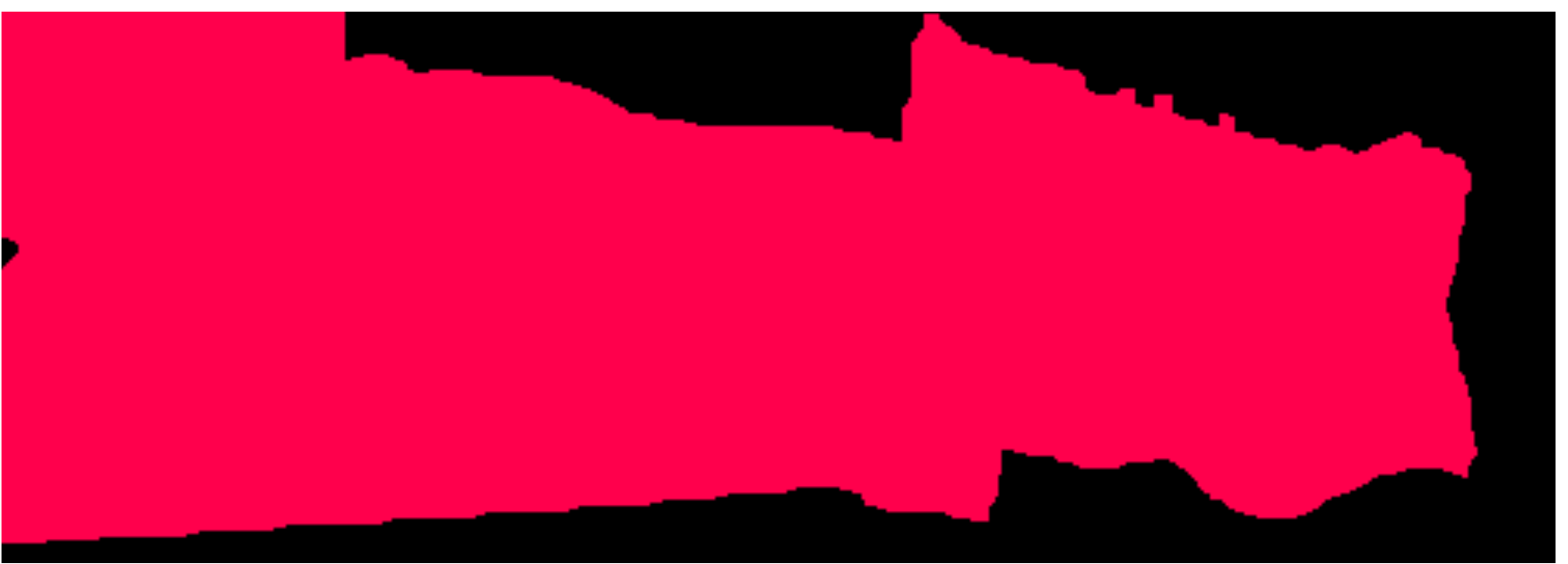}
	\end{subfigure}	
	\vspace{0.1cm}	
	\begin{subfigure}[b]{0.210\textwidth}
		\includegraphics[width=1\textwidth]{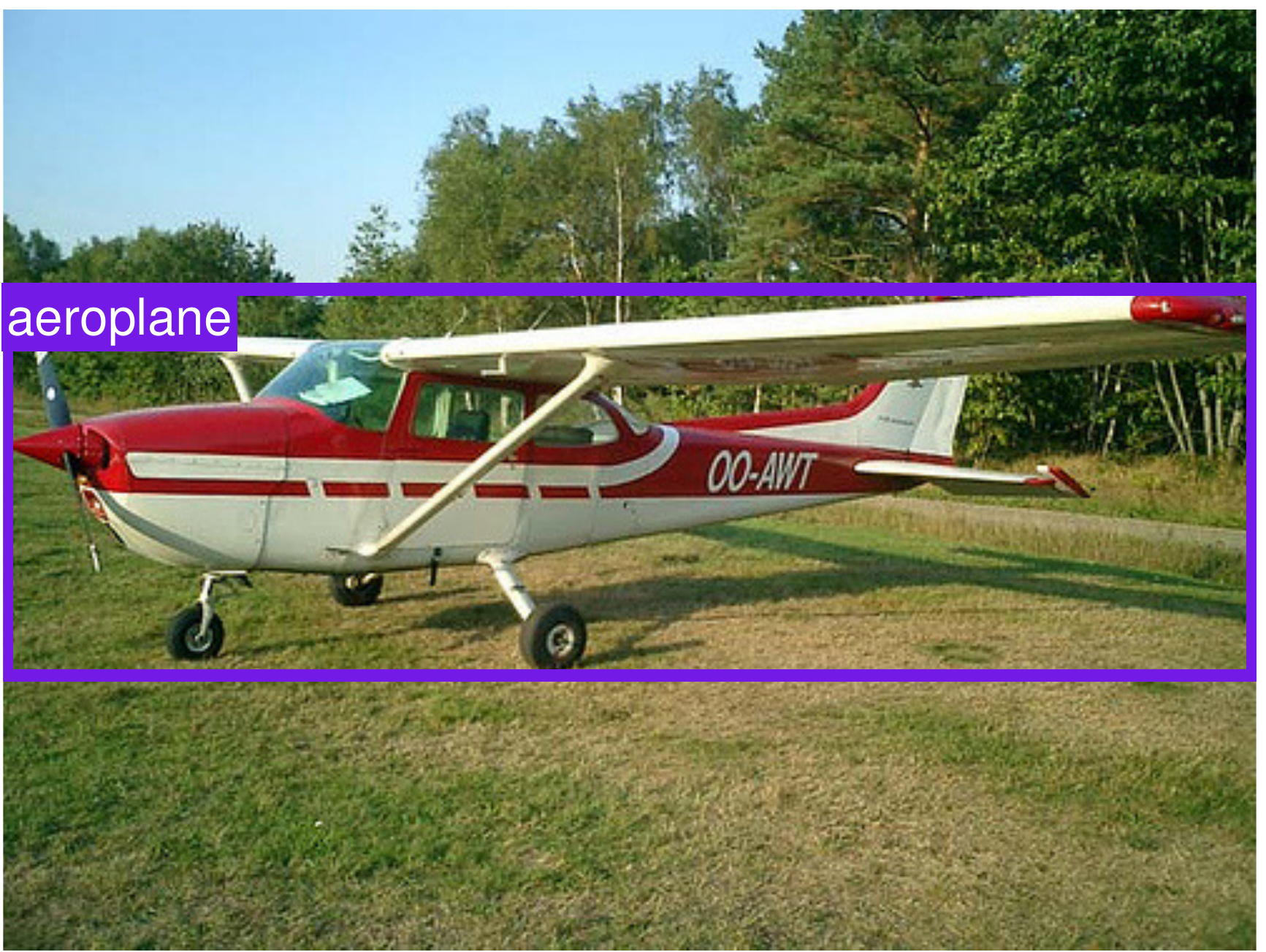}
	\end{subfigure}
	\begin{subfigure}[b]{0.210\textwidth}
		\includegraphics[width=1\textwidth]{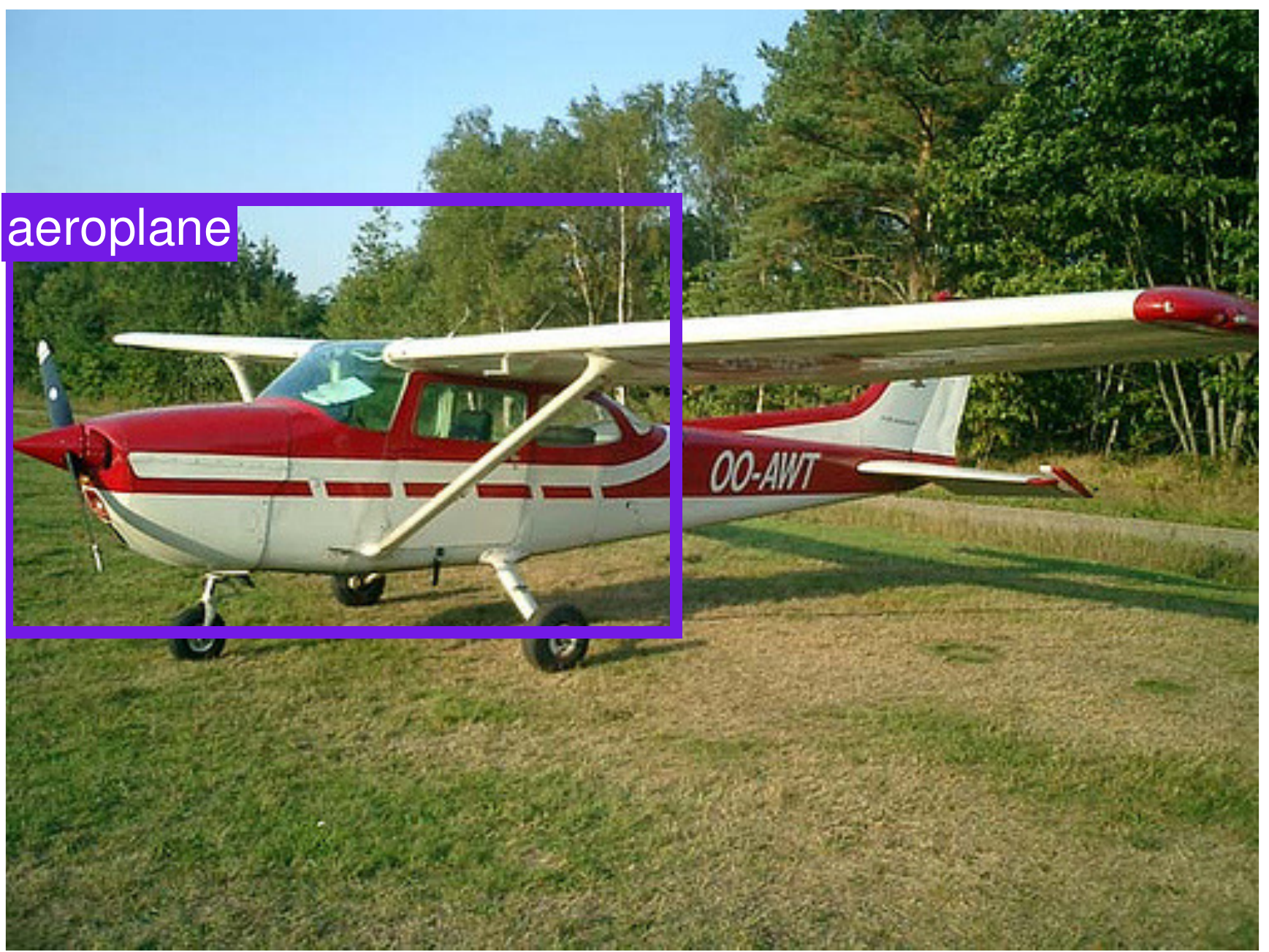}
	\end{subfigure}	
	\begin{subfigure}[b]{0.210\textwidth}
		\includegraphics[width=1\textwidth]{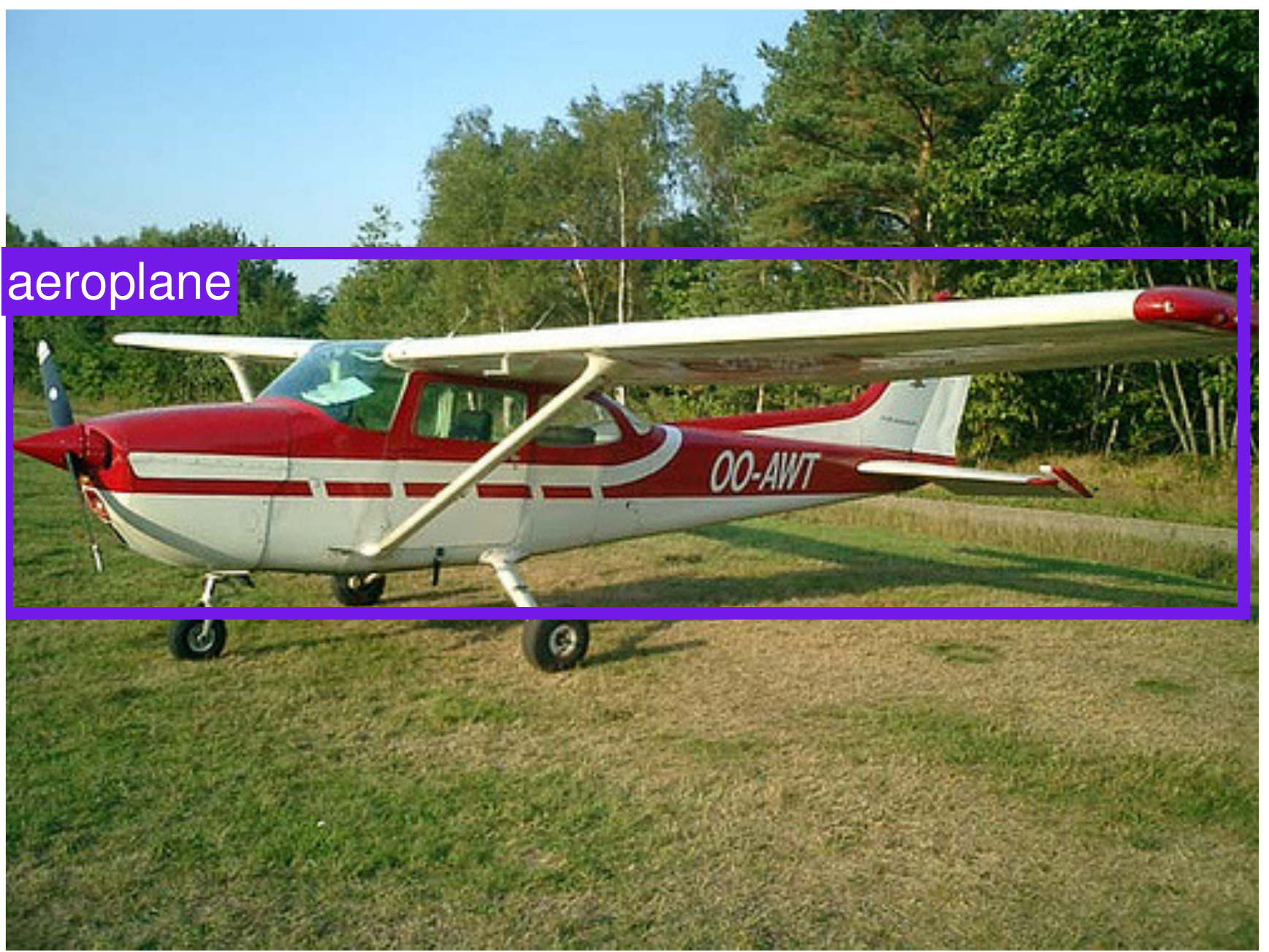}
	\end{subfigure}	
	\begin{subfigure}[b]{0.210\textwidth}
		\includegraphics[width=1\textwidth]{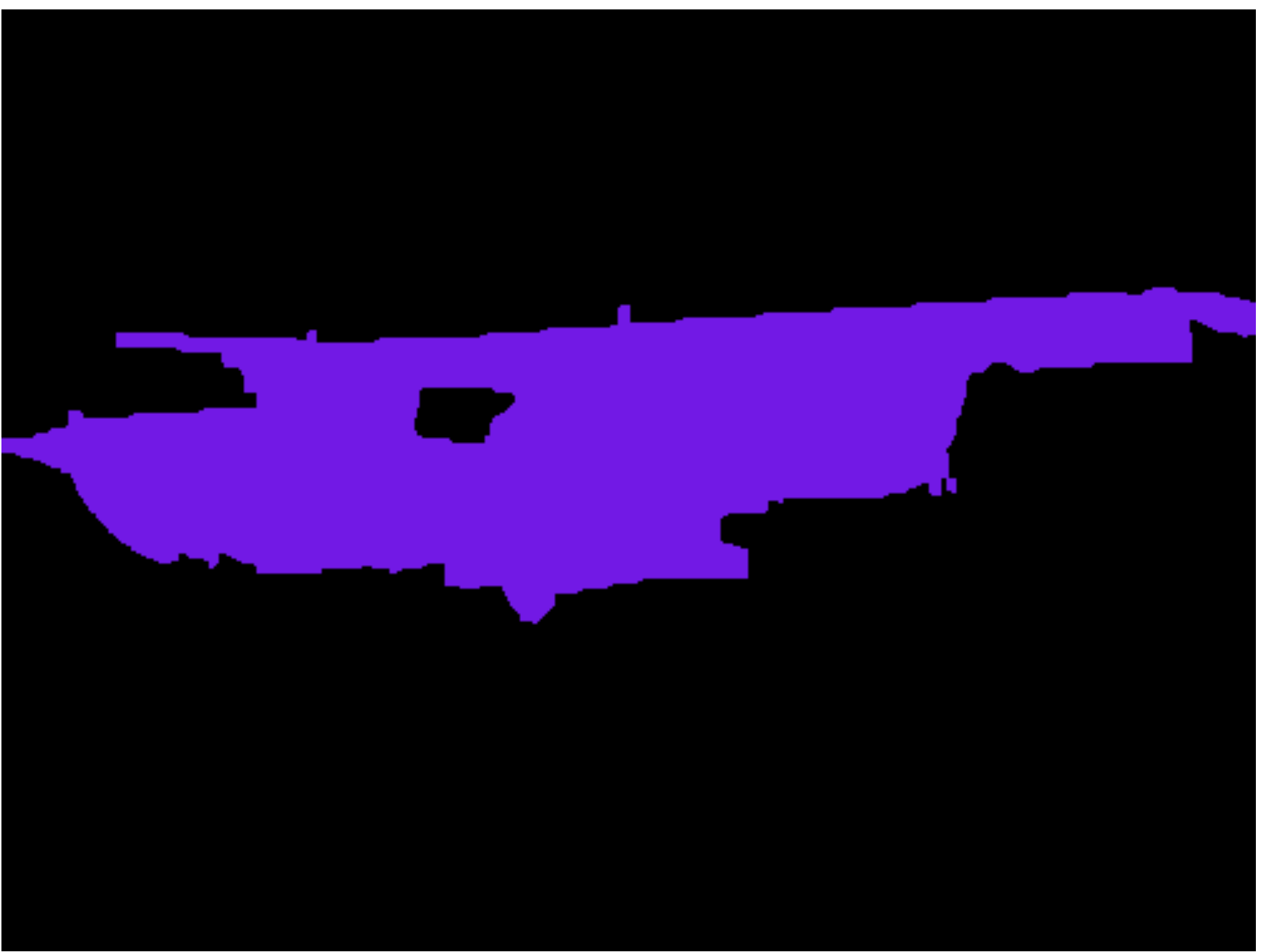}
	\end{subfigure}	
	\vspace{0.1cm}	
	\begin{subfigure}[b]{0.210\textwidth}
		\includegraphics[width=1\textwidth]{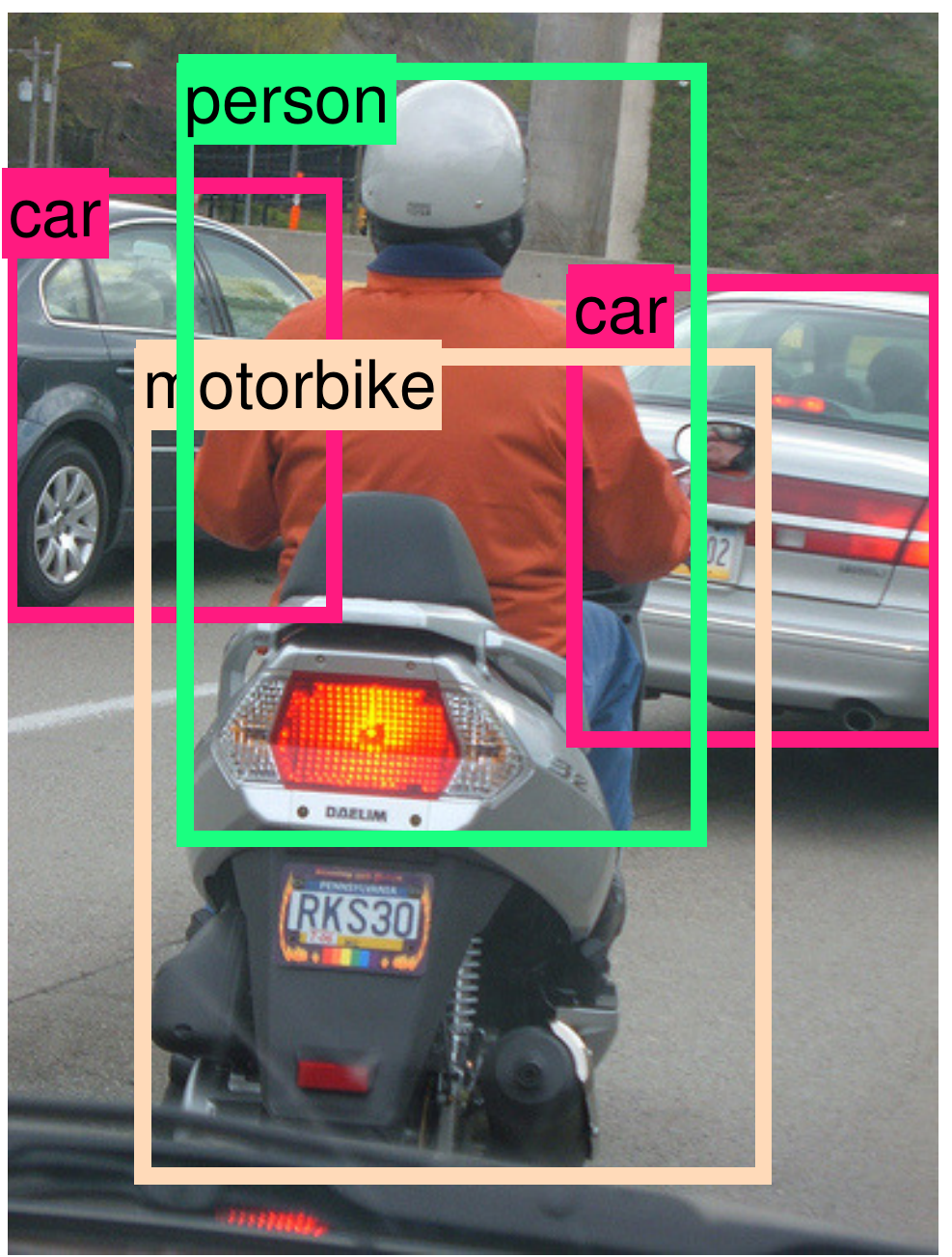}
	\end{subfigure}
	\begin{subfigure}[b]{0.210\textwidth}
		\includegraphics[width=1\textwidth]{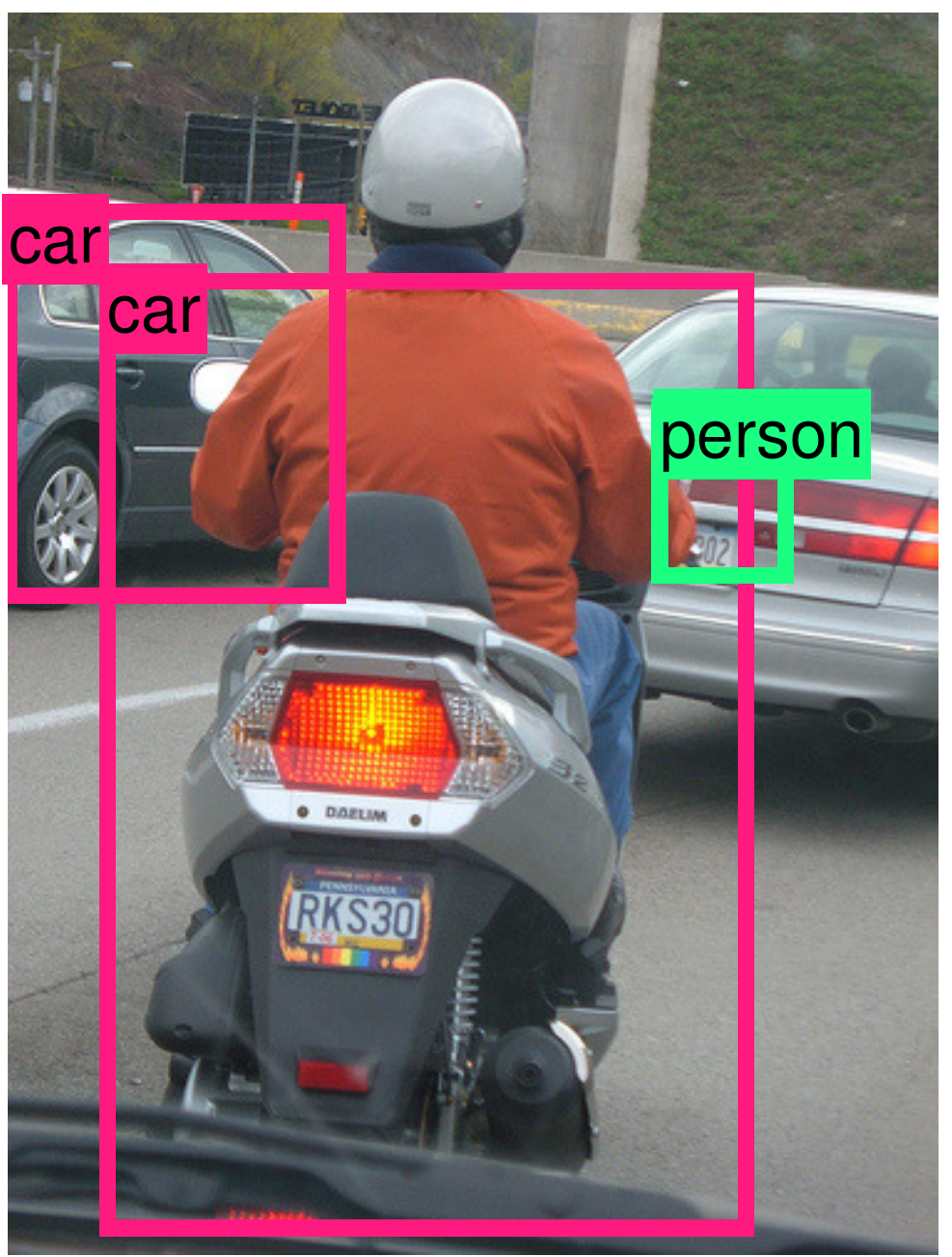}
	\end{subfigure}	
	\begin{subfigure}[b]{0.210\textwidth}
		\includegraphics[width=1\textwidth]{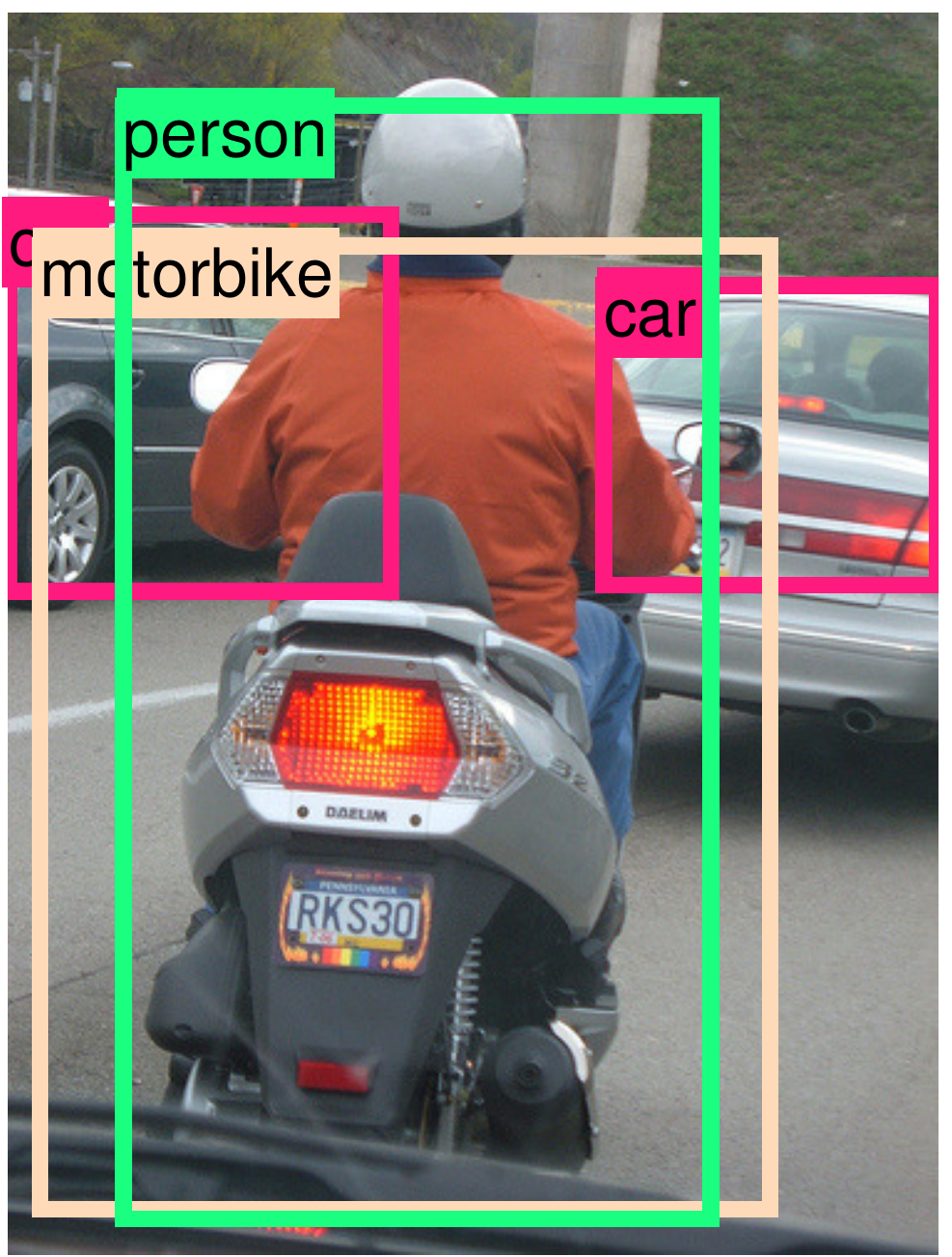}
	\end{subfigure}	
	\begin{subfigure}[b]{0.210\textwidth}
		\includegraphics[width=1\textwidth]{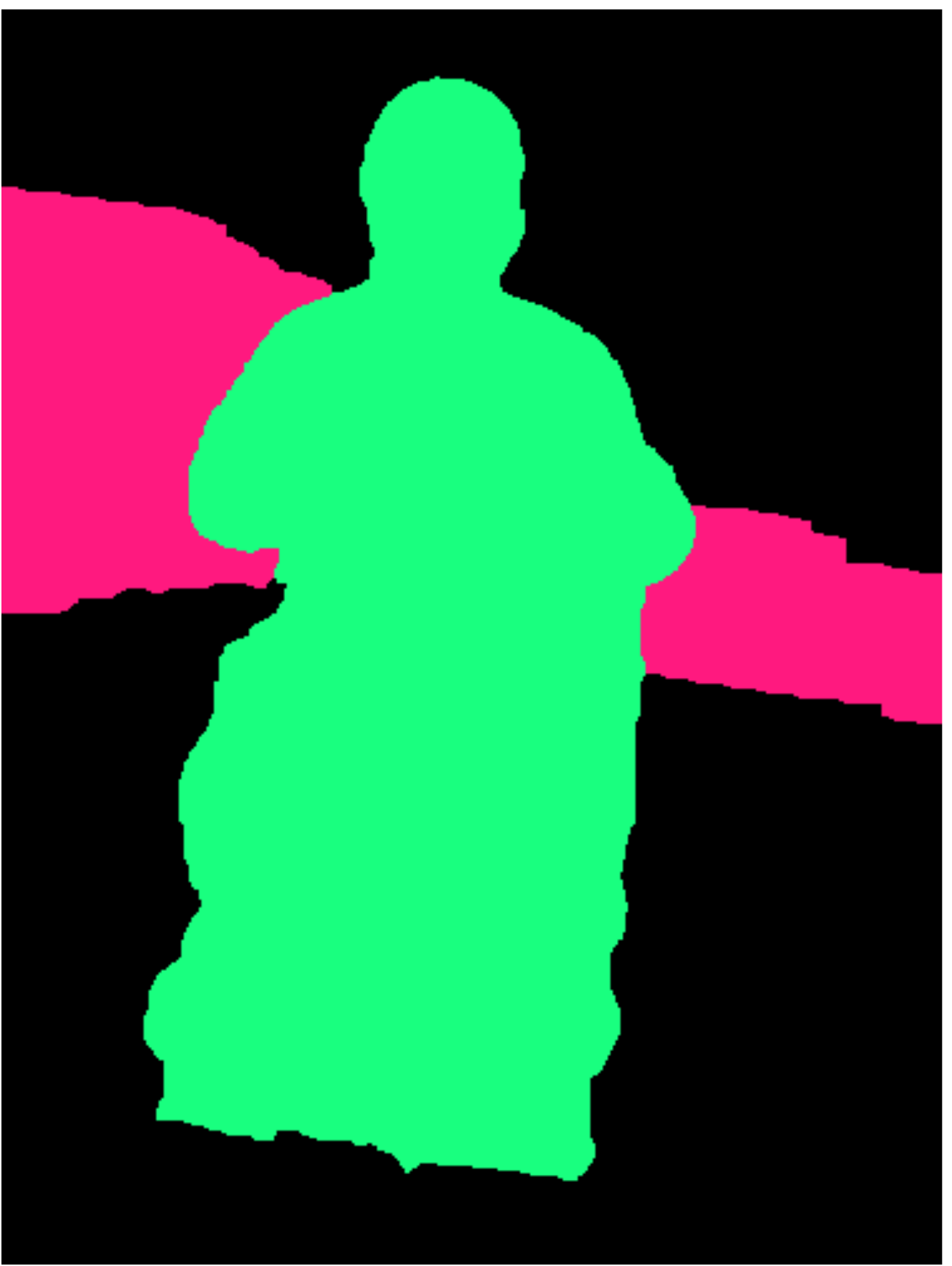}
	\end{subfigure}	
	\vspace{0.1cm}	
	\begin{subfigure}[b]{0.210\textwidth}
		\includegraphics[width=1\textwidth]{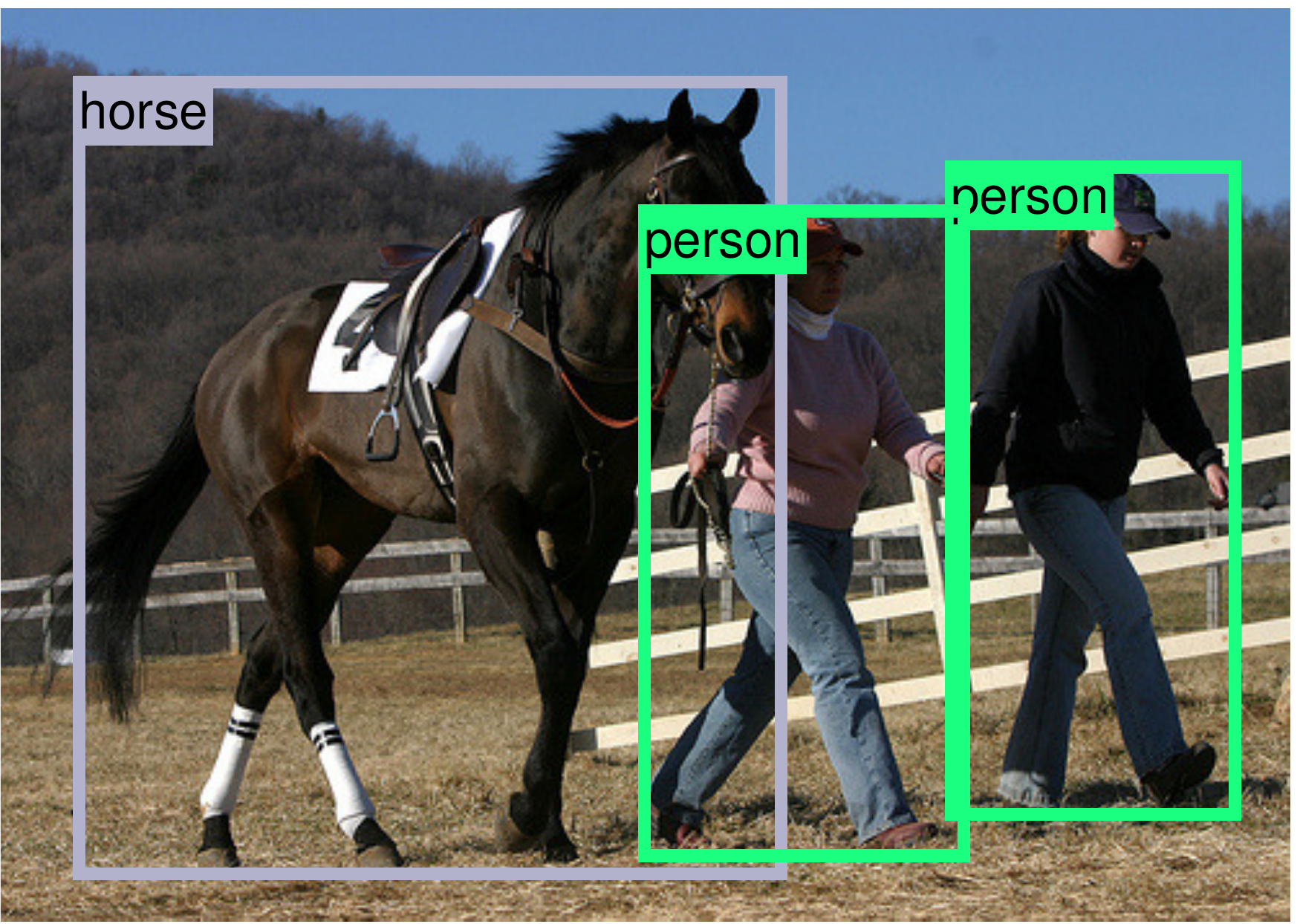}
	\end{subfigure}
	\begin{subfigure}[b]{0.210\textwidth}
		\includegraphics[width=1\textwidth]{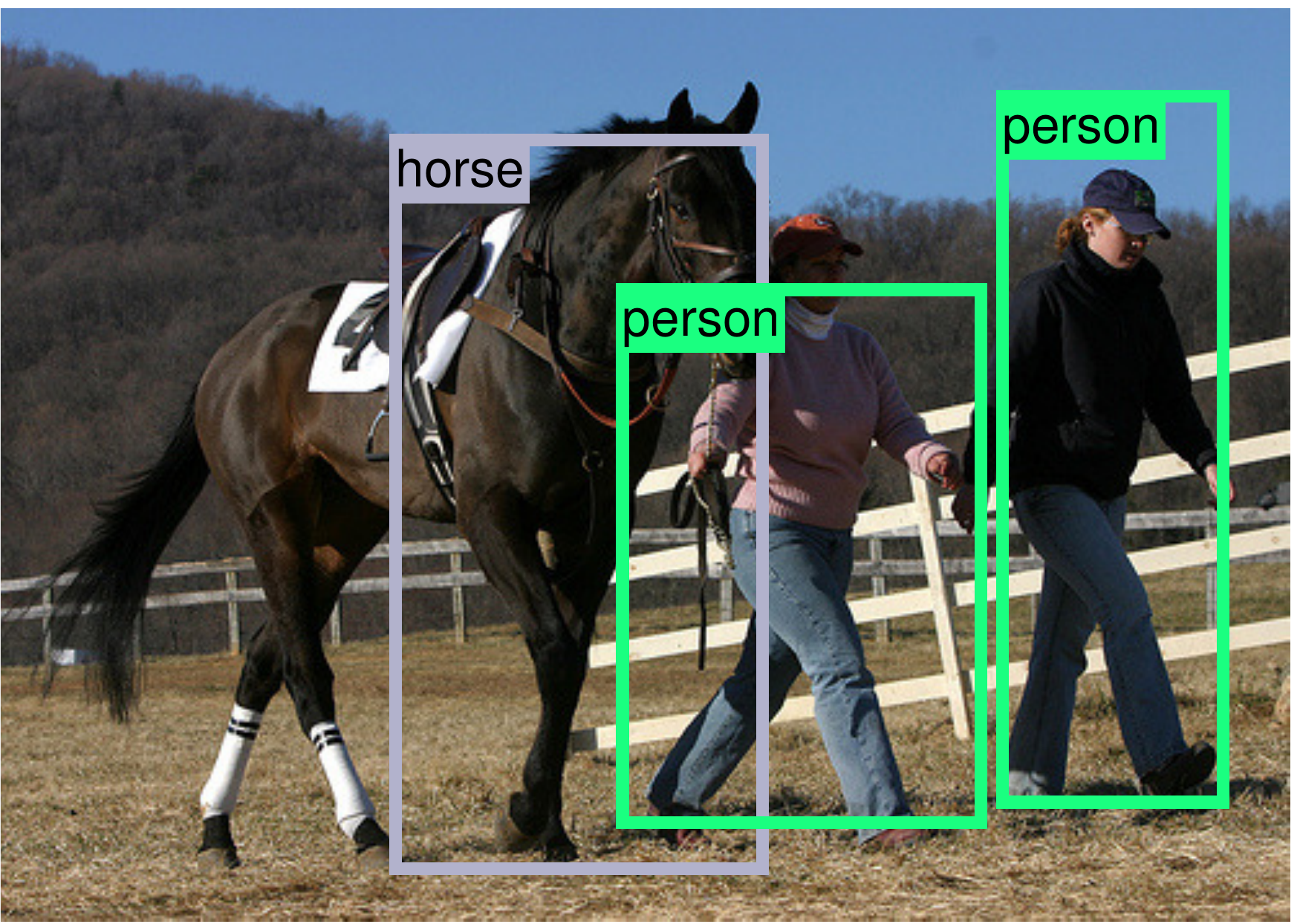}
	\end{subfigure}	
	\begin{subfigure}[b]{0.210\textwidth}
		\includegraphics[width=1\textwidth]{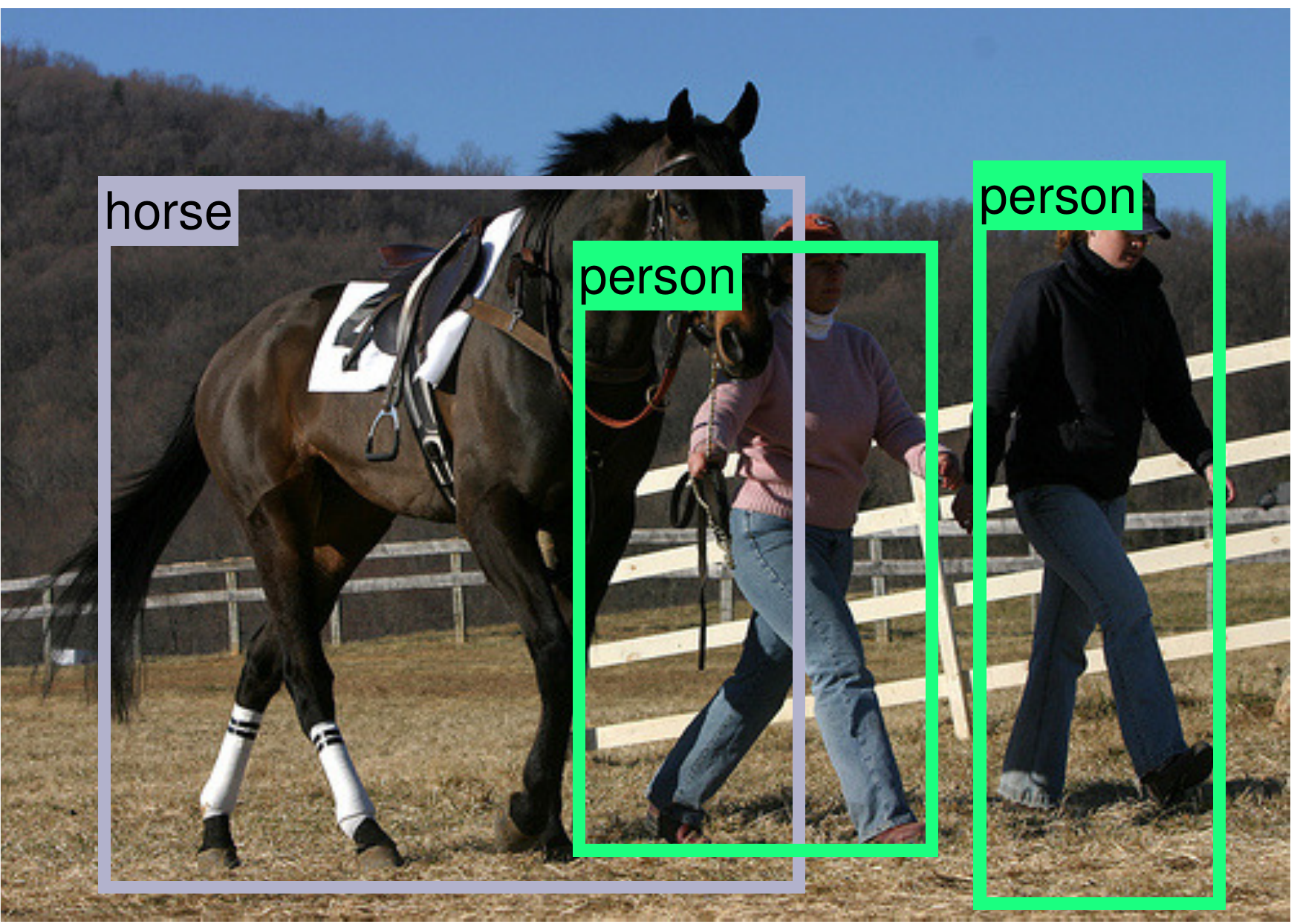}
	\end{subfigure}	
	\begin{subfigure}[b]{0.210\textwidth}
		\includegraphics[width=1\textwidth]{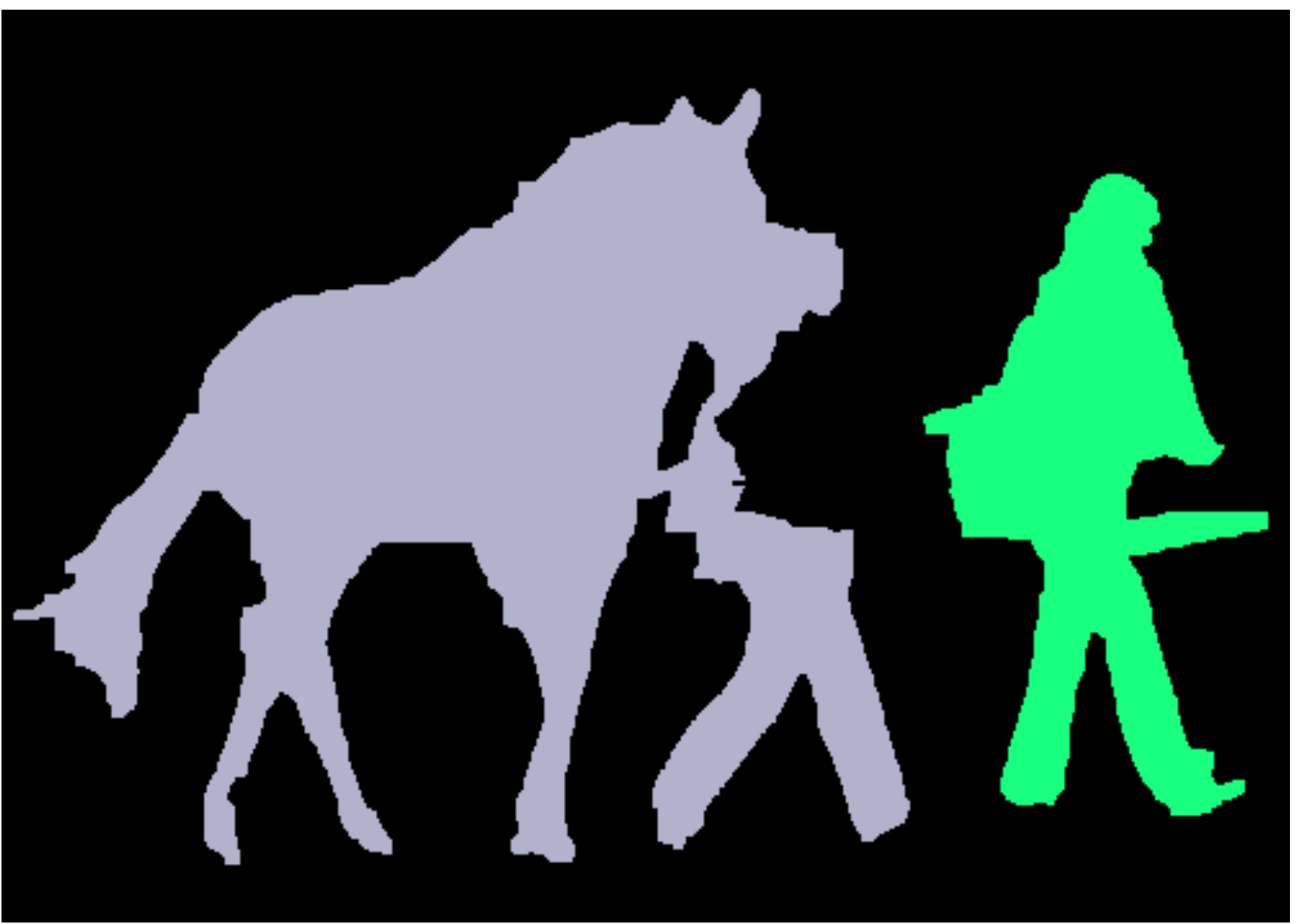}
	\end{subfigure}	
\vspace{0.1cm}	
	\begin{subfigure}[b]{0.210\textwidth}
		\includegraphics[width=1\textwidth]{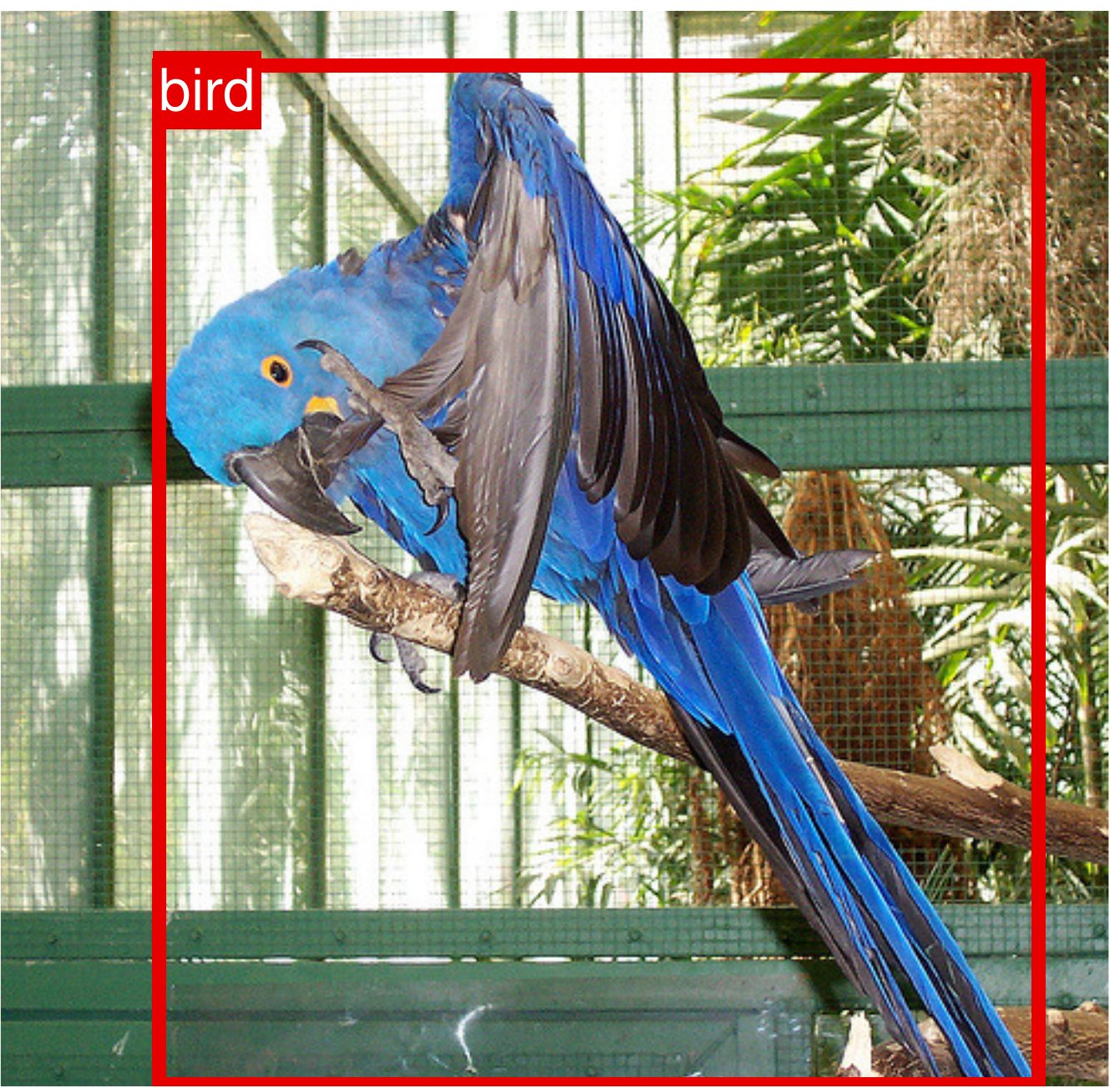}
		\caption{GroundTruth}
	\end{subfigure}
	\begin{subfigure}[b]{0.210\textwidth}
		\includegraphics[width=1\textwidth]{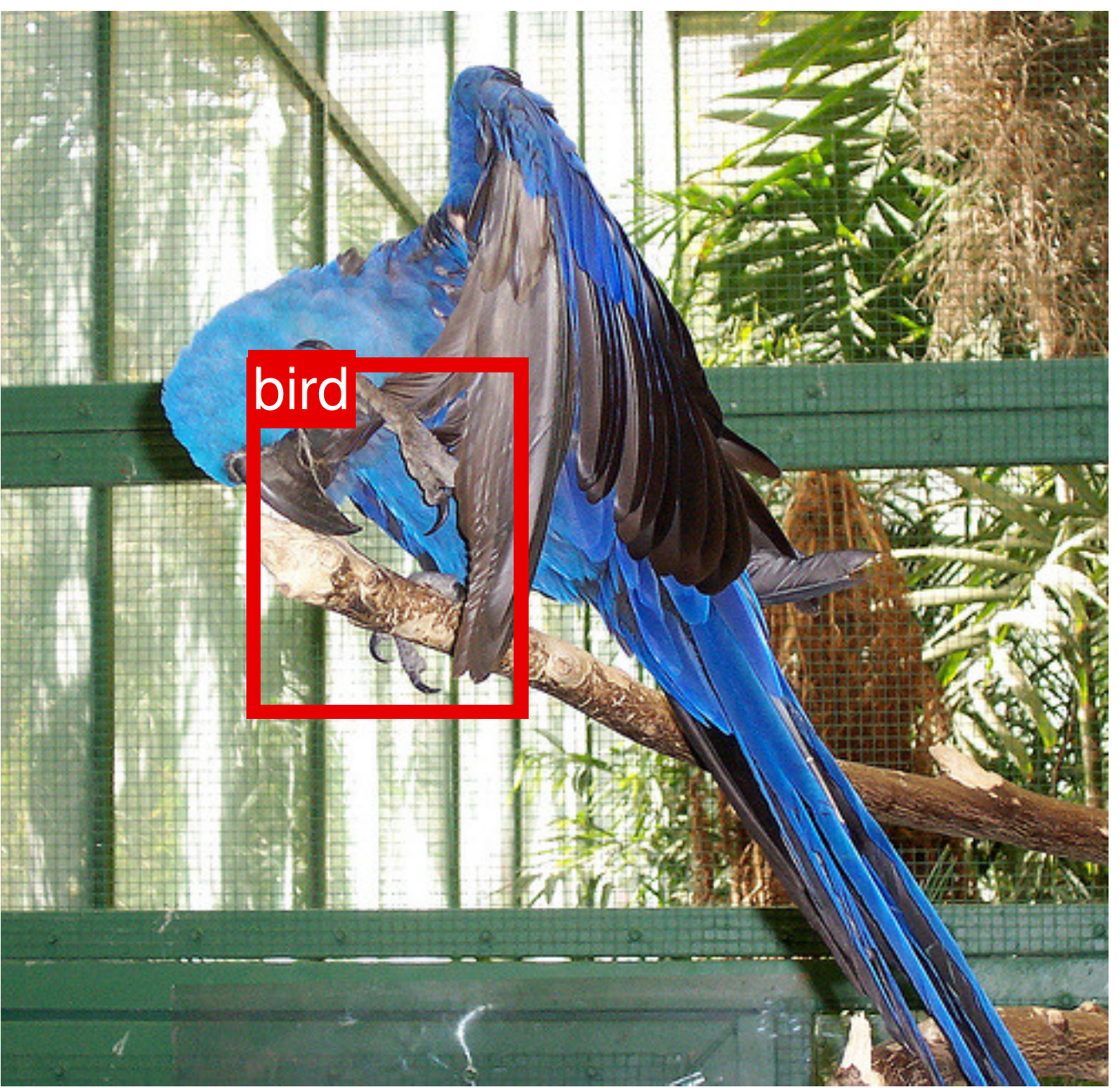}
		\caption{RCNN detection}
	\end{subfigure}	
	\begin{subfigure}[b]{0.210\textwidth}
		\includegraphics[width=1\textwidth]{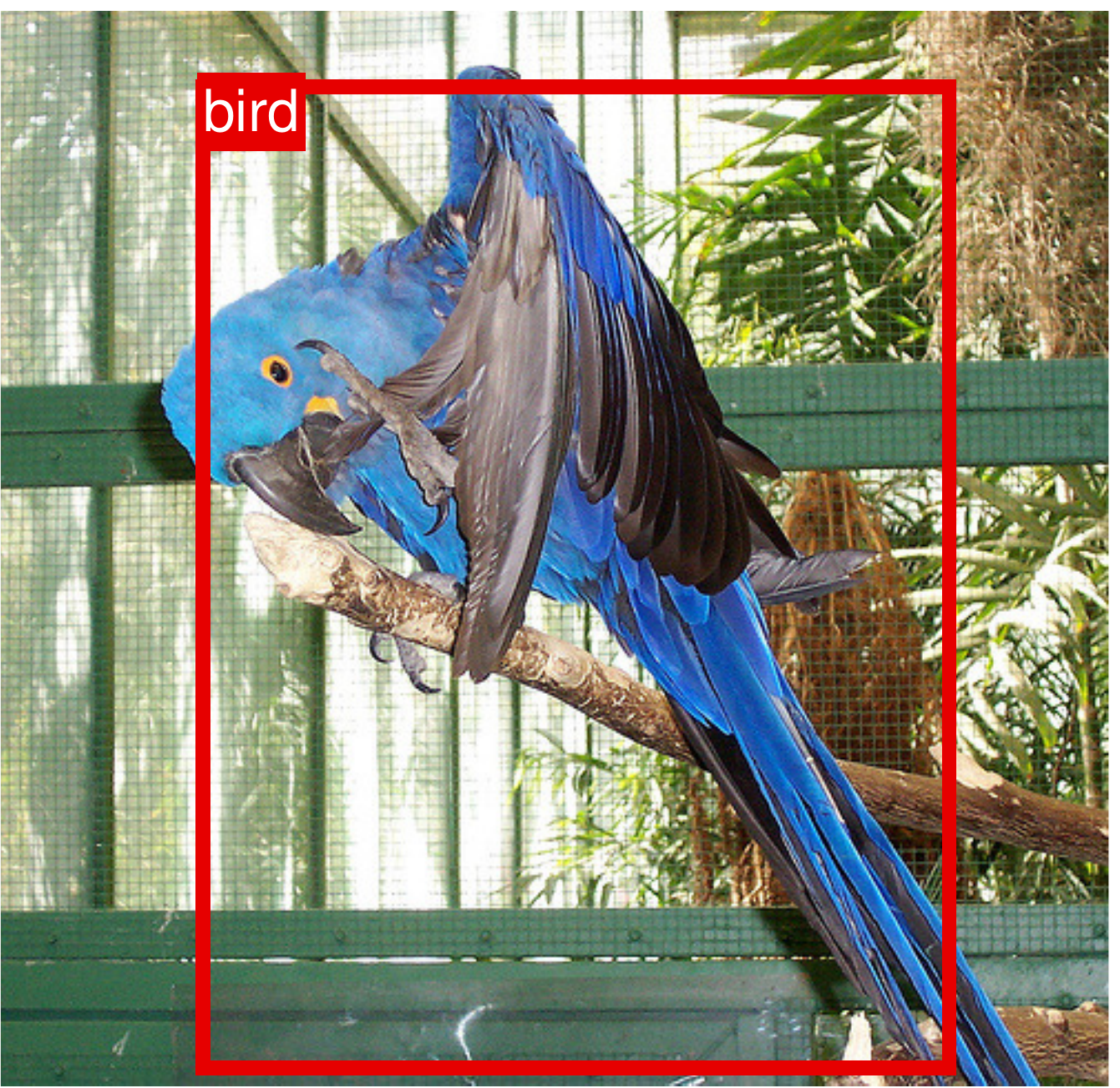}
		\caption{segDeepM detection}
	\end{subfigure}	
	\begin{subfigure}[b]{0.210\textwidth}
		\includegraphics[width=1\textwidth]{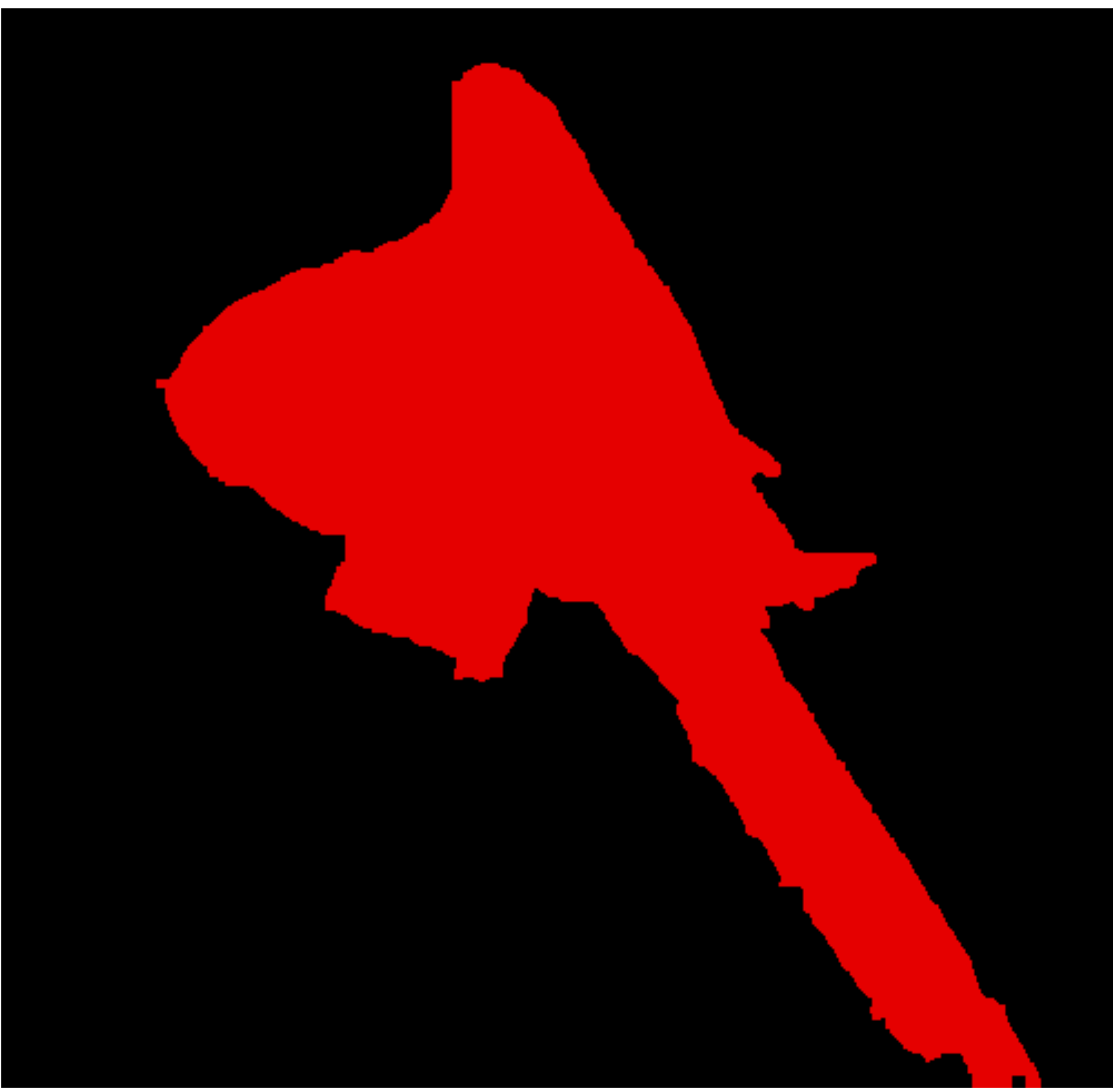}
		\caption{best segments selected}
	\end{subfigure}	
	\vspace{-3mm}
\caption{Qualitative results. We show the top scoring detections for each ground-truth class. For our method, we also show the segment chosen by our model.}\label{figure:qual}
\vspace{-2.5mm}
\end{figure*}

Our full approach, in the setting without any post-processing,  outperforms the strong baseline detector~\cite{girshick2013rich} by $6.3\%$, a significant improvement. After post-processing, the improvement is slightly lower, achieving a $4.8\%$ performance gain. We note that we improve over the baseline in 19 out of 20 object classes. The PR curves for the first 10 classes are shown in Figure~\ref{figure:pr} and the qualitative results are shown in Figure~\ref{figure:qual}. A detailed error analysis as proposed in~\cite{Hoiem14} of R-CNN and our detector is shown in Figure~\ref{fig:error}.

\vspace{-1mm}
\paragraph{Performance vs. grid size and \# of segments.} We evaluate the influence of different grid sizes and different number of CPMC segments per image. For each CPMC segment we compute the best O2P ranking score across  all classes, and choose the top $\eta$ segments according to these scores. Figure~\ref{ap_vs_numseg}, left panel, shows that the highest performance gain is due to the few best scoring segments. The differences are minor across different values of $K$ and $\eta$. Interestingly, the model performs worse with more segments and a coarse grid, as additional low-quality segments add noise and make L-SVM training more difficult. When using a finer grid, the performance peaks when more segments are use, and achieves an overall improvement over a single-cell grid.

\paragraph{Performance \wrt expansion ratio.} We evaluate the influence of the box expansion ratio $\rho$ used in our contextual model. The results for varying values of $\rho$ are illustrated in Figure~\ref{ap_vs_numseg}, right panel. Note that even a small expansion ratio ($10\%$ in each direction) can boost the detection performance by a significant $1.5\%$, and  the performance reaches its peak at $\rho=0.5$. This indicates that richer contextual information leads to a better object recognition. Notice also that the detection performance decreases beyond $\rho=0.5$. This is most likely due to the fact that most contextual boxes obtained this way will cover most or even the full image, and thus the positive and negative training instances in the same image will share the identical contextual features. This confuses our classifier and results in a performance loss. If we take the full image as context, the gain is less than $1\%$. 

\paragraph{Iterative bounding box prediction.} We next study the effect of  iterative bounding box prediction. We report a $1.4\%$ gain over the original R-CNN by 
starting with our set of re-localized boxes (one iteration).  Note that re-localization in the first iteration only affects $52\%$ of boxes (only $52\%$ of boxes change more than $20\%$ from the original set, thus feature re-computation only affects half of the boxes).  This performance gain persists when combined with our full model. If we apply another bounding box prediction as a post-processing step, this approach still obtains a $0.6\%$ improvement over R-CNN with bounding box prediction. In this iteration, re-localization affects $42\%$ of boxes. We have noticed that the performance saturates after two iterations. The second iteration improves mAP by only a small margin (about $0.1\%$). The interesting side result is that, the mean Average Best Overlap (mABO) measure used by bottom-up proposal generation techniques~\cite{van2011segmentation} to benchmark their proposals, remains exactly the same ($85.6\%$) with or without our bounding box prediction, but has a significant impact on the detection performance. This may indicate that mABO is not the best or at least not the only indicator of a good bottom-up grouping technique.

\def\IH{2.95cm}
\begin{figure}[t!]
\begin{minipage}{1\linewidth}
	\centering
	\begin{subfigure}[b]{0.24\linewidth}
		\includegraphics[height=\IH]{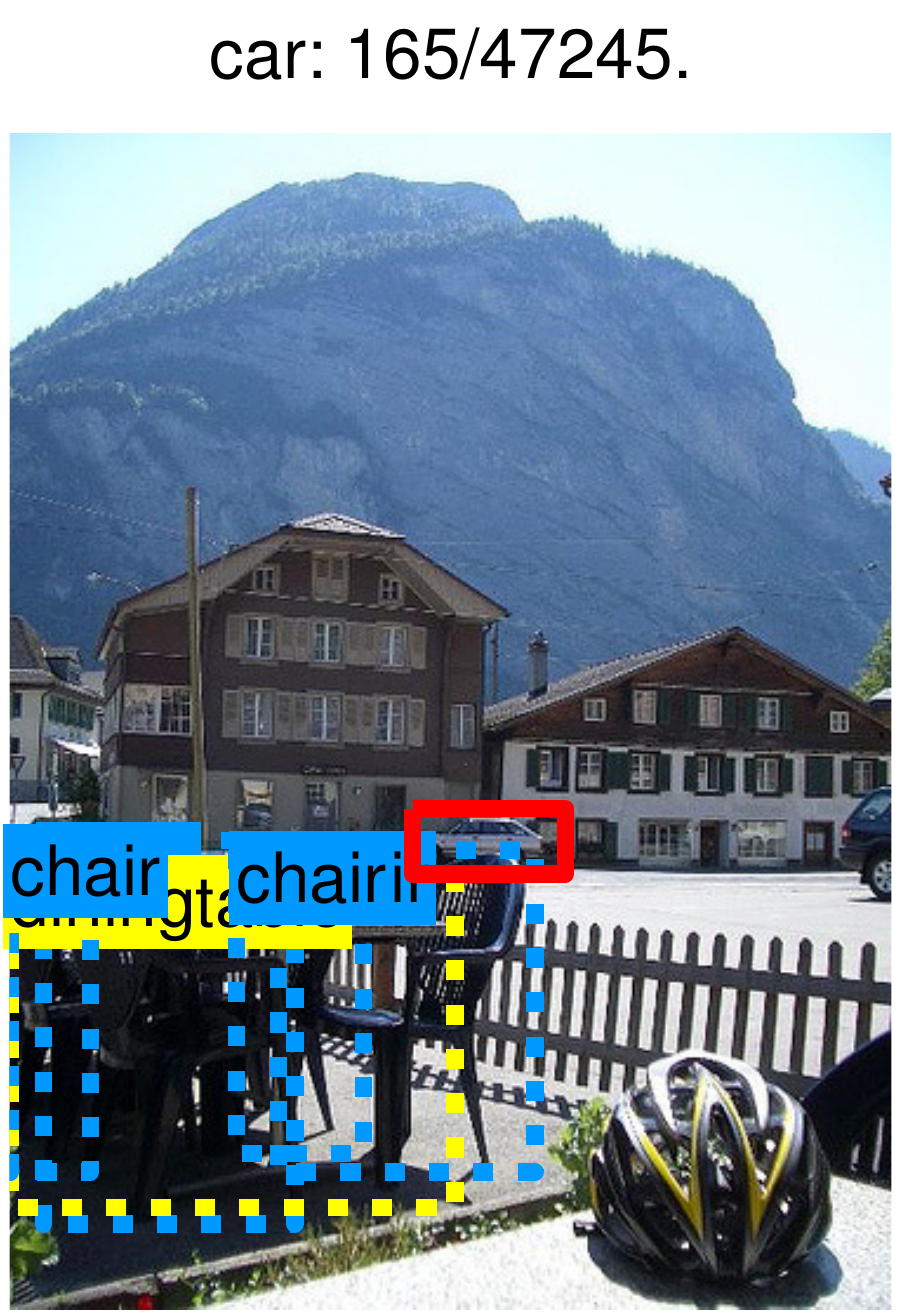}
		\caption{3rd FP for ``car"}\label{figure:missed_dets1}
	\end{subfigure}
	\begin{subfigure}[b]{0.24\linewidth}
		\includegraphics[height=\IH]{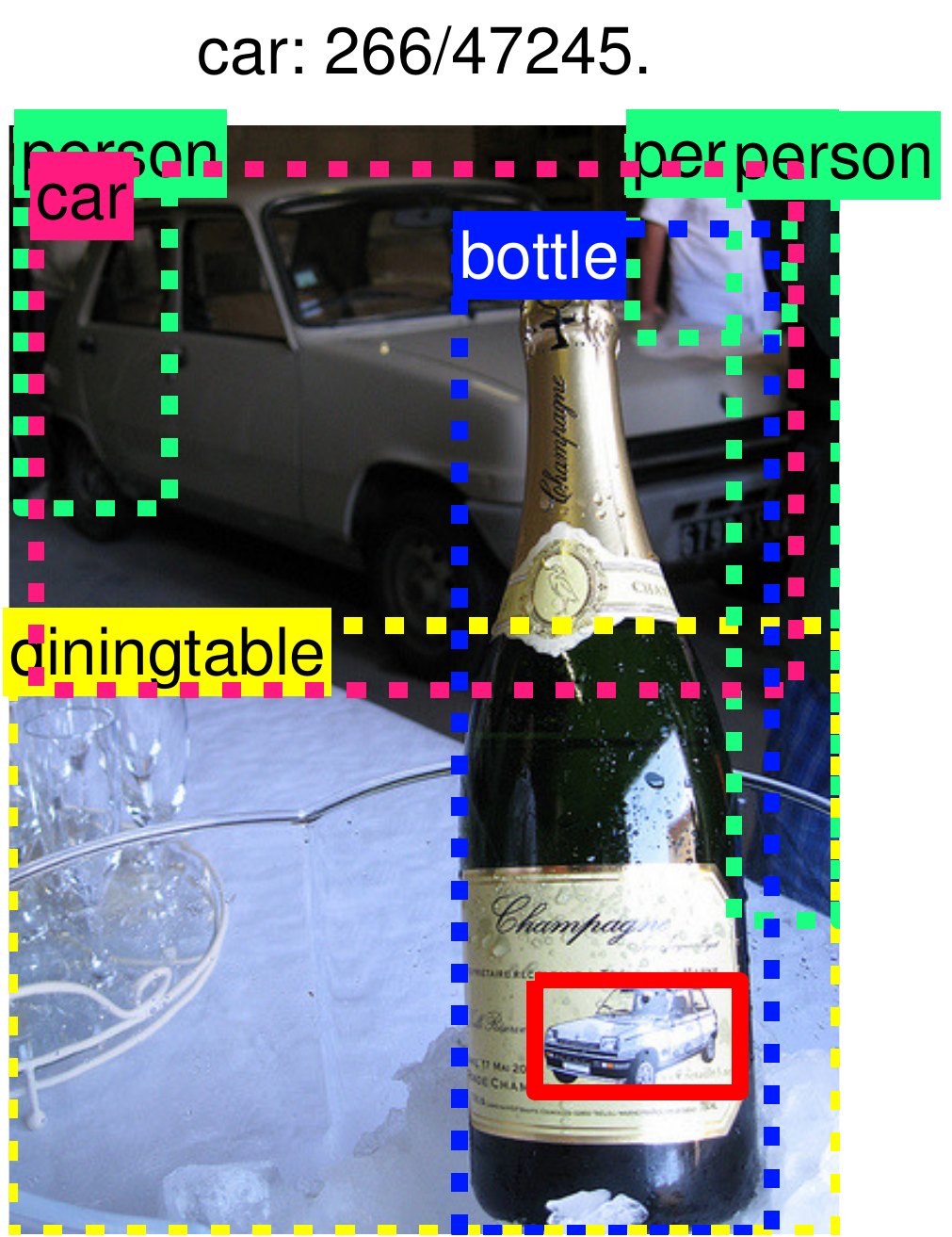}
		\caption{10th FP ``car"}\label{figure:missed_dets2}
	\end{subfigure}
	\begin{subfigure}[b]{0.24\linewidth}
		\includegraphics[height=\IH]{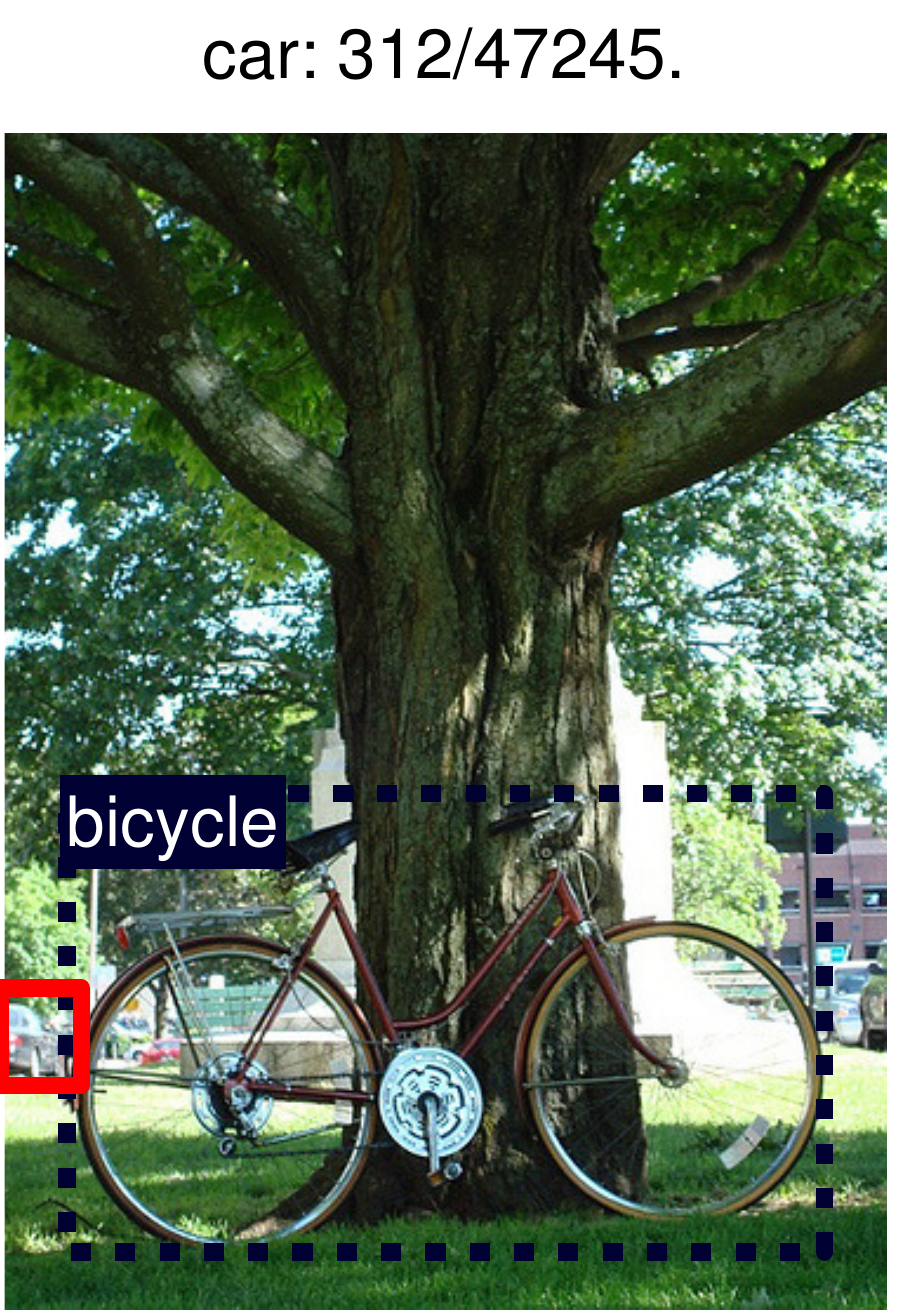}
		\caption{16th FP ``car"}\label{figure:missed_dets3}
	\end{subfigure}
	\begin{subfigure}[b]{0.24\linewidth}
		\includegraphics[height=\IH]{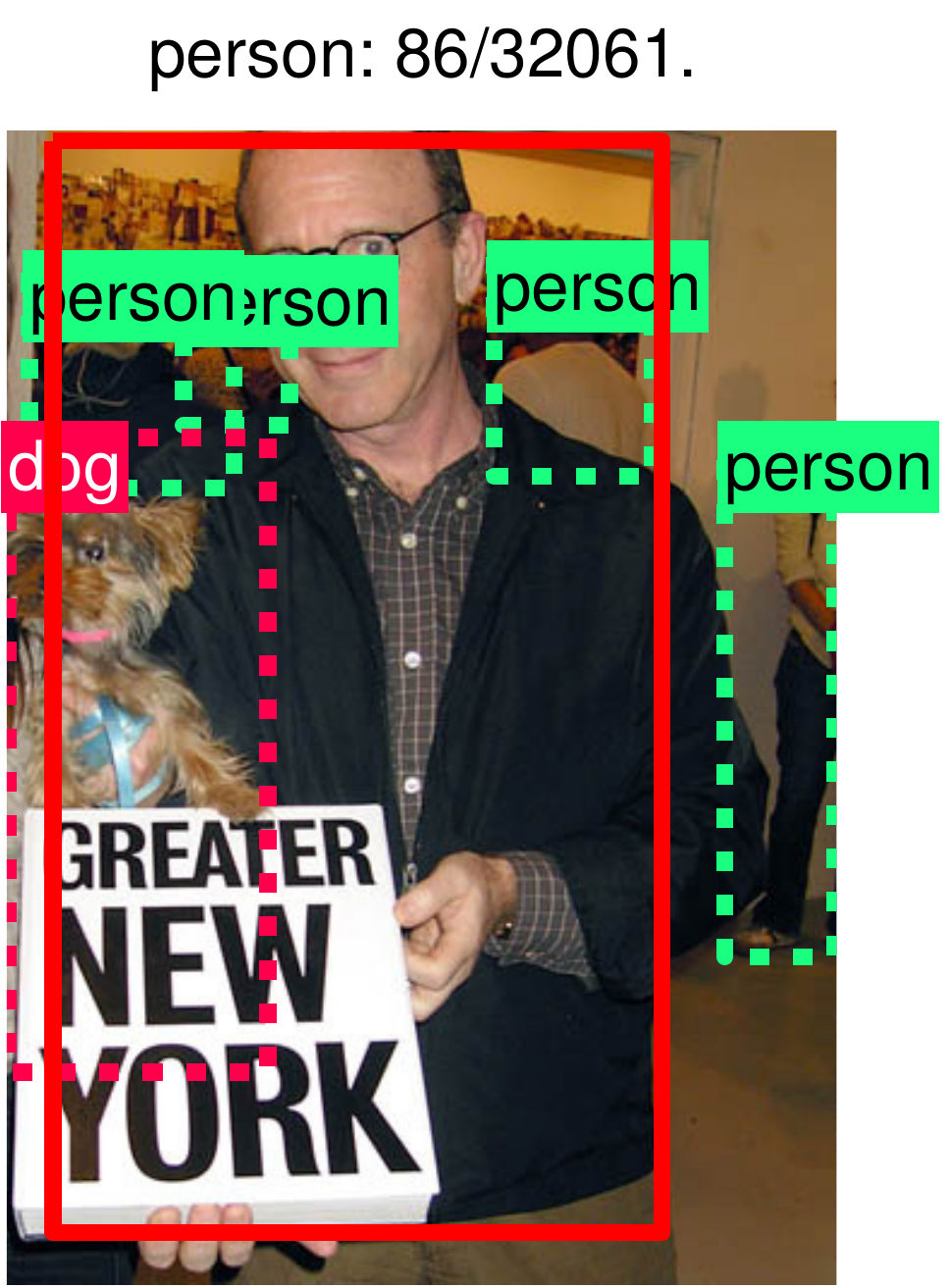}
		\caption{3rd FP ``person"}\label{figure:missed_dets4}
	\end{subfigure}
	\vspace{-3mm}
	\caption{Missing annotations detected by segDeepM. The red solid rectangle indicates segDeepM detection and dashed rectangle represents GT. We show all GT labels, included those marked as 'difficult'.}
	\label{figure:missed_dets}
	\vspace{-4mm}
	\end{minipage}
\end{figure}

\paragraph{Missing annotations.} An interesting issue arises when analyzing the top false-positives of our segDeepM. We have noticed that a non-neglible number of false-positives are due to missing annotations in PASCAL's ground-truth. Some examples are shown in Figure~\ref{figure:missed_dets}. These missed annotations are mostly due to small objects (Figure~\ref{figure:missed_dets1},~\ref{figure:missed_dets3}), ambiguous definition of an ``object'' (Figure~\ref{figure:missed_dets2}), and labelers' mistakes (Figure~\ref{figure:missed_dets4}). 
While missing annotations were not an issue a few years ago when performance was at $30\%$, it is becoming a problem now, indicating that perhaps a re-annotation is needed.

\subsection{Comparison with State-of-The-Art on Test}
\label{sec:test}

We evaluate our approach on the PASCAL VOC 2010 $test$ subset in Table~\ref{table:test}. For this experiment we trained our segDeepM model, as well as its potentials (the CPMC class regressor) on the PASCAL VOC $trainval$ subset using the best parameters tuned on the $train$/$val$ split. We only submitted one result to the evaluation server, thus no tuning on the test set was involved. 
Table~\ref{table:test} shows results of our full segDeepM  (including all post-processing steps).
We achieve a $4.1\%$ improvement over R-CNN with a 7-layer network, and a $1.4\%$ over the best reported method using a 7-layer network. Notice that the best results on the current leader board are achieved by the recently released 16-layer network~\cite{verydeep}. This network has $160$ million parameters, compared to $60$ million parameters used in our network. 
Our approach, with only a few additional parameters, scores rather high relative to the much larger network. Our result is ``only'' $2\%$ lower than the very deep state-of-the-art.
 
We also run our method using a recently released 16-layer OxfordNet~\cite{verydeep}. The results on $train$/$val$ and $trainval$/$test$ are shown in Table~\ref{table:val_16} and Table~\ref{table:test} respectively. On the $test$ set, our segDeepM achieves $67.2\%$ mean AP and outperforms others in  $20$ out of $20$ object classes. 

\paragraph{Performance on PASCAL VOC 2012.} We also test our full segDeepM model on PASCAL VOC 2012. We use the parameters tuned on the PASCAL VOC 2010 $train$/$val$ split. The result are reported and compared to the current state-of-the-art in Table~\ref{table:val2012}.

\vspace{-1mm}
\section{Conclusion}
\vspace{-1mm}

We proposed a MRF model that scores appearance as well as context for each detection, and allows each candidate box to select a segment and score the agreement between them. We additionally proposed a sequential localization scheme, where we iterate between scoring our model and re-positioning the box (changing the spatial scope of the input to the model). We demonstrated that our approach achieves a significant boost over the RCNN baseline, $4.1\%$ on PASCAL VOC 2010 test in the 7-layer setting and $4.3\%$ in the 16-layer setting. The final result places segDeepM at the top of the current PASCAL's leaderboard.

{\bf Acknowledgments.} This research was supported in part by  Toyota. The GPUs used in this research were generously donated by  NVIDIA Corporation.

\clearpage
{\small
\bibliographystyle{ieee}
\bibliography{segDeepM}

\begin{thebibliography}{10}\itemsep=-1pt

\bibitem{bourdev10}
L.~Bourdev, S.~Maji, T.~Brox, and J.~Malik.
\newblock Detecting people using mutually consistent poselet activations.
\newblock In {\em ECCV}, 2010.

\bibitem{carreira2012semantic}
J.~Carreira, R.~Caseiro, J.~Batista, and C.~Sminchisescu.
\newblock Semantic segmentation with second-order pooling.
\newblock In {\em ECCV}, pages 430--443. Springer, 2012.

\bibitem{carreira2010constrained}
J.~Carreira and C.~Sminchisescu.
\newblock Constrained parametric min-cuts for automatic object segmentation.
\newblock In {\em CVPR}, pages 3241--3248. IEEE, 2010.

\bibitem{PartsCVPR14}
X.~Chen, R.~Mottaghi, X.~Liu, N.-G. Cho, S.~Fidler, R.~Urtasun, and A.~Yuille.
\newblock Detect what you can: Detecting and representing objects using
  holistic models and body parts.
\newblock In {\em CVPR}, 2014.

\bibitem{dai12}
Q.~Dai and D.~Hoiem.
\newblock Learning to localize detected objects.
\newblock In {\em CVPR}, 2012.

\bibitem{dong2014towards}
J.~Dong, Q.~Chen, S.~Yan, and A.~Yuille.
\newblock Towards unified object detection and semantic segmentation.
\newblock In {\em ECCV}, pages 299--314. Springer, 2014.

\bibitem{felzenszwalb2010object}
P.~F. Felzenszwalb, R.~B. Girshick, D.~McAllester, and D.~Ramanan.
\newblock Object detection with discriminatively trained part-based models.
\newblock {\em TPAMI}, 32(9):1627--1645, 2010.

\bibitem{FidlerBottomCVPR2013}
S.~Fidler, R.~Mottaghi, A.~Yuille, and R.~Urtasun.
\newblock Bottom-up segmentation for top-down detection.
\newblock In {\em CVPR}, 2013.

\bibitem{girshick2013rich}
R.~Girshick, J.~Donahue, T.~Darrell, and J.~Malik.
\newblock Rich feature hierarchies for accurate object detection and semantic
  segmentation.
\newblock {\em arXiv preprint arXiv:1311.2524}, 2013.

\bibitem{gu09}
C.~Gu, J.~Lim, P.~Arbelaez, and J.~Malik.
\newblock Recognition using regions.
\newblock In {\em CVPR}, 2009.

\bibitem{HariharanSimultaneousECCV2014}
B.~Hariharan, P.~Arbel{\'a}ez, R.~Girshick, and J.~Malik.
\newblock Simultaneous detection and segmentation.
\newblock In {\em ECCV}, 2014.

\bibitem{Hoiem14}
D.~Hoiem, Y.~Chodpathumwan, and Q.~Dai.
\newblock Diagnosing error in object detectors.
\newblock In {\em ECCV}, 2014.

\bibitem{Kiros}
R.~Kiros, R.~Salakhutdinov, and R.~Zemel.
\newblock Multimodal neural language models.
\newblock In {\em ICML}, 2014.

\bibitem{koltun11}
P.~Kr\"{a}henb\"{u}hl and V.~Koltun.
\newblock Efficient inference in fully connected crfs with gaussian edge
  potentials.
\newblock In {\em NIPS}, 2011.

\bibitem{krizhevsky2012imagenet}
A.~Krizhevsky, I.~Sutskever, and G.~E. Hinton.
\newblock Imagenet classification with deep convolutional neural networks.
\newblock In {\em NIPS}, pages 1097--1105, 2012.

\bibitem{ladicky10}
L.~Ladicky, P.~Sturgess, K.~Alahari, C.~Russell, and P.~H. Torr.
\newblock What, where and how many? combining object detectors and crfs.
\newblock In {\em ECCV}, 2010.

\bibitem{maire11}
M.~Maire, S.~X. Yu, and P.~Perona.
\newblock Object detection and segmentation from joint embedding of parts and
  pixels.
\newblock In {\em ICCV}, 2011.

\bibitem{Memisevic}
R.~Memisevic and C.~Conrad.
\newblock Stereopsis via deep learning.
\newblock In {\em NIPS}, 2011.

\bibitem{mottaghirole}
R.~Mottaghi, X.~Chen, X.~Liu, N.-G. Cho, S.-W. Lee, S.~Fidler, R.~Urtasun, and
  A.~Yuille.
\newblock The role of context for object detection and semantic segmentation in
  the wild.
\newblock {\em CVPR}, 2014.

\bibitem{parkhi11}
O.~Parkhi, A.~Vedaldi, C.~V. Jawahar, and A.~Zisserman.
\newblock The truth about cats and dogs.
\newblock In {\em ICCV}, 2011.

\bibitem{verydeep}
K.~Simonyan and A.~Zisserman.
\newblock Very deep convolutional networks for large-scale image recognition.
\newblock In {\em arXiv:1409.1556}, 2014.

\bibitem{GoogLeNet}
C.~Szegedy, W.~Liu, Y.~Jia, P.~Sermanet, S.~Reed, D.~Anguelov, D.Erhan,
  V.~Vanhoucke, and A.~Rabinovich.
\newblock Going deeper with convolutions.
\newblock {\em arXiv preprint arXiv:1409.4842}, 2014.

\bibitem{van2011segmentation}
K.~E. Van~de Sande, J.~R. Uijlings, T.~Gevers, and A.~W. Smeulders.
\newblock Segmentation as selective search for object recognition.
\newblock In {\em ICCV}, pages 1879--1886. IEEE, 2011.

\bibitem{Weinzaepfel}
P.~Weinzaepfel, J.~Revaud, Z.~Harchaoui, and C.~Schmid.
\newblock {DeepFlow: Large displacement optical flow with deep matching}.
\newblock In {\em ICCV}, 2013.

\bibitem{yang11}
Y.~Yang, S.~Hallman, D.~Ramanan, and C.~Fowlkes.
\newblock Layered object models for image segmentation.
\newblock {\em PAMI}, 2011.

\bibitem{Yao12}
J.~Yao, S.~Fidler, and R.~Urtasun.
\newblock Describing the scene as a whole: Joint object detection, scene
  classification and semantic segmentation.
\newblock In {\em CVPR}, 2012.

\end{thebibliography}
}

\end{document}